%% file: main.tex
\documentclass{article}

\usepackage{iclr2027_conference,times}
\newif\ifderailpreprint
\derailpreprinttrue
\ifderailpreprint
  \iclrfinalcopy
\fi

\usepackage{amssymb}
\usepackage{graphicx}   
\usepackage{float}
\usepackage{booktabs}
\usepackage{adjustbox}
\usepackage{multirow}
\usepackage{makecell}
\usepackage{amsmath}
\usepackage{wrapfig}
\usepackage{enumitem}
\usepackage{color}
\usepackage{xcolor}
\usepackage{titletoc}
\usepackage[normalem]{ulem}

\newcommand{\rev}[1]{#1}
\newcommand{\revdel}[1]{}
\usepackage{subcaption}
\usepackage{hyperref}
\ifderailpreprint
  \hypersetup{hidelinks}
\fi
\usepackage{url}
\graphicspath{{../}}
\definecolor{darkgreen}{RGB}{0,118,42}
\usepackage{placeins}

\ifderailpreprint
\title{DeRail: Escaping Trajectory Priors via Counterfactual Action Steering for Mid-Execution Instruction Switching\par
  \vspace{8pt}% Keep the added separation from the author block.
}
\DeclareMathSizes{10.5}{10}{7}{5}
\author{%
  \parbox{\textwidth}{%
    \normalfont\centering\fontsize{10.5}{13}\selectfont
    \textbf{Yitong Hong}$^{1,2}$ \quad \textbf{Xiangdong Feng}$^{2,3}$ \quad \textbf{Xiaoxi Sun}$^{4}$ \quad \textbf{Senqiao Yang}$^{5}$\\
    \textbf{Boyao Han}$^{6}$ \quad \textbf{Zhejian Yang}$^{7}$ \quad \textbf{Shiyang Zhou}$^{2,4}$ \quad \textbf{Yongcai Wang}$^{1,8}$\\
    \textbf{Pengguang Chen}$^{9}$ \quad \textbf{Shu Liu}$^{9}$ \quad \textbf{Li Jiang}$^{2,6}$ \quad \textbf{Zhuotao Tian}$^{2,4}$\\[0.2cm]
    {\fontsize{10}{12}\selectfont
      $^{1}$RUC \quad $^{2}$SLAI \quad $^{3}$BIT \quad $^{4}$HITSZ \quad $^{5}$CUHK \quad $^{6}$CUHKSZ \quad $^{7}$JLU \quad $^{8}$RIAS \quad $^{9}$SmartMore\\[3pt]
      \textrm{yitonghong@slai.edu.cn} \qquad \textrm{tianzhuotao@hit.edu.cn}\endgraf}%
  }%
}
\else
\title{DeRail: Escaping Trajectory Priors via Counterfactual Action Steering for Mid-Execution Instruction Switching}

\author{
  \parbox{5.5in}{
    \normalfont\centering
    Yitong Hong$^{1,2}$ \quad Xiangdong Feng$^{2,3}$ \quad Xiaoxi Sun$^{4}$ \quad Senqiao Yang$^{5}$\\
    Boyao Han$^{6}$ \quad Zhejian Yang$^{7}$ \quad Shiyang Zhou$^{2,4}$ \quad Yongcai Wang$^{1,8}$\\
    Pengguang Chen$^{9}$ \quad Shu Liu$^{9}$ \quad Li Jiang$^{2,6}$ \quad Zhuotao Tian$^{2,4}$\\[3pt]
    {\small $^{1}$RUC \quad $^{2}$SLAI \quad $^{3}$BIT \quad $^{4}$HITSZ \quad $^{5}$CUHK \quad $^{6}$CUHKSZ}\\
    {\small $^{7}$JLU \quad $^{8}$RIAS \quad $^{9}$SmartMore}\\[2pt]
    {\small \texttt{yitonghong@slai.edu.cn} \quad \texttt{tianzhuotao@hit.edu.cn}}
  }
}

\fi

\begin{document}
\ifderailpreprint
  \begingroup
    % Remove only the title's outer tabular padding; body tables are unchanged.
    \setlength{\tabcolsep}{0pt}%
    \vspace*{-27pt}% Move the title closer to the existing header rule.
    \maketitle
  \endgroup
  \fancyhead{}
  \fancyhead[L]{\textit{Preprint}}
\else
  \maketitle
\fi

\begin{abstract}
    Robots operating in the real world must often adapt when user instructions change mid-execution, yet current instruction-conditioned robot policies, including Vision-Language-Action (VLA) models and World-Action Models (WAMs), are largely evaluated under fixed-instruction episodes. We study mid-execution instruction switching and identify trajectory hijacking as a key failure mode, where the policy fails to redirect sufficiently toward the switched target, with its post-switch behavior remaining fully or partially drawn toward trajectories associated with non-target tasks. We propose Counterfactual Action Steering (CAS), a training-free, model-agnostic inference-time method that identifies interfering non-target tasks, queries the same policy with corresponding counterfactual instructions, and uses target--counterfactual action differences to steer execution away from interfering trajectory priors. Experiments on LIBERO-Goal-Switch and CALVIN-Custom-Switch across five VLA and WAM policies, together with real-robot evaluation, show that CAS consistently improves switching success; for example, on CALVIN-Custom-Switch in the $[15,21]$ switching range, CAS improves $\pi_{0.5}$ from 5.8\% to 54.6\%. Successful CAS-guided rollouts further provide higher-quality self-training supervision than vanilla successful rollouts with the same number of trajectories, yielding stronger policies that generalize to unseen later instruction switches. Code  will be made publicly available.
\end{abstract}
\section{Introduction}
\label{sec:intro}
Instruction-conditioned robot policies, including Vision-Language-Action (VLA) models~\citep{BlackK-RSS-25,black2025pi05,kim2025fine,li2024cogact} and World-Action Models (WAMs)~\citep{li2026causal,yuan2026fast}, have emerged as a promising approach to general-purpose robot manipulation, where a single policy can perform diverse tasks conditioned on natural-language instructions.
Despite this progress, existing benchmark evaluations typically provide a fixed task-level instruction at the beginning of each episode and assess whether the corresponding task is completed. This protocol implicitly assumes that the language command remains unchanged throughout execution~\citep{liu2023libero,li2024evaluating,zhang2025vlabench,chen2025robotwin}.

In real-world manipulation, however, instructions are not always fixed: users may update the task goal after execution has already begun~\citep{chen2026resteer}. For example, a user may first ask a robot to grasp an apple, then change the command to grasp an orange after the robot has already moved toward the apple. This demands robot policies capable of redirecting their behavior from an intermediate state in response to a new instruction. Consistent with recent work~\citep{chen2026resteer}, we observe a striking performance degradation across modern VLA and WAM policies under mid-execution instruction switching, even when both the source and target tasks have been individually well-mastered.

\begin{figure}[t]
    \centering

    \begin{subfigure}[b]{0.25\linewidth}
        \centering
        \includegraphics[width=\linewidth]{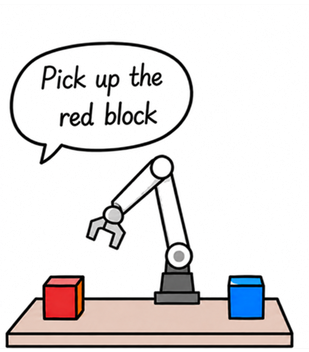}
        \caption{Before switch}
        \label{fig:ts1}
    \end{subfigure}
    \hfill
    \begin{subfigure}[b]{0.25\linewidth}
        \centering
        \includegraphics[width=\linewidth]{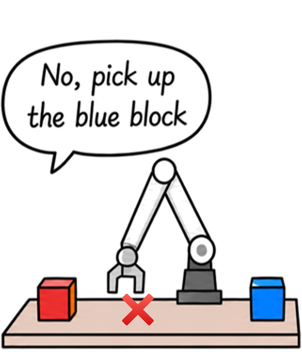}
        \caption{Without CAS: Hijacked}
        \label{fig:ts2_cross}
    \end{subfigure}
    \hfill
    \begin{subfigure}[b]{0.45\linewidth}
        \centering
        \includegraphics[width=\linewidth]{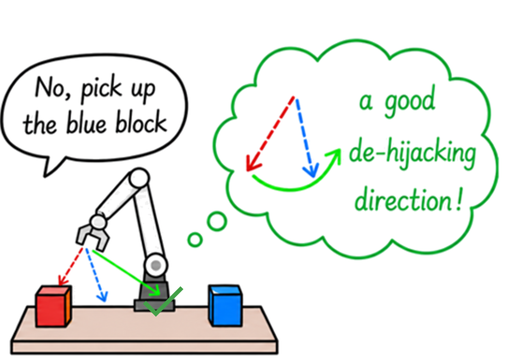}
        \caption{With CAS: Steered Away from Hijacking}
        \label{fig:ts3_tick}
    \end{subfigure}

    \caption{
\textbf{Illustration of trajectory hijacking and the intuition behind CAS.}
The robot is first instructed to pick up the red block and starts moving toward it. The instruction is then changed to pick up the blue block. Without CAS, the policy may fail to redirect sufficiently toward the blue block; instead, its post-switch behavior can remain drawn toward the red-block trajectory, a failure mode we refer to as trajectory hijacking. In (c), the \textcolor{red}{\textbf{dashed red arrow}} denotes the action predicted under the red-block instruction, and the \textcolor{blue}{\textbf{dashed blue arrow}} denotes the action predicted under the blue-block instruction. CAS contrasts these two actions, and their difference provides a corrective direction, producing the \textcolor{darkgreen}{\textbf{green corrected action}} that steers the robot away from the interfering trajectory prior and toward the switched target.
}

    \label{fig:teaser}
\end{figure}

\paragraph{Motivation.}
Fig.~\ref{fig:teaser}~(a,b) illustrates a key failure mode in mid-execution instruction switching---\emph{trajectory hijacking}.
The robot is initially instructed to pick up the red block. After its end-effector has moved toward the red block, the instruction is changed to pick up the blue block. However, the policy often fails to redirect its motion sufficiently toward the blue block. Instead, its post-switch behavior remains fully or partially drawn toward the red-block trajectory, as if its motion were hijacked by this trajectory. We analyze this failure mode in Sec.~\ref{obs}, where evidence suggests that visual cues in the current state can induce trajectory priors that contribute to trajectory hijacking.

To address instruction switching failures, ReSteer~\citep{chen2026resteer} relies on a relatively elaborate multi-stage pipeline for steering-data generation and iterative policy refinement, resulting in considerable engineering and implementation complexity. These limitations motivate us to seek a lightweight and broadly applicable strategy for instruction switching.

Fig.~\ref{fig:teaser}~(c) illustrates the design intuition behind our approach. In this representative example, after the instruction switches from the red block to the blue block, the target-conditioned action may still be insufficient to fully redirect the robot toward the blue block. Nevertheless, compared with the counterfactual action predicted under the red-block instruction, it is still biased more toward the desired blue-block behavior. Motivated by this, we ask: \emph{Can action differences between target and hijacking instructions provide a corrective signal for steering execution away from interfering non-target trajectories during mid-execution instruction switching?}

\paragraph{Our solution.}
In this paper, we propose \emph{Counterfactual Action Steering} (CAS), a training-free, model-agnostic inference-time method for mitigating trajectory hijacking. CAS first estimates which non-target tasks may exert strong trajectory priors at the current robot state by comparing the current end-effector position with task demonstration trajectories. It then queries the policy with the target instruction and with counterfactual instructions corresponding to the identified hijacking tasks. The difference between the target-conditioned action and each counterfactual action provides a corrective direction for steering execution away from the corresponding non-target trajectory prior.

We evaluate CAS across five VLA and WAM policies on our two instruction-switching benchmarks, \emph{LIBERO-Goal-Switch} and \emph{CALVIN-Custom-Switch}, as well as in real-robot experiments. CAS significantly improves mid-execution instruction switching success rates and outperforms PCD~\citep{wu2025policy}; for example, on CALVIN-Custom-Switch in the $[15,21]$ switching range, CAS improves $\pi_0$ from $6.7\%$ to $45.4\%$ and $\pi_{0.5}$ from $5.8\%$ to $54.6\%$.
% We further show that successful CAS-guided rollouts exhibit more effective de-hijacking behavior, offering higher-quality supervision for self-training than vanilla successful rollouts.
Beyond inference-time steering, CAS also supports policy self-improvement: successful CAS-guided rollouts exhibit more effective de-hijacking behavior, providing higher-quality self-training supervision than vanilla successful rollouts. As a result, self-training on CAS-guided rollouts improves the original policy more effectively than vanilla self-training with the same number of trajectories, and this advantage persists when evaluated on unseen, later instruction switches.
To summarize, our contributions are as follows:
\begin{itemize}[leftmargin=0.7cm]
    \item We study and analyze mid-execution instruction switching across multiple VLA and WAM policies, identifying trajectory hijacking as a key failure mode driven by visual trajectory priors.
    
    \item We propose CAS, a \textbf{simple, training-free, and model-agnostic method} for mitigating trajectory hijacking, and demonstrate its effectiveness in simulation and real-robot experiments.
    
    \item We show that CAS-guided rollouts contain more effective de-hijacking behaviors and provide higher-quality self-training supervision than vanilla successful rollouts. With matched numbers of self-training trajectories, CAS-guided self-training yields stronger policies that generalize to unseen, later instruction switches.
\end{itemize}
\section{Problem Formulation and Motivation}
\label{sec:problem}
We first formulate the mid-execution instruction switching problem and then present key observations that motivate our method. Further discussion of related work is provided in Appendix~\ref{app:related_work}.

\subsection{Mid-Execution Instruction Switching}

We study mid-execution instruction switching for instruction-conditioned robot policies, including both VLA and WAM models. Starting from environment step $t=0$, the robot acts under a \emph{source instruction} $I_\mathrm{src}$. At step $t_s$, referred to as the \emph{switching step}, the active instruction is changed to a \emph{target instruction} $I_\mathrm{tgt}$. From step $t_s$ onward, the policy is expected to redirect its behavior from the current state to complete the target task specified by $I_\mathrm{tgt}$. The success of an instruction-switching episode is determined by completion of the target task.

\subsection{Key Observations}
\label{obs}
\begin{figure}[t]
    \centering
    \includegraphics[width=\linewidth]{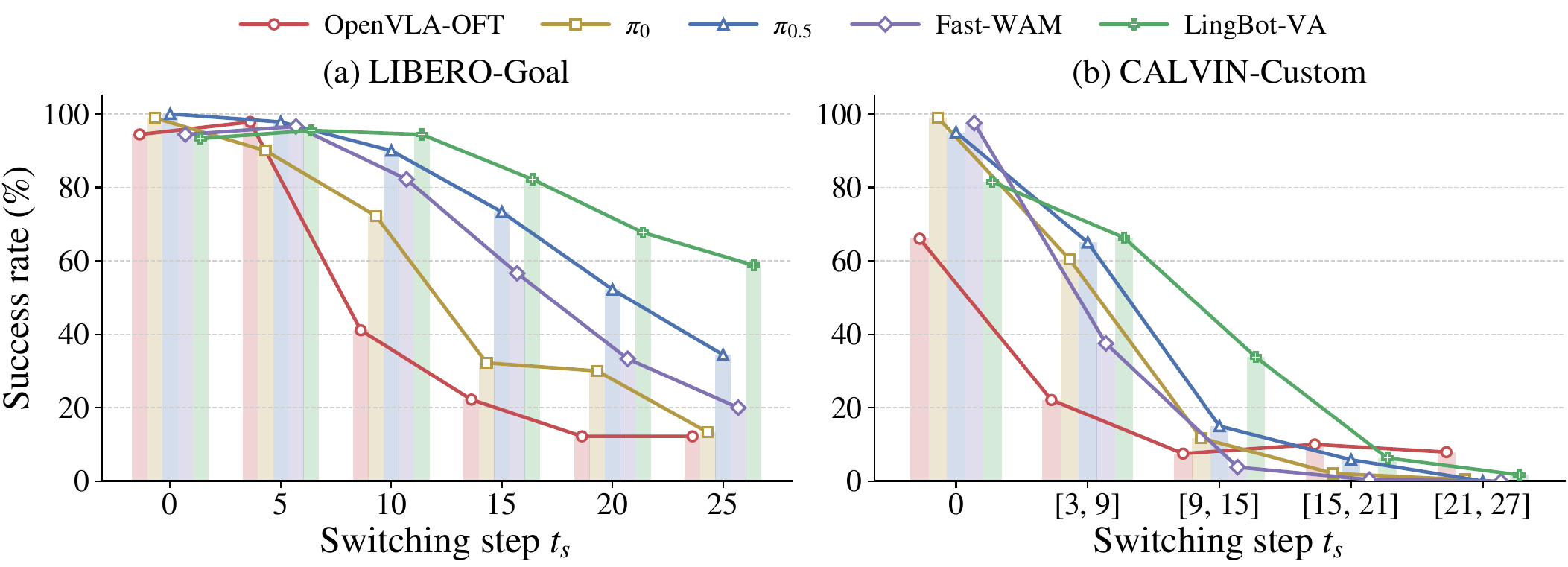}
    \caption{
    \textbf{Success rates of OpenVLA-OFT, $\pi_0$, $\pi_{0.5}$, Fast-WAM and LingBot-VA at different switching steps.} 
    Performance generally degrades with later switches.
    }
    \label{fig:obs1}
    % Keep the attention visualization immediately below Fig. 2 for layout review.
    \par\vspace{3mm}
    \includegraphics[width=0.75\linewidth]{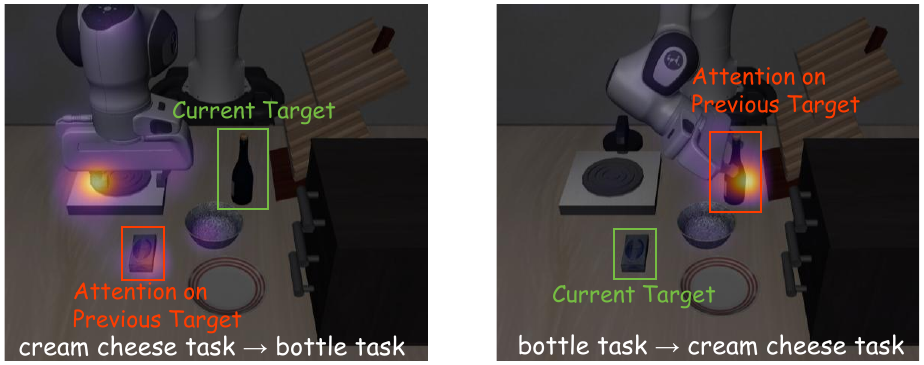}
    \caption{\rev{\textbf{Visualization of attention persistence after instruction switching.}
    We visualize action-to-vision attention in OpenVLA-OFT at the last language-backbone layer (layer 31), averaged over attention heads and action-token positions.
    At an intermediate execution state, we switch only the instruction while keeping the visual observation unchanged.
    In both examples, attention remains concentrated on the gripper and the previous target object (red), rather than shifting toward the new target (green), suggesting that visual attention can remain biased toward previously relevant objects despite instruction changes.}}
    \label{fig:attention_switch}
\end{figure}
\paragraph{Observation 1: Mid-execution instruction switching degrades robot policy performance.}
\label{obs1}
Fig.~\ref{fig:obs1} evaluates OpenVLA-OFT, $\pi_0$, $\pi_{0.5}$, Fast-WAM, and LingBot-VA at different switching steps $t_s$ on LIBERO-Goal and CALVIN-Custom, where $t_s=0$ denotes directly executing the target task without an instruction switch. At the switch, we discard any unexecuted source-conditioned actions in the current action chunk and clear cached temporal context to eliminate explicit carryover from the source phase. Performance generally declines with increasing switching step $t_s$, with particularly severe degradation at larger $t_s$ on CALVIN-Custom, although robustness varies across policies. These results show that current robot policies struggle under mid-execution instruction switching. Further experimental details are provided in Appendix~\ref{app:obs1_details}.

\paragraph{Observation 2: Robot policies are strongly influenced by visual priors.}
\label{obs2}
One possible explanation for this degradation is that policy behavior may be insufficiently sensitive to language instructions. This motivates us to examine how modalities affect policy behavior. To this end, we conduct inference-time modality interventions on LIBERO. We separately perturb vision, language, and, when available, proprioception: we replace visual inputs with a constant mean image, task instructions with ``remain still'', and proprioceptive inputs with the initial robot state. Across policies, we observe a clear contrast between language and visual interventions. When task-specific instructions are replaced, policies often retain non-trivial success rates and continue to exhibit recognizable task-like manipulation behaviors. In contrast, when visual inputs are disrupted, policies rarely exhibit recognizable task-like behaviors, instead producing erratic motions, repetitive movements along nearly fixed directions, or movements with substantially reduced magnitudes. Consistent with this behavioral degradation, success rates under visual intervention collapse to near zero, even though object positions in LIBERO are nearly fixed.

We further finetune the policy on the instruction ``remain still''. When the instruction is switched to ``remain still'' mid-execution, the policy often still fails to stop. For example, when the visual observation shows the gripper already close to an object, the policy tends to continue the grasping behavior cued by the observed gripper--object configuration, moving further toward the object instead of remaining stationary. These results suggest that robot policies are strongly influenced by visual priors.

\rev{Consistent with this behavioral persistence, we observe that visual attention in OpenVLA-OFT can remain focused on the previous target after an instruction switch (Fig.~\ref{fig:attention_switch}). Further analyses of visual priors are provided in Appendix~\ref{app:obs2}. Interestingly, we find that visual attention in OpenVLA-OFT becomes less responsive to instruction changes at intermediate execution states than at the initial state (Appendix~\ref{app:attention_persistence}). We also find that visual cues associated with the current gripper--object configuration play a particularly important role in sustaining task-like behavior, even when much of the surrounding scene information is removed (Appendix~\ref{app:robot_state_visual_prior}).}

\definecolor{winebottle}{HTML}{2563EB}
\definecolor{bowl}{HTML}{00A087}
\definecolor{creamcheese}{HTML}{E69F00}
\definecolor{stove}{HTML}{A855F7}

\begin{figure}[t]
  \centering
  \includegraphics[
    width=\textwidth,
    trim={11bp 187bp 172bp 34bp},
    clip
  ]{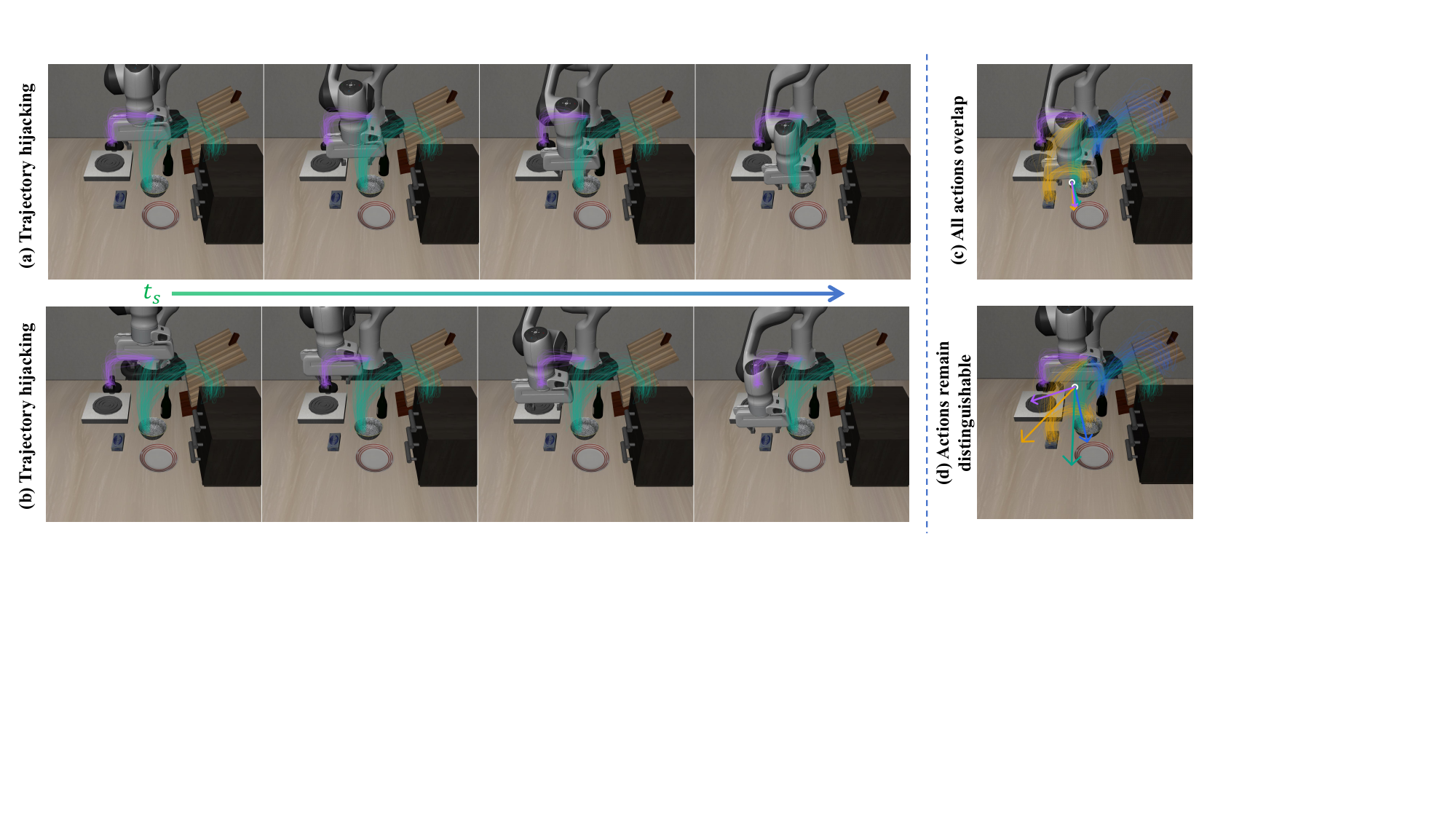}
    \caption{
    \textbf{Visualization of trajectory hijacking and instruction-conditioned action contrast.}
    Colors denote four tasks:
    \textcolor{winebottle}{\textit{put the wine bottle on top of the cabinet}},
    \textcolor{bowl}{\textit{put the bowl on top of the cabinet}},
    \textcolor{creamcheese}{\textit{put the cream cheese in the bowl}}, and
    \textcolor{stove}{\textit{turn on the stove}}.
    Colored curves show demonstration trajectories, and same-colored arrows show the corresponding 3D end-effector translational actions projected onto the 2D camera plane.
    Panels (a,b) show two forms of trajectory hijacking in bowl-to-stove switching rollouts, with the first frame at $t_s$ and subsequent frames after the switch. In (a), the policy completes the non-target bowl task; in (b), it partially drifts toward the bowl trajectory.
    Panels (c,d) show instruction-conditioned actions predicted from the same observation under the four task instructions: they nearly overlap in (c) but remain clearly distinguishable in (d).
    }
  \label{fig:obs3}
\end{figure}

% \begin{figure}[t]
%   \centering
%   \includegraphics[
%     width=\textwidth,
%     trim=150 170 150 100,
%     clip
%   ]{figures/fig3_arrow.pdf}
%     \caption{
%     \textbf{Visualization of trajectory hijacking and instruction-conditioned action contrast.}
%     After the switch, the target-conditioned action may fail to redirect the robot sufficiently toward the target task, and the behavior can remain fully or partially drawn toward a non-target trajectory, resulting in complete execution of a non-target task in (a) or partial drift toward a non-target trajectory in (b). For the same observation, a strong trajectory prior can cause actions conditioned on different instructions to nearly overlap in (c); however, in many states, their relative differences remain informative and the action directions remain distinguishable, as in (d). Colored curves denote task-specific demonstration trajectories; same-colored arrows denote the corresponding 3D end-effector translational actions projected onto the 2D camera plane: \textcolor{winebottle}{\textit{put the wine bottle on top of the cabinet}}, \textcolor{bowl}{\textit{put the bowl on top of the cabinet}}, \textcolor{creamcheese}{\textit{put the cream cheese in the bowl}}, and \textcolor{stove}{\textit{turn on the stove}}.
%     }
%   \label{fig:obs3}
% \end{figure}

\paragraph{Observation 3: Non-target trajectory priors hijack post-switch behavior.}
A key failure mode in mid-execution instruction switching is insufficient redirection toward the target task.
Consider the switch shown in Fig.~\ref{fig:obs3}~(a,b), from grasping the bowl to turning on the stove.
By the switching step, the gripper has already moved close to the bowl under the source instruction.
After the instruction changes, the target-conditioned action may fail to redirect the robot sufficiently toward the stove, and the subsequent behavior remains fully or partially drawn toward the bowl-grasping trajectory, as if hijacked by this trajectory.
More generally, the interfering trajectory need not correspond to the source task: another non-target trajectory can similarly bias the post-switch behavior away from the target.
We refer to this failure mode as \emph{trajectory hijacking}, encompassing both complete execution of a non-target task and partial drift toward a non-target trajectory.
We observe non-target task completion across multiple policies and switching ranges, as detailed in Appendix~\ref{app:wrong_task}.

% before rectified
% Observation~2 suggests that misleading visual priors play an important role in trajectory hijacking.
% For example, in standard demonstrations seen during training, states in which the gripper is already close to an object are typically followed by actions that continue toward grasping that object, whereas counterfactual examples are scarce.
% We refer to the tendency induced by such visual priors to continue the task behavior suggested by the current state as a \emph{trajectory prior}.

% after rectified
Observation~2 suggests that misleading visual priors play an important role in trajectory hijacking.
For example, in standard demonstrations, states in which the gripper is already close to an object are typically followed by actions that continue toward grasping that object, whereas counterfactual examples are scarce, training the policy to follow visually suggested behavior.
We refer to this learned tendency to follow the task behavior suggested by the current state as a \emph{trajectory prior}.

To further examine how trajectory priors affect action prediction, we compare actions predicted from the same observation under different task instructions.
In the extreme case shown in Fig.~\ref{fig:obs3}~(c), a strong trajectory prior causes actions conditioned on different instructions to nearly overlap.
\rev{To investigate this action collapse, we trace task-conditioned representations inside $\pi_{0.5}$ and find that task differences persist throughout the language backbone, but become much weaker in the action expert (Appendix~\ref{app:representation_analysis}).}
However, in many states, the relative differences among actions conditioned on different task instructions remain informative: different instructions still induce distinguishable action directions, as shown in Fig.~\ref{fig:obs3}~(d). This suggests that instruction-conditioned action differences can provide a useful signal for correcting trajectory-induced bias.

\section{Method}
\label{sec:method}
\begin{figure}[t]
  \centering
  \includegraphics[
    width=\textwidth,
    trim={15bp 80bp 126bp 68bp},
    clip
  ]{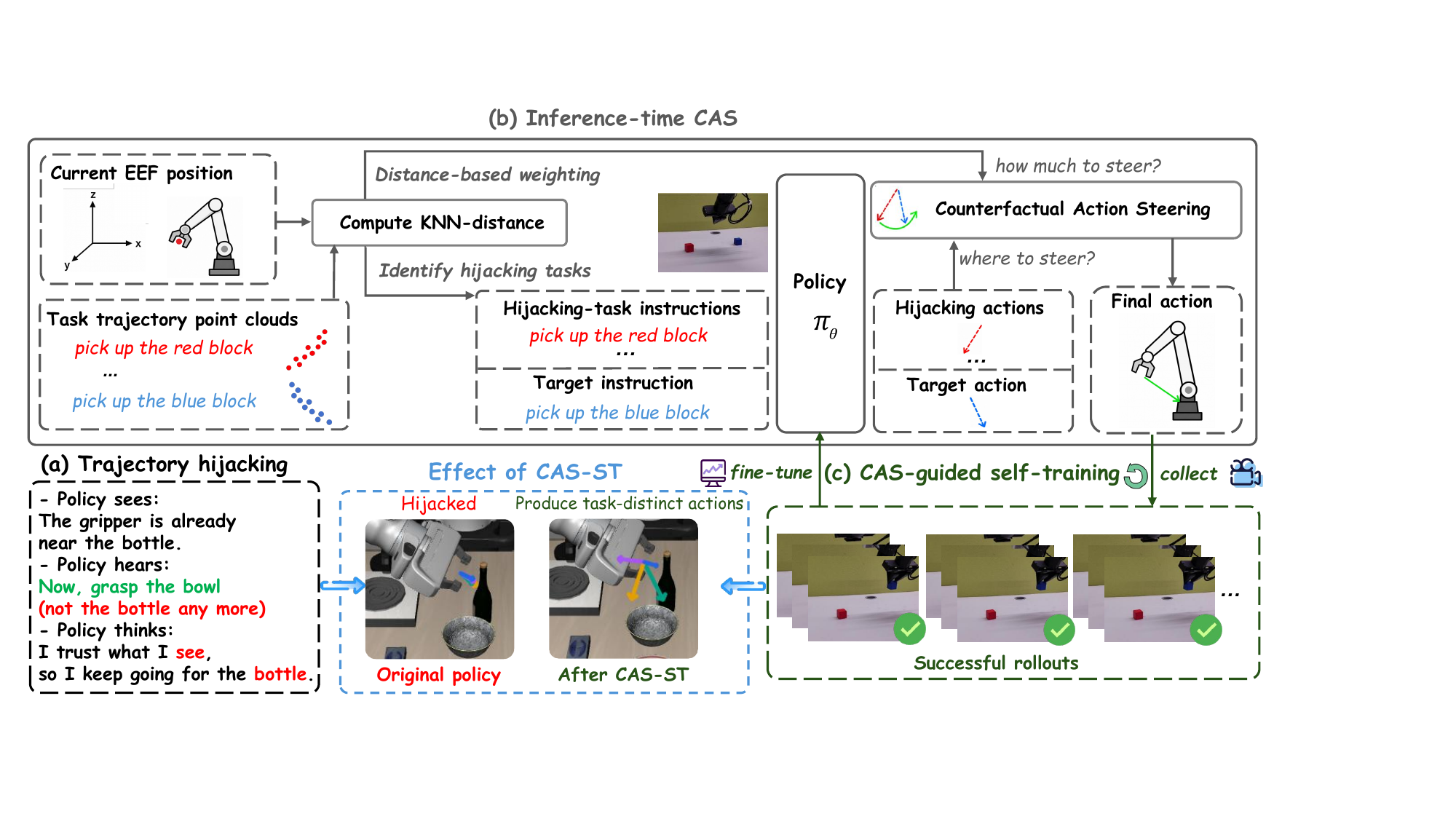}
    \caption{
      \textbf{Overview of CAS.}
      (a) Under mid-execution instruction switching, strong visual priors can cause the policy to remain drawn toward a non-target trajectory, resulting in trajectory hijacking.
      (b) Inference-time CAS identifies interfering non-target tasks, uses target--counterfactual action differences to determine where to steer, and trajectory proximity to determine how much to steer. In the red-to-blue block example, the hijacking task happens to coincide with the source task, but this need not hold in general. For clarity, the arrows show a single action from the predicted action chunk.
      (c) Successful CAS-guided rollouts can be used for self-training (CAS-ST), and this process can be applied iteratively. The blue-boxed comparison illustrates the effect of CAS-ST: given the same observation, the policy produces more task-distinct actions after self-training.
    }
  \label{fig:overview}
\end{figure}

\subsection{Overview}
Figure~\ref{fig:overview} provides an overview of our approach. CAS is a training-free, model-agnostic inference-time method for mitigating trajectory hijacking during mid-execution instruction switching. Beyond inference, successful CAS-guided rollouts can be used for self-training, providing de-hijacking supervision and enabling iterative improvement of the underlying policy.

\subsection{Inference-Time Counterfactual Action Steering}
\label{sec:inference_time_cas}

At a post-switch step $t$, let $V_t$ be the visual observation and
$p_t\in\mathbb{R}^3$ the end-effector position, and assume the scene
contains $n$ candidate tasks. The policy predicts the target-conditioned action
chunk
$A_{\mathrm{tgt}}=\pi_\theta(V_t,I_{\mathrm{tgt}})
=[a_{\mathrm{tgt}}^1,\ldots,a_{\mathrm{tgt}}^H]$,
where $H$ is the prediction horizon and $a_{\mathrm{tgt}}^k$ denotes the
$k$-th action in the chunk. CAS first identifies non-target tasks whose
trajectory priors may interfere with the current state, and then determines
where and how strongly to steer away from them.

For each task $i\in\{1,\ldots,n\}$, let $D_i$ denote its task-specific
trajectory point cloud, formed by end-effector positions from training
demonstrations. \rev{The demonstration library covers a predefined set of downstream candidate tasks (e.g., LIBERO-Goal) and need not include all tasks encountered during pretraining.}
In standard demonstrations, states in which the end effector is
already near an object are typically followed by actions that continue
interacting with that object, while counterfactual continuations are scarce.
Repeated exposure to such gripper--object configurations may bias the policy toward
continuing the corresponding task behavior. Motivated by this observation, we use
the average $K$-nearest-neighbor distance as an intuitive and lightweight geometric
proxy for identifying potentially interfering trajectory priors.
\begin{equation}
    d_i(p_t)
    =
    \frac{1}{K}
    \sum_{p\,\in\,\mathrm{KNN}(p_t,\,D_i)}
    \|p_t - p\|_2 .
\end{equation}
A smaller $d_i(p_t)$ means that $p_t$ is close to many demonstrated states of
    task $i$, indicating stronger local support for continuing that task. 
    CAS retains non-target tasks with $d_i(p_t)<\delta$ as hijacking tasks, where $\delta$ is a hyperparameter threshold, denoted by $\mathcal{H}_t=\{h_1,\ldots,h_m\}$.

For each $h_j\in\mathcal{H}_t$, CAS queries the same policy under the same
observation with the corresponding hijacking instruction $I_{h_j}$, obtaining
the counterfactual action chunk
$A_{h_j}=\pi_\theta(V_t,I_{h_j})
=[a_{h_j}^1,\ldots,a_{h_j}^H]$.
At horizon index $k$, the contrast
$a_{\mathrm{tgt}}^k-a_{h_j}^k$ determines \emph{where to steer}: it provides a
direction away from the trajectory prior of $h_j$ relative to the target
behavior. The trajectory distance determines \emph{how much to steer}: CAS weights this
direction by $\lambda/d_{h_j}(p_t)$, giving stronger corrections to hijacking
tasks with stronger trajectory priors. The corrected action at horizon
index $k$ is
\begin{equation}
    a_{\mathrm{final}}^k
    =
    a_{\mathrm{tgt}}^k
    +
    \lambda
    \sum_{h_j\in\mathcal{H}_t}
    \frac{1}{d_{h_j}(p_t)}
    \bigl(a_{\mathrm{tgt}}^k-a_{h_j}^k\bigr),
    \qquad k=1,\ldots,H,
    \label{eq:cas_final}
\end{equation}
where $\lambda>0$ controls the overall steering strength. CAS applies this
correction to all action dimensions except the gripper, whose command is taken
from the target-conditioned action. The corrected action chunk steers execution
away from interfering trajectory priors and toward the switched target.
\rev{Optional safeguards against excessive corrections, such as capping each per-task weight and bounding the corrected action, are described in Appendix~\ref{app:cas_details} alongside additional implementation details and qualitative examples.}

As a lightweight variant, \textit{Source-only CAS} uses only the source instruction as the counterfactual and applies a fixed steering weight, requiring neither trajectory
  data nor additional candidate-task instructions while retaining the core counterfactual steering mechanism of CAS. Although less effective than full CAS, it still
  noticeably improves switching success, as shown in Sec.~\ref{sec:ablations}.

\subsection{CAS-Guided Self-Training}
Beyond inference-time steering, CAS can also improve the underlying policy through self-training. We collect successful CAS-guided rollouts and finetune the original policy on their post-switch segments under the updated instruction. These trajectories provide explicit de-hijacking supervision, helping the policy learn to escape interfering trajectory priors and generalize recovery to later switching states unseen during self-training. Additional details are provided in Appendix~\ref{app:self_training_details}.
\section{Experiments}
\label{sec:experiments}
\subsection{Benchmark Construction}
\label{sec:benchmark_construction}

We construct two benchmarks for evaluating mid-execution instruction switching: LIBERO-Goal-Switch, built on LIBERO-Goal~\citep{liu2023libero}, and CALVIN-Custom-Switch, built on the CALVIN simulator~\citep{mees2022calvin}.
For \emph{LIBERO-Goal-Switch}, we evaluate switching at
$t_s \in \{10, 15, 20, 25\}$, covering progressively later stages of source-task execution.
For \emph{CALVIN-Custom-Switch}, episodes are grouped into three switching ranges,
$t_s \in [9, 15]$, $[15, 21]$, and $[21, 27]$, corresponding to progressively later
progress along the source-task trajectory.
Further benchmark construction and evaluation details are provided in
Appendix~\ref{app:benchmark_construction_details}.

\subsection{CAS Mitigates Trajectory Hijacking at Inference Time}
\paragraph{Simulation evaluation.}
\input{TABLE_MAIN_RESULTS}
We evaluate CAS on \emph{LIBERO-Goal-Switch} and \emph{CALVIN-Custom-Switch}
using OpenVLA-OFT, $\pi_0$, $\pi_{0.5}$, Fast-WAM, and LingBot-VA.
As a training-free baseline, we compare against Policy Contrastive Decoding (PCD)
\citep{wu2025policy}, which mitigates spurious correlations inherited from
pre-training trajectories through inference-time contrastive decoding. \rev{Additional comparisons with CAG-TF, AVA-VLA, OpenVLA-OFT+ from LIBERO-Plus, and ReSteer are provided in Appendix~\ref{app:related_method_comparisons}.}
We omit PCD for OpenVLA-OFT because PCD is designed for autoregressive and
diffusion-based policies and is not directly applicable to the mainstream
OpenVLA-OFT implementation.
As shown in Table~\ref{tab:main_results}, CAS consistently improves over the
corresponding base policies across both benchmarks, all five evaluated policies,
and all switching settings, while also outperforming PCD in every setting where
PCD is applicable.
The gains are particularly pronounced on \emph{CALVIN-Custom-Switch}.
For example, in the $[9,15]$ switching range, CAS improves the success rate
of $\pi_0$ from 22.1\% to 73.8\% and that of $\pi_{0.5}$ from 23.3\% to
77.1\%, corresponding to gains of 51.7 and 53.8 percentage points, respectively.
We further analyze the inference-time efficiency of CAS by comparing the base
policy, PCD, and CAS using $\pi_0$ on \emph{CALVIN-Custom-Switch} in the
$[15,21]$ switching range, with all methods evaluated under the same hardware
and evaluation setup.
Despite using a serial implementation of counterfactual policy queries without
batching, CAS incurs only moderate inference overhead.
Its average inference latency per replan is $1.54\times$ that of the base policy,
whereas PCD requires $7.71\times$ the base latency.
Interestingly, CAS also achieves a lower average episode time than the base
policy (12.13\,s vs.\ 14.14\,s), as its substantially higher switching success
allows more episodes to complete the target task before reaching the maximum
episode horizon. Details are provided in Appendix~\ref{app:efficiency}.

We further evaluate CAS under diverse settings. CAS consistently improves over base policies across varying trajectory concentrations and fully randomized object placements (Appendix~\ref{app:diverse_trajectory_regimes}). It also remains effective under increasing training--evaluation appearance gaps (Appendix~\ref{app:scene_appearance_shift}) and consistently outperforms the base policy across inference random seeds (Appendix~\ref{app:random_seed})\rev{ and across different initial states (Appendix~\ref{app:initial_states})}.
\paragraph{Real-world evaluation.}
\begin{figure}[t]
  \centering
  \includegraphics[width=\textwidth]{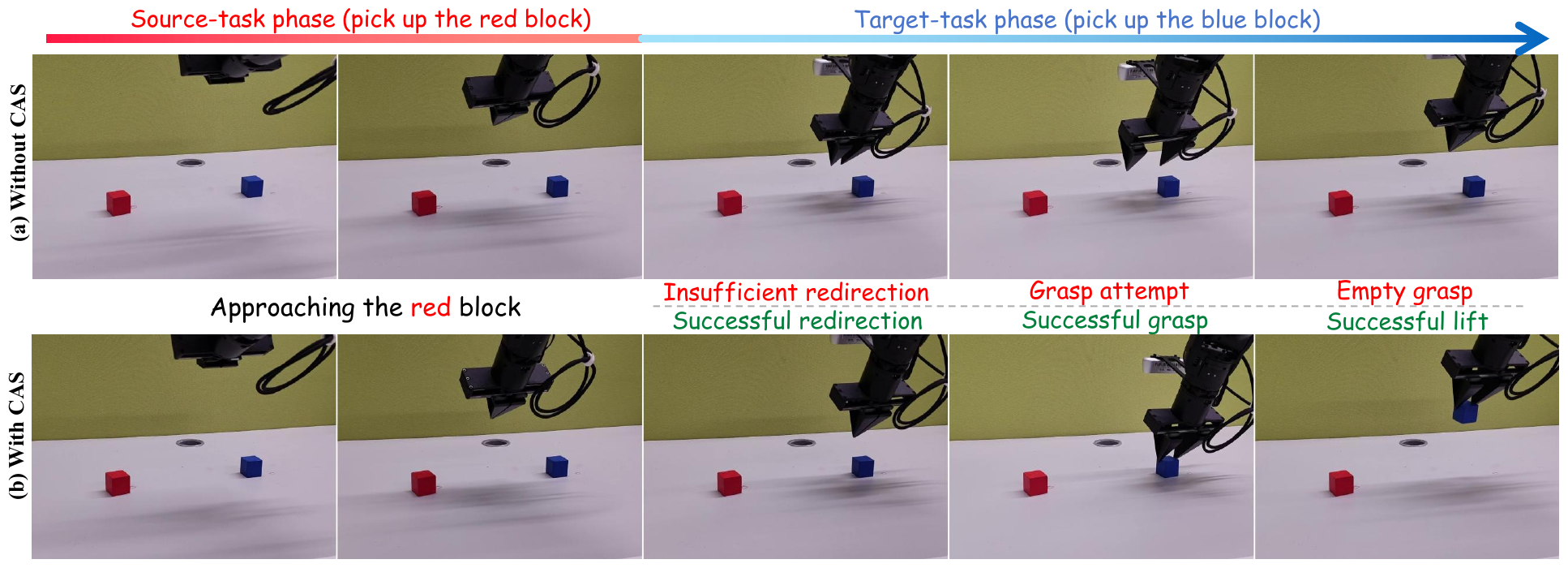}
    \caption{
    \textbf{Real-world comparison of mid-execution instruction switching.}
    The instruction switches from \textcolor{red}{\textit{picking up the red block}} to \textcolor{blue}{\textit{picking up the blue block}}. Without CAS, insufficient redirection leads to an empty grasp; with CAS, the robot successfully picks up the blue block.
    }
  \label{fig:real}
\end{figure}
% \begin{figure}[t]
%   \centering
%   \includegraphics[
%     width=\textwidth,
%     trim=0 120 0 80,
%     clip
%   ]{figures/1915real_300dpi.png}
%   \caption{
%   \textbf{Real-world comparison of mid-execution instruction switching.}
% The robot is initially instructed to pick up the red block, and the instruction is then switched to pick up the blue block.
% Without CAS, the policy remains biased by the red-block trajectory prior and fails to redirect sufficiently: the gripper closes on the empty space to the left of the blue block and then lifts without grasping it.
% With CAS, the robot is steered away from the interfering prior, redirected onto the blue block, and successfully picks it up.
% }
%   \label{fig:real}
% \end{figure}

We evaluate CAS on an AIRBOT Play robot arm. The robot first executes an instruction to pick up one block (red or blue), and the instruction is then switched to the other at step $t_s$. We test both switching directions at $t_s=20$ and $t_s=30$, with 20 trials per setting, using matched object placements for the base and CAS trials. As shown in Table~\ref{tab:real_world}, CAS improves $\pi_0$ in all settings. The real-world policy predicts delta joint actions rather than the delta end-effector actions used in simulation, suggesting that CAS is not tied to a specific action representation. Fig.~\ref{fig:real} shows a representative switch from picking up the red block to picking up the blue block. Without CAS, the policy remains biased toward the red-block trajectory prior and fails to redirect sufficiently; with CAS, it escapes the interfering prior and successfully picks up the blue block.

\subsection{Beyond Inference: CAS Produces Higher-Quality Rollouts for Self-Training}
\label{sec:ST}
\input{TABLE_SELF_TRAINING}

We finetune the original policy on successful CAS-guided rollouts, denoted as CAS-ST, and compare it with the base policy and vanilla self-training (Vanilla-ST), which finetunes the same base policy on the same number of successful rollouts collected without CAS. All finetuned policies are evaluated without CAS at inference time.
We evaluate direct self-training gains at the data-collection setting and generalization at unseen later switches, where greater progress along the source-task trajectory makes redirection more challenging. For LIBERO-Goal-Switch, self-training trajectories are collected at $t_s=25$; we evaluate at $t_s=25$ and further test generalization at the unseen later switch $t_s=30$. For CALVIN-Custom-Switch, trajectories are collected in the $t_s \in [9,15]$ switching range and evaluated in both the same range and the unseen later range $t_s \in [15,21]$.

As shown in Table~\ref{tab:ST_main}, Vanilla-ST improves over the base policy, while CAS-ST consistently yields further gains across both benchmarks and both evaluated policies, both at the data-collection settings and at unseen later switches. These results show that successful CAS-guided rollouts provide higher-quality supervision for self-training than vanilla successful rollouts with the same number of trajectories. Qualitative comparisons of paired successful Base and CAS rollouts in Appendix~\ref{app:better_ST} further show that CAS can produce more effective de-hijacking behavior even when both methods ultimately complete the target task, highlighting differences in trajectory quality that are not captured by binary success alone. The consistent gains of CAS-ST at unseen later switches further indicate that the recovery behaviors learned from CAS-guided trajectories extrapolate beyond the switching regimes used for self-training. 

Motivated by this extrapolation, we further investigate whether CAS-guided self-training can be applied iteratively to progressively improve recovery at later switches. Starting from the one-round $\pi_{0.5}$ CAS-ST checkpoint on LIBERO-Goal-Switch, whose self-training trajectories were collected at $t_s=25$, we collect additional successful CAS-guided trajectories at $t_s=40$ and perform a second round of self-training. As shown in Table~\ref{tab:round2_cas_st}, the one-round checkpoint already outperforms the base policy at $t_s = 40, 50, 60$, well beyond its data-collection setting. The second round further improves performance at all three switching steps. Notably, although the second-round data are collected only at $t_s=40$, the gains persist at the unseen later switches $t_s=50$ and $t_s=60$. These results show that CAS-guided self-training can be iterated to progressively improve recovery under increasingly challenging switching conditions. To further analyze the effect of self-training on instruction-conditioned action predictions, we compare instruction-conditioned actions before and after two rounds of CAS-ST, as in Fig.~\ref{fig:obs3}~(d). After self-training, the policy produces more distinguishable actions that are better aligned with the corresponding task directions, even in challenging states where the base policy predicts nearly overlapping actions; see Appendix~\ref{app:action_vis}.

We provide additional analyses of CAS-guided self-training in the appendix. 
CAS-ST is substantially more finetuning-efficient than Vanilla-ST (Appendix~\ref{app:self_training_budget}). Pair-level analysis identifies cases where CAS-ST outperforms Vanilla-ST despite fewer retained self-training trajectories (Appendix~\ref{app:self_training_pairwise}). CAS-ST's advantage also persists at a later unseen switch (Appendix~\ref{app:self_training_later_switches}) and under fully randomized trajectory distributions (Appendix~\ref{app:calvin_random_self_training}).
\rev{A paired self-training experiment further shows that CAS-ST's advantage persists when switching states, source--target task pairs, and trajectory counts per pair are matched, despite using fewer dataset frames (Appendix~\ref{app:matched_self_training}).}
\rev{We also compare action statistics and trajectory lengths in the original and paired self-training datasets (Appendix~\ref{app:st_action_analysis}).}
% -------------------------------------------------------------------------------------------
% detail version
% We conduct additional experiments in the appendix. In Appendix~\ref{app:self_training_budget}, we study finetuning efficiency and find that, on CALVIN-Custom-Switch with Fast-WAM, CAS-ST trained for only 4k steps already outperforms Vanilla-ST trained for 20k steps by more than 20 percentage points. In Appendix~\ref{app:self_training_pairwise}, we further analyze the CALVIN-Custom-Switch results at the source--target pair level. CAS-ST achieves higher success on two task pairs despite using fewer retained self-training trajectories, while among all seven pairs for which CAS-ST uses more trajectories, six improve and one ties, with no observed regression. In Appendix~\ref{app:self_training_later_switches}, we further extend the one-round CAS-ST versus Vanilla-ST comparison on LIBERO-Goal-Switch to $t_s=35$, where CAS-ST continues to outperform Vanilla-ST for both $\pi_{0.5}$ and Fast-WAM. In Appendix~\ref{app:calvin_random_self_training}, we evaluate self-training when trajectory distributions are fully randomized in both training and evaluation, where CAS-ST remains substantially more effective than Vanilla-ST.

\begin{table}[!htbp]
\centering
\noindent
\begin{minipage}[t]{0.40\linewidth}
\vspace{0pt}
\centering
\captionof{table}{Real-world switching.}
\label{tab:real_world}

\small
\setlength{\tabcolsep}{3.5pt}
\begin{tabular}{@{}lccc@{}}
\toprule
Policy & $t_s$ & Blue$\rightarrow$Red & Red$\rightarrow$Blue \\
\midrule
Base     & 20 & 7/20  & 7/20 \\
+ CAS & 20 & \textbf{13/20} & \textbf{15/20} \\
\midrule
Base     & 30 & 2/20  & 6/20 \\
+ CAS & 30 & \textbf{10/20} & \textbf{14/20} \\
\bottomrule
\end{tabular}
\end{minipage}%
% \hspace{0.025\linewidth}%
\hfill
\begin{minipage}[t]{0.575\linewidth}
\vspace{0pt}
\centering
\captionof{table}{
Iterative CAS-guided self-training on LIBERO-Goal-Switch with $\pi_{0.5}$.
}
\label{tab:round2_cas_st}

\small
\setlength{\tabcolsep}{4.0pt}
\begin{tabular}{@{}lccc@{}}
\toprule
Method & $t_s=40$ & $t_s=50$ & $t_s=60$ \\
\midrule
Base
    & 28.9 & 24.4 & 17.8 \\
CAS-ST (1 round)
    & 36.7 & 32.2 & 28.9 \\
CAS-ST (2 rounds)
    & \textbf{46.7} & \textbf{42.2} & \textbf{37.8} \\
\bottomrule
\end{tabular}
\end{minipage}

\end{table}

\begin{table}[!htbp]
\centering
    \caption{
    \textbf{CAS variants and design ablation.}
    We compare full CAS, its lightweight Source-only CAS variant, and All non-target Fixed.
    Source-only CAS requires no trajectory data and achieves $35.4\%$ Switch SR.
    Full CAS achieves the highest success rate of $45.4\%$ with state-dependent task selection and adaptive steering across relevant hijacking tasks.
    }

\label{tab:cas_design_ablation}

\small
\setlength{\tabcolsep}{3.0pt}
\renewcommand{\arraystretch}{1.12}

\resizebox{\linewidth}{!}{
\begin{tabular}{@{}lcccc@{}}
\toprule
Method &
Counterfactual task set &
Required task trajectory data &
Steering weight &
Switch SR $\uparrow$ \\
\midrule

Base
& --
& None
& --
& 6.7 \\

Source-only CAS
& Source task
& None
& $w=0.6$
& 35.4 \\

All non-target Fixed
& All non-target tasks
& None
& $w=0.4$
& 23.3 \\

\textbf{CAS (Full)}
& KNN-selected non-target tasks
& All candidate tasks
& Adaptive
& \textbf{45.4} \\

\bottomrule
\end{tabular}
}
\end{table}

\subsection{Ablation Study}
\label{sec:ablations}
We conduct all ablation studies using $\pi_0$ on CALVIN-Custom-Switch in the switching range $t_s \in [15,21]$, covering both design and hyperparameter ablations\rev{, and additionally evaluate alternative action steering strategies on LIBERO-Goal-Switch at $t_s=20$}.

\paragraph{Design ablations.}
Table~\ref{tab:cas_design_ablation} compares full CAS with its lightweight Source-only CAS variant and All non-target Fixed, which uniformly steers against all non-target tasks.
To optimize each variant's performance, we sweep the fixed steering weight over
$w \in \{0.1, 0.2, 0.4, 0.6, 0.8\}$
for both fixed-weight variants and report the best result for each variant.
For full CAS, we use the same $\lambda=0.03$ as in the main results (Table~\ref{tab:main_results}).
The complete sweep is provided in Appendix~\ref{app:design}.
Both fixed-weight variants outperform the base policy under every tested steering weight. Source-only CAS improves Switch SR from $6.7\%$ to $35.4\%$, demonstrating that the core counterfactual steering mechanism remains effective even without trajectory-based task selection and weighting. All non-target Fixed can over-steer against weakly interfering trajectories, yielding $23.3\%$ Switch SR.
Full CAS adaptively weights the identified hijacking tasks and achieves the highest Switch SR of $45.4\%$.

\rev{Beyond task selection and adaptive weighting, we compare alternative action correction strategies in Appendix~\ref{app:action_steering} (Table~\ref{tab:alternative_action_steering}). Simply scaling the target-conditioned action or reversing the CAS correction degrades performance on both benchmarks. CAS also remains effective when its final action norm is matched to the target-conditioned action, supporting the importance of correction direction beyond action amplification.}

\paragraph{Hyperparameter ablations.}
We vary $K$, $\delta$, and $\lambda$ and find that CAS consistently outperforms the base policy across all tested settings. \rev{CAS also clearly improves over the base policy across a broad range of $\lambda$ in all three switching ranges.} Detailed results are provided in Appendix~\ref{app:hyper}.

\section{Conclusion}
\label{sec:conclusion}

We studied mid-execution instruction switching across VLA and WAM policies and identified trajectory hijacking as a key failure mode. We proposed CAS, a training-free, model-agnostic inference-time method that steers the policy away from interfering trajectory priors using target--counterfactual action differences. Across two switching benchmarks and real-robot evaluation, CAS consistently improves switching success. CAS-guided rollouts also provide better self-training supervision, yielding stronger policies that generalize to unseen later switches.
\newpage
\section*{AI Use Statement}
In this work, we did not use generative AI tools for any tasks with required disclosure. Additionally, we used generative AI tools to edit the research paper to improve readability, including language polishing, grammar correction, and refinement of wording. We have reviewed all AI-assisted edits to ensure that they did not alter the scientific meaning, claims, results, or conclusions of the work. We take responsibility for the final content of this work, including text, claims, or artifacts produced with the aid of generative AI.

\section*{Reproducibility Statement}
We provide detailed information to facilitate reproduction of our results.
The formulation and implementation details of CAS are described in
Sec.~\ref{sec:method} and Appendix~\ref{app:cas_details}. The construction of
LIBERO-Goal-Switch and CALVIN-Custom-Switch, together with their evaluation
protocols, is documented in Appendix~\ref{app:benchmark_construction_details}
and Appendix~\ref{app:obs1_details}. Additional experimental details include
real-robot settings (Appendix~\ref{real_world_exp_details}), robustness across
random seeds and CAS hyperparameter analyses (Appendices~\ref{app:random_seed}
and~\ref{app:abla}), self-training data collection, subsampling, finetuning,
and evaluation configurations (Appendix~\ref{app:self_training_details}), and
implementation details of the PCD baseline
(Appendix~\ref{app:pcd_implementation_details}). Code and the CAS-guided self-training checkpoints underlying our main results will be made publicly available.

\bibliographystyle{iclr2027_conference}
\bibliography{references}
\clearpage
\appendix
\begin{center}
    {\Large\bfseries Appendix}
\end{center}
\vspace{1em}

\startcontents[appendix]
\printcontents[appendix]{ }{1}{%
    \section*{Contents}%
    \setcounter{tocdepth}{2}%
}
\clearpage
\section{Related Work}
\label{app:related_work}
\paragraph{Generalist instruction-conditioned robot policies.}
Instruction-conditioned robot policies have emerged as a promising paradigm for general-purpose robot manipulation, enabling a single policy to perform diverse tasks from visual observations and natural-language instructions. A prominent line of work develops Vision-Language-Action (VLA) policies that directly connect multimodal perception and language understanding to action generation. $\pi_0$ formulates robot control as a vision-language-action flow model~\citep{BlackK-RSS-25}, while $\pi_{0.5}$ extends this framework toward open-world generalization through co-training on heterogeneous robot data, high-level semantic prediction, web data, and hybrid multimodal examples~\citep{black2025pi05}. OpenVLA-OFT focuses on efficient adaptation of pretrained VLA models to downstream manipulation tasks through improved action decoding, action representations, and finetuning objectives~\citep{kim2025fine}. 
VLAct further explores representation-centric continued pre-training of VLM backbones for improved transfer across action heads, tasks, and robot embodiments~\citep{yang2026beyond}, while Mage-VL investigates codec-native visual encoding for efficient multimodal perception~\citep{yang2026mage}. 
More recently, World-Action Models (WAMs) have explored predictive world modeling for robot control. LingBot-VA studies causal world modeling for action generation~\citep{li2026causal}, while Fast-WAM investigates whether explicit future imagination is necessary at test time and develops a more efficient WAM formulation for action prediction~\citep{yuan2026fast}. Together, these VLA and WAM approaches represent a diverse family of generalist instruction-conditioned robot policies.

Alongside model development, benchmarks and evaluation frameworks such as LIBERO, simulation-based evaluation of real-world robot manipulation policies, VLABench, and RoboTwin 2.0 provide broad testbeds for language-conditioned manipulation, knowledge transfer, long-horizon reasoning, and robust bimanual control~\citep{liu2023libero,li2024evaluating,zhang2025vlabench,chen2025robotwin}. However, standard evaluations typically assume that the task instruction remains fixed throughout an episode. They therefore measure whether a policy can execute a specified task from an initial state, but do not directly test whether it can redirect its behavior from an intermediate state after the instruction changes. We study this setting under \emph{mid-execution instruction switching} and show that strong policies across both VLA and WAM families can suffer substantial performance degradation, with trajectory hijacking emerging as a key failure mode. These observations motivate approaches that directly address switching failures after execution has already begun.

\paragraph{Instruction switching in instruction-conditioned robot policies.}
A small but growing body of work has begun to study whether robot policies can follow newly issued instructions from intermediate execution states. Existing approaches have primarily focused on VLA policies, while mid-execution instruction switching in WAMs remains comparatively underexplored.

ReSteer~\citep{chen2026resteer} studies a closely related problem under the notion of policy steerability, asking whether a multitask robot policy can respond to newly issued task instructions from intermediate execution states. In addition to an exhaustive rollout-based evaluation protocol, ReSteer introduces conditional mutual information (CMI) between language and action as a proxy for identifying states in which the policy is weakly conditioned on the instruction. Low-CMI states are then prioritized for additional steering supervision. To improve steerability, ReSteer develops a multi-stage data-centric refinement pipeline. It first segments manipulation demonstrations into semantic stages, such as pre-grasp, transport, and place, using kinematic and contact cues. For a selected low-steerability source state, it searches for a stage-compatible state from a target-task demonstration and synthesizes a cross-task transition between the two, followed by execution under the target instruction. The resulting steering trajectories are added to the training data for policy finetuning. ReSteer further applies self-refining behavioral cloning (SRBC), collecting successful task-switching rollouts from the updated policy and using them for further refinement. Overall, ReSteer improves steerability through a relatively elaborate workflow that combines state-wise steerability estimation, stage-aware trajectory processing, cross-task transition synthesis, policy finetuning, and self-refinement, resulting in considerable methodological and engineering complexity.

This work establishes the importance of execution-time instruction switching, but addresses switching failures through additional data generation and training. In contrast, the core CAS method targets instruction-conditioned robot policies broadly, including both VLA and WAM policies, and operates as a training-free, model-agnostic inference-time intervention. Rather than learning or synthesizing a dedicated transition mechanism, CAS queries the same policy under the target instruction and counterfactual hijacking-task instructions, and uses their action differences to steer execution away from interfering trajectory priors. Successful CAS-guided rollouts can further be reused for self-training, providing more effective de-hijacking supervision than vanilla successful rollouts.

\section{CAS Implementation Details and Illustrative Examples}
\label{app:cas_details}
\subsection{CAS Implementation Details}
\label{app:cas_implementation_details}

We describe several practical details for implementing CAS during inference.
The core update follows Eq.~\eqref{eq:cas_final}, while lightweight safeguards
such as weight capping or action clipping can be used to improve stability when
the counterfactual correction becomes large. We also discuss
gripper-dimension exclusion and batched computation of
instruction-conditioned action chunks.

\paragraph{Capping per-task steering weights.}
For each hijacking task $h_j\in\mathcal{H}_t$, the scalar weight associated
with its counterfactual correction term in Eq.~\eqref{eq:cas_final} is
\begin{equation}
    r_{h_j}=\frac{\lambda}{d_{h_j}(p_t)} .
\end{equation}
When the end-effector is very close to the demonstrated trajectory of task
$h_j$, the distance $d_{h_j}(p_t)$ can become small, resulting in an
excessively large steering weight. In this case, at horizon index $k$, the
corresponding correction term,
$r_{h_j}\bigl(a_{\mathrm{tgt}}^k-a_{h_j}^k\bigr)$,
may dominate the aggregated correction. To avoid such over-amplification, we
cap each per-task steering weight by a maximum value $r_{\max}$:
\begin{equation}
    \tilde{r}_{h_j}
    =
    \min\left(
    \frac{\lambda}{d_{h_j}(p_t)},\,
    r_{\max}
    \right).
\end{equation}
This cap is applied independently to each hijacking task before the correction
terms are aggregated. When capping is used, $\tilde{r}_{h_j}$ replaces
$\lambda/d_{h_j}(p_t)$ in Eq.~\eqref{eq:cas_final}.

\paragraph{Bounding the corrected action.}
After aggregating the counterfactual correction terms, a corrected action
$a_{\mathrm{final}}^k$ may occasionally have a larger magnitude than desired,
especially when multiple hijacking tasks are selected or when their KNN
distances are small. In such cases, bounding the corrected action can improve
execution stability. Bounding is applied only to the non-gripper dimensions.
A simple choice is per-dimension clipping, which clamps
each steered dimension to the valid action range of the policy,
such as the normalized range $[-1,1]$. Another possible choice is norm
rescaling, which scales the non-gripper components of each corrected action together so that their
magnitude remains within a fixed factor of the corresponding
target-conditioned action $a_{\mathrm{tgt}}^k$ restricted to the same dimensions, while preserving the
direction of the steered non-gripper action vector. The per-task cap $r_{\max}$ limits the
contribution of each individual hijacking task before summation, whereas this
final bounding step constrains the overall magnitude of
$a_{\mathrm{final}}^k$ after all correction terms have been aggregated.

\rev{\paragraph{Settings used in our experiments.}
All simulation experiments use the per-task cap with $r_{\max}=0.6$. Post-CAS per-dimension clipping to $[-1,1]$ is applied for OpenVLA-OFT, Fast-WAM, and LingBot-VA on CALVIN-Custom-Switch. For $\pi_0$ and $\pi_{0.5}$ on CALVIN-Custom-Switch, the policy outputs are clipped to $[-1,1]$ before CAS, and no additional post-CAS bounding is applied. On LIBERO-Goal-Switch, no additional post-CAS bounding is applied. CAS remains effective both with and without post-CAS bounding (Table~\ref{tab:main_results}), so this step serves as an optional safeguard rather than a requirement.

\paragraph{Low-level controller constraints.}
The corrected action is executed through the native low-level controller of each environment, which provides additional protection. In LIBERO, the OSC controller clips the 6-D input to $[-1,1]$ and maps it to bounded end-effector deltas, and the resulting joint torques are clipped to the actuator limits. In CALVIN, the end-effector target is converted by inverse kinematics into joint targets that are clipped to the joint limits, and these targets are executed by position control with velocity and force limits. Linear extrapolation in action space therefore does not produce unbounded joint commands.}

\paragraph{Excluding the gripper dimension from steering.}
CAS steering is applied to all action dimensions except the gripper. In
simulation, the steered dimensions correspond to delta end-effector
translation and rotation. On the real robot, they correspond to delta joint
commands. For each horizon index $k$, the gripper open/close command is taken
directly from the corresponding target-conditioned action
$a_{\mathrm{tgt}}^k$ and is not modified by CAS.

\paragraph{Computing multiple instruction-conditioned action chunks.}
CAS requires action-chunk predictions under multiple instructions at the same
observation, including the target instruction and the counterfactual
instructions associated with the selected hijacking tasks. A simple
implementation is to compute these chunks through multiple policy forward
passes: one forward pass for $I_{\mathrm{tgt}}$ and one additional forward
pass for each $h_j\in\mathcal{H}_t$ using instruction $I_{h_j}$. Each policy query returns an action
chunk, and CAS applies Eq.~\eqref{eq:cas_final} to corresponding horizon
indices across the target and counterfactual chunks. This implementation requires only minimal changes to an existing robot-policy inference pipeline, since the policy can be queried repeatedly with the same visual observation and different language instructions.

When inference efficiency is important, these instruction-conditioned action
chunks can instead be computed in a single batched forward pass. Specifically,
the current visual observation $V_t$ is repeated across the batch dimension,
while the language inputs are set to
$\{I_{\mathrm{tgt}}, I_{h_1}, \ldots, I_{h_m}\}$.
The policy then returns the target-conditioned action chunk and all
counterfactual action chunks together. In this batched implementation, CAS
does not introduce additional sequential policy-forward time for obtaining
multiple action chunks in principle.

\subsection{Illustrative Examples}
\label{app:visualizations}
\label{app:cas_example}
\begin{figure}[t]
    \centering
    \includegraphics[width=0.6\textwidth]{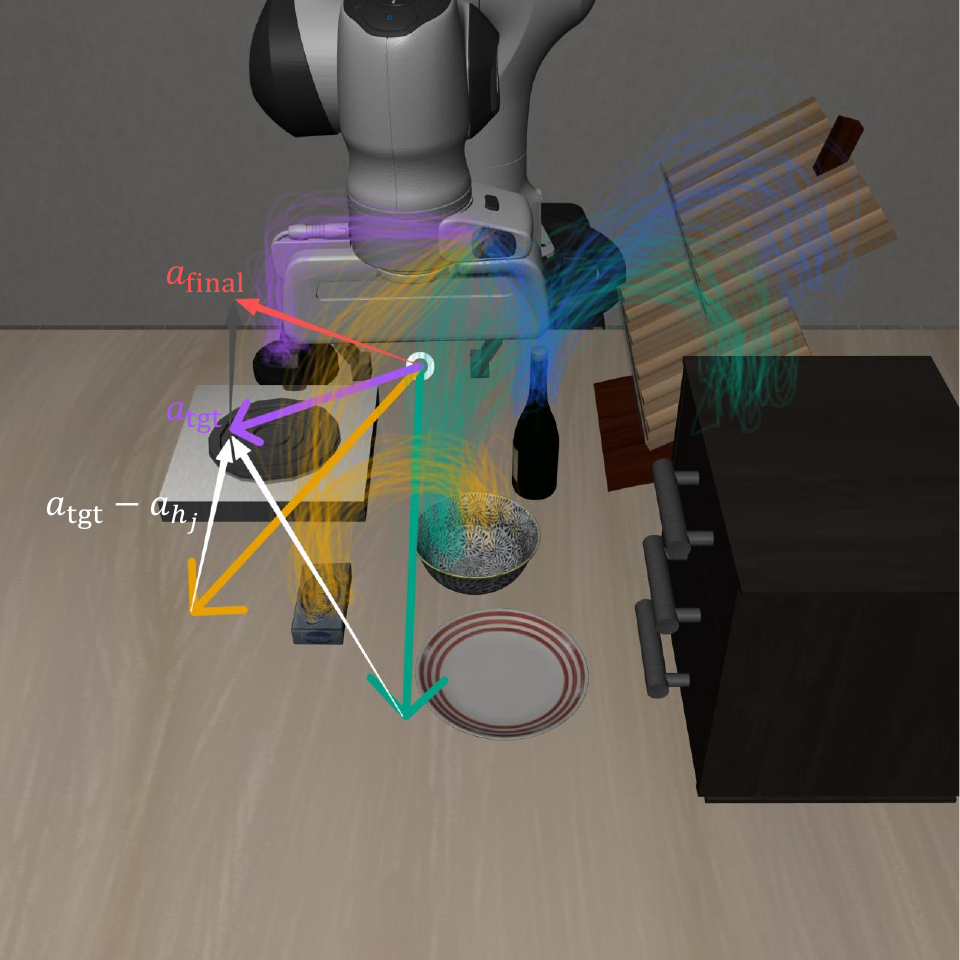}
    \caption{
    \textbf{Action-level illustration of Counterfactual Action Steering.}
    The robot switches from the source task \emph{put the bowl on top of the cabinet} to the target task
    \emph{turn on the stove}. Since the current end-effector position $p_t$ is close to the demonstration trajectory distributions of the bowl-related tasks, CAS selects \emph{put the bowl on top of the cabinet} (green) and
    \emph{put the cream cheese in the bowl} (yellow) as hijacking tasks based on their small KNN
    distances, while excluding \emph{put the wine bottle on top of the cabinet} (blue) due to its larger
    KNN distance. CAS contrasts the target-conditioned action $a_{\mathrm{tgt}}$ with the selected counterfactual
    actions $a_{h_j}$ and aggregates the weighted directions $a_{\mathrm{tgt}}-a_{h_j}$ to produce $a_{\mathrm{final}}$ (red),
    which steers the robot away from interfering trajectory priors and toward the stove knob.
    For clarity, the arrows visualize the translational components of a single action from each predicted action chunk, taken at the same horizon index.
    }
    \label{fig:cas_example}
\end{figure}
Fig.~\ref{fig:cas_example} illustrates how CAS corrects the policy action in a concrete
mid-execution instruction-switching episode. We illustrate the correction at a single horizon index shared across the predicted action chunks and omit the superscript $k$ for clarity. The robot is first commanded to \emph{put the bowl
on top of the cabinet}. After the robot has moved for several steps under this source instruction,
the user switches the target task to \emph{turn on the stove}. At the switching state, the current
end-effector position $p_t$ is already close to the bowl and to the trajectory distribution of the
source task. As a result, directly querying the policy with the switched target instruction produces
the target-conditioned action $a_{\mathrm{tgt}}$, shown by the purple arrow, but this action can still be biased
toward the bowl-related trajectory prior.

CAS uses the current end-effector position to identify which non-target tasks are likely to interfere
with the switched target behavior. In this example, two task trajectories are selected as hijacking sources:
\emph{put the bowl on top of the cabinet}, shown in green, and
\emph{put the cream cheese in the bowl}, shown in yellow. For both tasks, the corresponding
trajectory point clouds contain demonstration points close to the current end-effector position
$p_t$, resulting in small KNN distances. CAS therefore treats them as hijacking tasks and queries
the policy with their corresponding counterfactual instructions. In
contrast, the trajectory of \emph{put the wine bottle on top of the cabinet}, shown in blue, is farther
from the current end-effector position. Since its KNN distance is larger than the threshold, CAS
does not include it in the hijacking set.

For each selected hijacking task $h_j\in\mathcal{H}_t$, CAS compares the target-conditioned action with the action
predicted under the corresponding counterfactual instruction. Let $a_{h_j}$ denote this counterfactual action
at the same horizon index. The contrast $a_{\mathrm{tgt}}-a_{h_j}$ indicates a direction that moves the
policy away from that non-target trajectory prior. In Fig.~\ref{fig:cas_example}, these contrastive
directions are shown by the white arrows. The correction is weighted by the inverse KNN distance,
so the smaller the KNN distance to the demonstration trajectory of a hijacking task, the more strongly it contributes to the final
correction. Following the core update in Eq.~\eqref{eq:cas_final}, we obtain
\begin{equation}
a_{\mathrm{final}}
=
a_{\mathrm{tgt}}+
\lambda
\sum_{h_j\in\mathcal{H}_t}
\frac{1}{d_{h_j}(p_t)}
\bigl(a_{\mathrm{tgt}}-a_{h_j}\bigr),
\end{equation}
where $\mathcal{H}_t$ contains the selected hijacking tasks, $d_{h_j}(p_t)$ is the KNN distance from
$p_t$ to the trajectory point cloud $D_{h_j}$ of task $h_j$, and $\lambda>0$ controls the overall steering strength.

The resulting action $a_{\mathrm{final}}$ is shown by the red arrow. Compared with the original
target-conditioned action $a_{\mathrm{tgt}}$, the corrected action is steered away from the bowl-related and
cream-cheese-related trajectory priors and points more directly toward the stove knob. In this way,
CAS helps the policy escape the interfering trajectory prior induced by the source-side behavior and
redirects execution toward the switched target task, \emph{turn on the stove}.

Let us now return to Figure~\ref{fig:teaser}~(c) and explain the example more concretely in terms of CAS. After the instruction switches to the target task, i.e., picking up the blue block, CAS first considers which non-target task may interfere with the current state. At this moment, the end-effector has already moved along the previous red-block trajectory, so the local trajectory prior associated with the red-block task is strong. The red-block task is therefore identified as a hijacking task. CAS then queries the policy twice at the same observation: once with the target instruction ``pick up the blue block'' and once with the counterfactual hijacking instruction ``pick up the red block.'' The blue-conditioned action gives the policy's current attempt to follow the switched target, while the red-conditioned action reveals the direction induced by the interfering trajectory prior. CAS contrasts these two actions and adds the resulting correction to the target-conditioned action. In this example, the correction suppresses the red-block bias and shifts the final action toward the blue block, producing the green corrected action shown in the figure.

\section{Benchmark Construction Details}
\label{app:benchmark_construction_details}

\subsection{Complementary Online and Offline Protocols}
\label{app:benchmark_protocol}

We construct two primary simulation benchmarks, \emph{LIBERO-Goal-Switch} and
\emph{CALVIN-Custom-Switch}, using complementary protocols for evaluating
mid-execution instruction switching in VLA and WAM policies. We additionally introduce 
\emph{CALVIN-Custom-Switch-Random} as a
supplementary variant of \emph{CALVIN-Custom-Switch}, in which object
placements are broadly randomized to evaluate mid-execution instruction
switching without fixed nominal object locations.

\emph{LIBERO-Goal-Switch} adopts an \emph{online} switching protocol. The
evaluated policy first executes a source instruction and, at a designated
switching step $t_s$, receives a new target instruction within the same rollout.
The switching state is therefore produced naturally by the policy's own
preceding execution, closely reflecting realistic instruction-switching
scenarios in which a user changes the instruction while the robot is already
acting. A minor consequence of this protocol is that different policies may
make slightly different spatial progress along the source-task trajectory by
the same switching step $t_s$, due to small differences in execution speed.
This can introduce modest variations in switching difficulty across models.
Nevertheless, comparisons between methods built on the same base policy remain
well controlled, since they use the same pre-switch policy and any intervention
is activated only after the instruction changes.

\emph{CALVIN-Custom-Switch}, in contrast, adopts a controlled \emph{offline}
switching protocol. Each episode is initialized from a predefined intermediate
state representing partial progress toward a source task, and the policy is
then evaluated on whether it can complete the updated target instruction from
that state. All evaluated policies and methods receive exactly the same robot
state, object configuration, and target instruction. This state-matched design
removes differences caused by pre-switch execution speed and avoids small
variations introduced by policy stochasticity before the switch, making
comparisons across all evaluated models and methods strictly fair by
construction.

The two primary benchmarks thus provide complementary views of instruction
switching. \emph{LIBERO-Goal-Switch} emphasizes realistic online adaptation
from states naturally reached by the evaluated policy, whereas
\emph{CALVIN-Custom-Switch} isolates post-switch adaptation under identical
physical conditions. \emph{CALVIN-Custom-Switch-Random} retains the controlled
offline protocol of \emph{CALVIN-Custom-Switch} while removing its fixed
nominal object locations, providing a complementary evaluation under broadly
randomized object placements. In all benchmarks, success is determined by completion of the target task after the instruction change.

\subsection{CALVIN-Custom-Switch Details}
\label{app:calvin_custom_switch_details}

\paragraph{Scene and tasks.}
CALVIN-Custom-Switch is built on top of the CALVIN simulator. We design a tabletop scene containing three colored blocks---red, pink, and blue---and a press button. The benchmark contains four atomic tasks with fixed natural-language instructions: \emph{``lift the red block''}, \emph{``lift the pink block''}, \emph{``lift the blue block''}, and \emph{``press the button''}.

\paragraph{Demonstration generation.}
Since the evaluated pretrained policies do not directly solve these custom tasks, we construct a small demonstration dataset for finetuning. Demonstrations are generated by a rule-based scripted oracle. For each episode, object poses and the robot initial configuration are perturbed around nominal values, invalid configurations such as object collisions are rejected, and the scene is allowed to settle before execution.

For block-lifting tasks, the oracle moves above the target block, descends to grasp it, closes the gripper, and lifts it to a fixed height. For the button-pressing task, the oracle moves above the button and presses it downward with the gripper closed. We retain only successful rollouts verified by the CALVIN task oracle. In our experiments, we collect 50 demonstrations per task and use them to finetune the evaluated policies.

These 200 demonstrations contain only standard single-instruction episodes and no mid-execution instruction switches. Thus, the evaluated policies learn the four underlying manipulation tasks without being explicitly trained on the switching behavior evaluated by the benchmark.

\paragraph{Switching-state construction.}
To construct switching states, we sample intermediate robot configurations along a scripted reference path for each source task. The path moves from the default robot pose toward a pre-grasp or pre-contact pose and then toward the corresponding interaction pose. We sample a progress value along this path, solve for the corresponding robot joint configuration using inverse kinematics, and save the settled robot and scene state. Object poses are sampled near fixed nominal locations with small perturbations, and all states are generated with random seed 0.

We divide the reference path into early, middle, and late progress intervals. Using a 30-step reference schedule, these intervals are reported as the switching-step ranges $[9,15]$, $[15,21]$, and $[21,27]$. These ranges indicate the progress of the saved state along the source-task reference path; they do not imply that the evaluated policy executes the source instruction online for the corresponding number of actions. States from later ranges lie further along the source-task behavior and generally make redirection toward a new target more difficult.

\paragraph{Evaluation protocol.}
For each switching range, we generate 20 object configurations and construct one intermediate state for each of the four possible source tasks. Each intermediate state is paired with the three distinct target tasks, yielding
\[
    20 \times 4 \times 3 = 240
\]
instruction-switching episodes per range. The state-generation scripts additionally produce 80 diagonal source--target controls in which the instruction remains unchanged. These no-switch controls are excluded from the switch-only results in Table~\ref{tab:main_results} and from the reported Switch Success Rate.

At evaluation time, the policy is initialized directly from the saved intermediate state and receives only the target instruction. It may execute at most 180 environment actions. An episode is considered successful only if the CALVIN task oracle verifies completion of the target task; completing the source task or another non-target task does not count as success. All policies and methods are evaluated from exactly the same saved states.

\subsection{CALVIN-Custom-Switch-Random Details}
\label{app:calvin_random_details}

The standard CALVIN-Custom-Switch benchmark samples object poses as small
perturbations around fixed nominal locations. As a result, demonstration
trajectories of the same task tend to occupy relatively concentrated regions
of the workspace, which may induce strong and spatially localized trajectory
priors. This raises a natural question: does the difficulty of mid-execution
instruction switching persist when such fixed object-centered spatial
regularities are removed, and can CAS remain effective under randomized object
placements?

To study this question, we construct
\emph{CALVIN-Custom-Switch-Random}, a qualitatively different variant in which
block locations are sampled over a broader workspace region without fixed
nominal centers. Unlike the concentration-controlled variants, which preserve
task-specific nominal object locations and vary only the perturbation range,
the \emph{random} regime substantially changes the spatial organization of the
resulting task trajectories, as illustrated in
Fig.~\ref{fig:traj_regimes}~(d). Consequently, task trajectories are no longer
tightly associated with fixed object-centered regions of the workspace.

\paragraph{Randomized object placements and demonstrations.}
Instead of sampling block poses as perturbations around task-specific nominal
locations, we uniformly sample the center of each block over the entire valid
tabletop region. This removes the fixed spatial association between each block
and a particular region of the workspace, while ensuring that sampled block
locations remain within the valid manipulation area. We generate training
demonstrations under this randomized placement distribution using the same
scripted-oracle procedure as in CALVIN-Custom-Switch. For each of the four
tasks, we collect 100 successful demonstration trajectories, yielding 400
training trajectories in total.

\subsection{LIBERO-Goal-Switch Details}
\label{app:libero_goal_switch_details}

\paragraph{Task suite and shared initial condition.}
LIBERO-Goal-Switch is built on LIBERO-Goal, which contains ten language-conditioned manipulation tasks sharing the same scene layout. We use a common physical initial condition for every episode: the first official initial state associated with LIBERO-Goal task 0, whose instruction is \emph{``open the middle drawer of the cabinet''}. This association is used only to identify the stored physical state; task 0 otherwise has no privileged role in the evaluation. Reusing the same state across all source--target pairs ensures that every online rollout starts from an identical robot configuration and object arrangement. \rev{Results with a different initial state for every switching pair are reported in Appendix~\ref{app:initial_states}.}

\paragraph{Online switching protocol.}
After loading the shared initial state, we execute ten no-op actions to allow the simulator to stabilize. These actions are not predicted by the policy and are not counted toward the switching step. We denote the resulting observation by $V_0$. The policy then receives the source instruction and executes exactly $t_s$ environment actions, reaching the policy-dependent switching observation $V_{t_s}$. At this point, the active instruction is replaced by the target instruction. For the main evaluation, we consider
\[
    t_s \in \{10,15,20,25\},
\]
covering progressively later stages of source-task execution.

All evaluated policies predict action chunks rather than individual actions. At the switching boundary, any remaining source-conditioned actions in the current chunk are discarded, and the policy replans from $V_{t_s}$ under the updated target instruction. For policies that additionally maintain context across action chunks, such as LingBot-VA, we reset the accumulated cross-chunk context at the switching boundary, so that the target phase starts from the current observation and the updated instruction without retaining source-conditioned temporal context. All evaluated methods use the unmodified base policy before the switch; CAS and other post-switch interventions are activated only after the target instruction is issued.

\paragraph{Task pairs and success metric.}
We enumerate all ordered pairs of distinct source and target tasks. With ten tasks, each switching step contains
\[
    10 \times 9 = 90
\]
online instruction-switching episodes. Source--target pairs are ordered, so switching from task $i$ to task $j$ and switching from task $j$ to task $i$ are treated as two distinct switching scenarios. Each ordered pair is evaluated once at each value of $t_s$, yielding a total of 360 episodes per policy.

After the switch, the policy may execute at most 300 environment actions. Success is determined solely by the official LIBERO completion condition for the target task; completing the source task or any other non-target task does not count as success. For each switching step, we report the mean success rate over all 90 ordered off-diagonal task pairs:
\[
\mathrm{Switch\ Success\ Rate}(t_s)
=
\frac{1}{90}
\sum_{(i,j):\,i\neq j}
\mathbf{1}
\!\left[
\text{target task } j \text{ is completed after switching from task } i
\right].
\]

\section{Analyses of Mid-Execution Instruction Switching}
\label{app:additional_experimental_details_and_results}

\subsection{Analysis of Visual Priors}
\label{app:obs2}

\rev{We examine visual priors from four complementary perspectives: modality interventions test behavioral reliance on visual input, an online stop test measures persistence despite explicit stop supervision, fixed-state attention probes examine whether an updated instruction redirects visual attention, and targeted visual interventions examine which robot--object cues preserve task-like behavior under a conflicting instruction.}

\begin{table}[t]
    \centering
    \caption{
        Average LIBERO success rates (\%) under reported clean evaluation and inference-time modality interventions.
        Results are averaged over LIBERO-Spatial, LIBERO-Object, LIBERO-Goal, and LIBERO-Long.
    }
    \label{tab:obs2_modality_avg}
    \small
    \setlength{\tabcolsep}{7pt}
    \begin{tabular}{lcccc}
        \toprule
        & \multicolumn{1}{c}{Clean} & \multicolumn{3}{c}{Intervened Modality} \\
        \cmidrule(lr){2-2}
        \cmidrule(lr){3-5}
        Model & Reported & Vision & Instruction & Proprioception \\
        \midrule
        OpenVLA-OFT          & 97.1 & 0.0 & 60.3 & 41.7 \\
        $\pi_0$              & 94.2 & 1.7 & 19.3 & 7.9  \\
        $\pi_{0.5}$          & 96.9 & 1.9 & 55.6 & --   \\
        Fast-WAM             & 97.6 & 0.0 & 46.0 & 86.6 \\
        LingBot-VA$^\dagger$ & 98.5 & 0.1 & 58.7 & --   \\
        \bottomrule
    \end{tabular}

    \vspace{2pt}
    \begin{minipage}{0.97\linewidth}
        \footnotesize
        \textit{Note.}
        Clean results are taken from the corresponding papers or released checkpoints rather than re-evaluated in our intervention runs.
        A dash indicates that the standard policy does not consume an independent proprioceptive input.
        $^\dagger$For LingBot-VA, only LIBERO-Long uses the official released checkpoint in our intervention experiments.
        LIBERO-Spatial, LIBERO-Object, and LIBERO-Goal use community-released suite-specific checkpoints.
    \end{minipage}
\end{table}

\begin{table}[t]
    \centering
    \caption{
        Suite-level LIBERO success rates (\%) under reported clean evaluation and inference-time modality interventions.
        Each suite contains 10 tasks with 50 evaluation episodes per task.
    }
    \label{tab:obs2_modality_detail}
    \small
    \setlength{\tabcolsep}{6pt}
    \begin{tabular}{lccccc}
        \toprule
        Model & Spatial & Object & Goal & Long & Avg. \\
        \midrule

        \multicolumn{6}{l}{\textit{Clean (reported)}} \\
        OpenVLA-OFT          & 97.6 & 98.4  & 97.9 & 94.5 & 97.1 \\
        $\pi_0$              & 96.8 & 98.8  & 95.8 & 85.2 & 94.2 \\
        $\pi_{0.5}$          & 98.8 & 98.2  & 98.0 & 92.4 & 96.9 \\
        Fast-WAM             & 98.2 & 100.0 & 97.0 & 95.2 & 97.6 \\
        LingBot-VA$^\dagger$ & 98.5 & 99.6  & 97.2 & 98.5 & 98.5 \\
        \addlinespace

        \midrule
        \multicolumn{6}{l}{\textit{Vision intervention}} \\
        OpenVLA-OFT          & 0.0 & 0.0 & 0.0 & 0.0 & 0.0 \\
        $\pi_0$              & 0.0 & 0.0 & 6.8 & 0.0 & 1.7 \\
        $\pi_{0.5}$          & 0.0 & 0.0 & 7.4 & 0.0 & 1.9 \\
        Fast-WAM             & 0.0 & 0.0 & 0.0 & 0.0 & 0.0 \\
        LingBot-VA           & 0.4 & 0.0 & 0.0 & 0.0 & 0.1 \\
        \addlinespace

        \midrule
        \multicolumn{6}{l}{\textit{Instruction intervention}} \\
        OpenVLA-OFT          & 58.4 & 96.2 & 9.2  & 77.4 & 60.3 \\
        $\pi_0$              & 26.2 & 30.2 & 1.6  & 19.0 & 19.3 \\
        $\pi_{0.5}$          & 67.4 & 66.6 & 12.8 & 75.6 & 55.6 \\
        Fast-WAM             & 40.0 & 86.6 & 4.2  & 53.2 & 46.0 \\
        LingBot-VA           & 82.6 & 58.0 & 9.8  & 84.2 & 58.7 \\
        \addlinespace

        \midrule
        \multicolumn{6}{l}{\textit{Proprioception intervention}} \\
        OpenVLA-OFT          & 7.2  & 34.6 & 31.4 & 93.4 & 41.7 \\
        $\pi_0$              & 6.2  & 4.0  & 17.0 & 4.4  & 7.9 \\
        Fast-WAM             & 82.8 & 89.8 & 95.6 & 78.2 & 86.6 \\
        \bottomrule
    \end{tabular}

    \vspace{2pt}
    \begin{minipage}{0.97\linewidth}
        \footnotesize
        \textit{Note.}
        Clean results are taken from the corresponding papers or released checkpoints rather than re-evaluated in our intervention runs.
        For the intervention experiments, all non-intervened inputs and model-specific inference settings are kept unchanged within each policy.
        $\pi_{0.5}$ and LingBot-VA are omitted from the proprioception block because their standard inference configurations do not consume an independent proprioceptive input.
        $^\dagger$For LingBot-VA, only LIBERO-Long uses the official released checkpoint in our intervention experiments.
        LIBERO-Spatial, LIBERO-Object, and LIBERO-Goal use community-released suite-specific checkpoints.
    \end{minipage}
\end{table}

\subsubsection{Inference-Time Modality Interventions}

\paragraph{Experimental setup.}
We evaluate inference-time modality interventions on LIBERO-Spatial, LIBERO-Object, LIBERO-Goal, and LIBERO-Long.
Each suite contains 10 tasks, and we evaluate all 50 official initial states for each task, resulting in 500 episodes per model--suite--intervention condition and 2,000 episodes per model--intervention condition across the four suites.
For each policy, we preserve its native evaluation protocol and modify only the modality under study; all non-intervened inputs, environment dynamics, action generation and post-processing, rollout horizon, and success criteria remain unchanged.
Thus, the intervention conditions are controlled within each policy, while different policies retain their respective standard inference procedures.
No CAS steering or additional counterfactual policy queries are used in these experiments.

For the \emph{vision intervention}, we replace all visual observations provided to the policy with the same spatially constant RGB image.
The image has constant pixel value $(128,118,109)$, obtained from the channel-wise mean of a reference LIBERO observation.
For policies that consume both a third-person view and a wrist-camera view, both visual inputs are replaced by this same constant image at every policy query.
The original task instruction and, when applicable, the real-time proprioceptive state remain unchanged.
The environment itself continues to evolve and render normally; only the visual observations supplied to the policy are replaced.

For the \emph{instruction intervention}, every task-specific instruction is replaced by the fixed instruction ``remain still.'', while the visual observations and other available inputs remain unchanged.
The environment task and its success criterion are not modified: an episode is still counted as successful only if the policy completes the original LIBERO task.
Therefore, high success under this intervention indicates that task-specific behavior can persist despite removing the original task-specific language, whereas low success alone does not imply that the robot actually remains stationary.

For the \emph{proprioception intervention}, we freeze the proprioceptive input at its value from the first policy query of each episode.
After the model's standard stabilization procedure, we record the 8-D state consisting of the end-effector position, end-effector orientation, and gripper joint positions, and repeatedly provide this fixed state at all subsequent policy queries.
The physical robot state and visual observations continue to evolve normally.
This intervention therefore removes online proprioceptive updates rather than removing proprioception entirely.
We apply it to OpenVLA-OFT, $\pi_0$, and Fast-WAM.
The standard inference configurations of $\pi_{0.5}$ and LingBot-VA do not consume an independent proprioceptive input, so no corresponding intervention is reported for these models.

\paragraph{Model-specific evaluation details.}
For Fast-WAM, we use the base \texttt{libero\_uncond\_2cam224} checkpoint used by the official LIBERO evaluation, rather than an optional IDM variant.
We follow its native evaluation protocol with random seed 42, 30 stabilization steps after environment reset, replanning every 10 environment steps, and an action horizon of 32.
The same checkpoint and inference settings are used across the vision, instruction, and proprioception interventions, with only the intervened modality changed.

For LingBot-VA, checkpoint availability differs across LIBERO suites.
In our intervention experiments, LIBERO-Long uses the official released \texttt{lingbot-va-posttrain-libero-long} checkpoint.
Official released checkpoints for the other three suites were not available in the evaluation setup we used; therefore, LIBERO-Spatial, LIBERO-Object, and LIBERO-Goal are evaluated using community-released suite-specific LingBot-VA checkpoints.
All LingBot-VA intervention conditions within a given suite use the same corresponding checkpoint.
We retain the model's native inference mechanism, including its action/video history and KV cache, and intervene only on the external input modality under study.
LingBot-VA does not consume robot proprioception as an independent input in this inference configuration.

\paragraph{Modality intervention results.}
The modality interventions reveal a strikingly consistent dependence on visual input.
Under the reported clean settings, all five policies achieve high average LIBERO success rates, ranging from 94.2\% to 98.5\%.
In contrast, replacing the visual observations with a spatially constant image reduces the average success rate to at most 1.9\% for every evaluated policy.
OpenVLA-OFT and Fast-WAM drop to 0.0\%, LingBot-VA to 0.1\%, and $\pi_0$ and $\pi_{0.5}$ to 1.7\% and 1.9\%, respectively.
This collapse is also highly consistent across individual suites: almost all model--suite combinations achieve zero success under the vision intervention, with the main residual successes occurring on LIBERO-Goal for $\pi_0$ and $\pi_{0.5}$, which achieve 6.8\% and 7.4\%, respectively.
We inspect these successful rollouts and find that they do not resemble normal task-conditioned execution.
Instead, the policy repeatedly produces nearly the same motion, continuously translating toward the stove while rotating in a nearly fixed direction.
This repetitive motion occasionally contacts the stove knob and accidentally satisfies the ``turn on the stove'' success condition.
The residual success under vision intervention therefore largely reflects coincidental task completion rather than coherent execution based on the visual scene.

Replacing the task instruction produces a markedly different outcome.
Substantial task success remains for several policies even though they no longer receive the original task-specific language.
OpenVLA-OFT, $\pi_{0.5}$, Fast-WAM, and LingBot-VA retain average success rates of 60.3\%, 55.6\%, 46.0\%, and 58.7\%, respectively, while $\pi_0$ retains 19.3\%.
The residual performance is strongly suite-dependent.
For example, OpenVLA-OFT still achieves 96.2\% on LIBERO-Object and 77.4\% on LIBERO-Long, while LingBot-VA achieves 82.6\% on LIBERO-Spatial and 84.2\% on LIBERO-Long.
In contrast, success under the instruction intervention is much lower on LIBERO-Goal for all models, ranging from 1.6\% to 12.8\%.
These results do not imply that language is unimportant; rather, they show that removing the original task-specific instruction does not necessarily eliminate recognizable task-like manipulation behavior.

The proprioception intervention exhibits substantially greater variation across policies.
OpenVLA-OFT retains 41.7\% average success, $\pi_0$ drops to 7.9\%, whereas Fast-WAM remains at 86.6\%.
Thus, the contribution of online proprioceptive updates appears strongly policy-dependent.
Notably, $\pi_0$ is also sensitive to the proprioception intervention, suggesting that current robot-state information can play an important role in its behavior in addition to visual input.
This observation is compatible with our KNN-based design in CAS, which uses the current end-effector position to estimate proximity to task trajectories.

In contrast, the near-complete failure under visual intervention is remarkably consistent across all five evaluated architectures.

\paragraph{Interpreting residual success under instruction intervention.}
We further inspect why substantial success can remain under the instruction intervention.
Except for LIBERO-Goal, the other three LIBERO suites contain visual scene or configuration cues that can often provide information about which task is likely to be executed.
Such visual cues may therefore partially explain the relatively high residual success rates when task-specific language is replaced.
Although these visual cues can be task-discriminative, they do not necessarily drive the policy toward the very task behavior associated with the original environment task.
For example, by inspecting rollouts on LIBERO-Object, we observe many cases in which the policy continues to exhibit a recognizable task-like manipulation behavior after the instruction is replaced by ``remain still.'', but the resulting behavior corresponds to a different task from the one originally specified.

Importantly, our interpretation that robot policies are strongly influenced by visual priors is not based solely on the success-rate differences across modality interventions.
Success rate only indicates whether the resulting behavior happens to satisfy the original task predicate; it does not fully characterize what behavior the policy is attempting to execute.
A failed episode under instruction intervention may still contain a highly recognizable task-like manipulation behavior, whereas, as discussed above, a successful episode under vision intervention may result from accidental interaction with the environment.

\paragraph{Behavioral analysis.}
Our main evidence also comes from direct inspection and visualization of policy behavior.
Consistent with the observation in Sec.~\ref{obs2}, under instruction intervention, policies frequently continue to exhibit recognizable task-like manipulation behaviors, as if still executing a plausible task.
This remains true even in cases where the resulting behavior corresponds to a task different from the one originally specified.
In contrast, under vision intervention, such task-likeness is substantially degraded, manifesting as erratic motions, repetitive motion along a nearly fixed direction, or movements with noticeably reduced magnitudes.

\begin{figure*}[t]
    \centering
    \includegraphics[width=\textwidth]{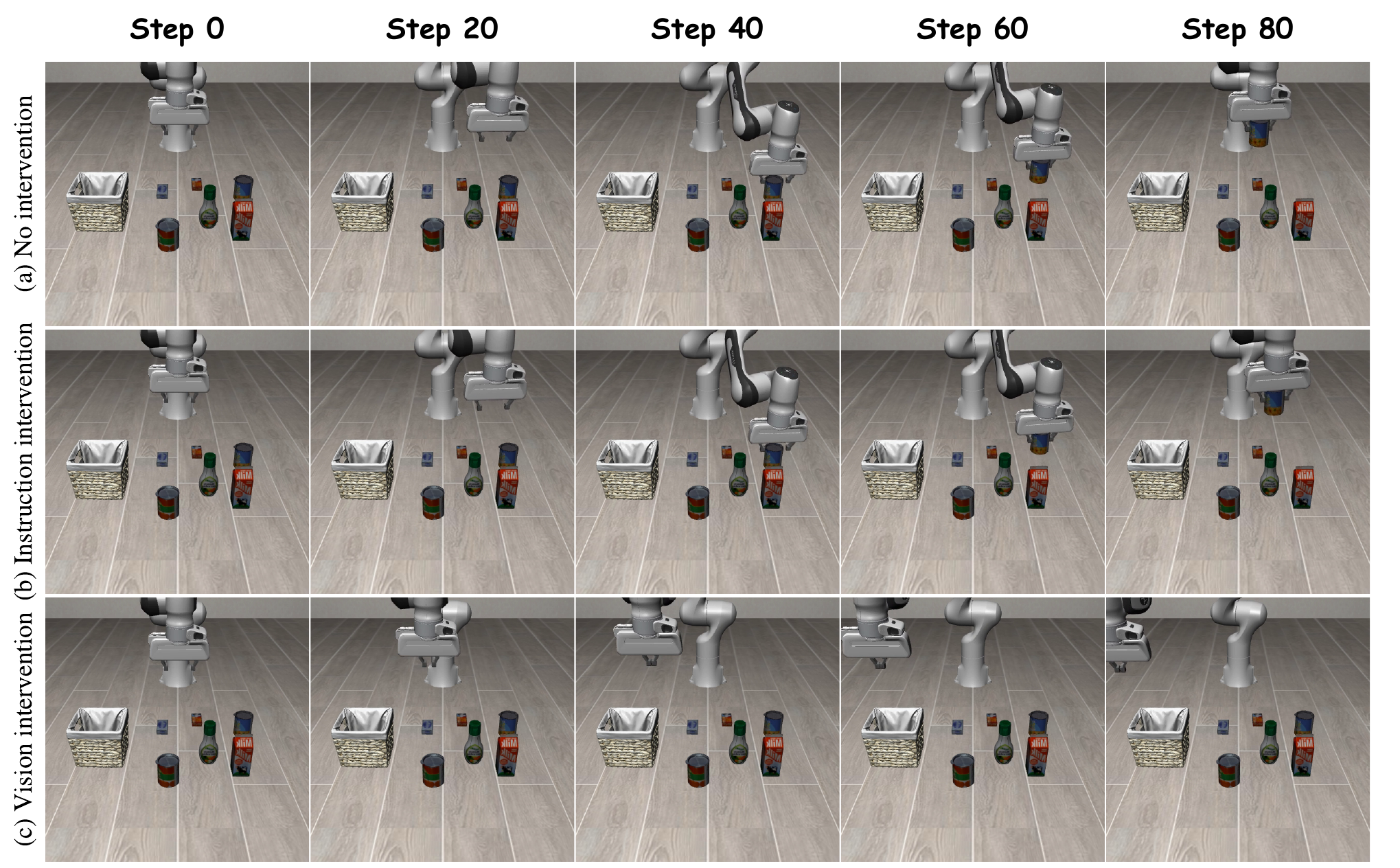}
    \caption{
    \textbf{Representative rollout sequences for OpenVLA-OFT.}
    Under instruction intervention, the policy behaves similarly to clean inference and successfully grasps the correct object, whereas under vision intervention the gripper repeatedly translates toward the right without exhibiting a recognizable manipulation behavior.
    }
    \label{fig:obs2_oft}
\end{figure*}

\begin{figure*}[t]
    \centering
    \includegraphics[width=\textwidth]{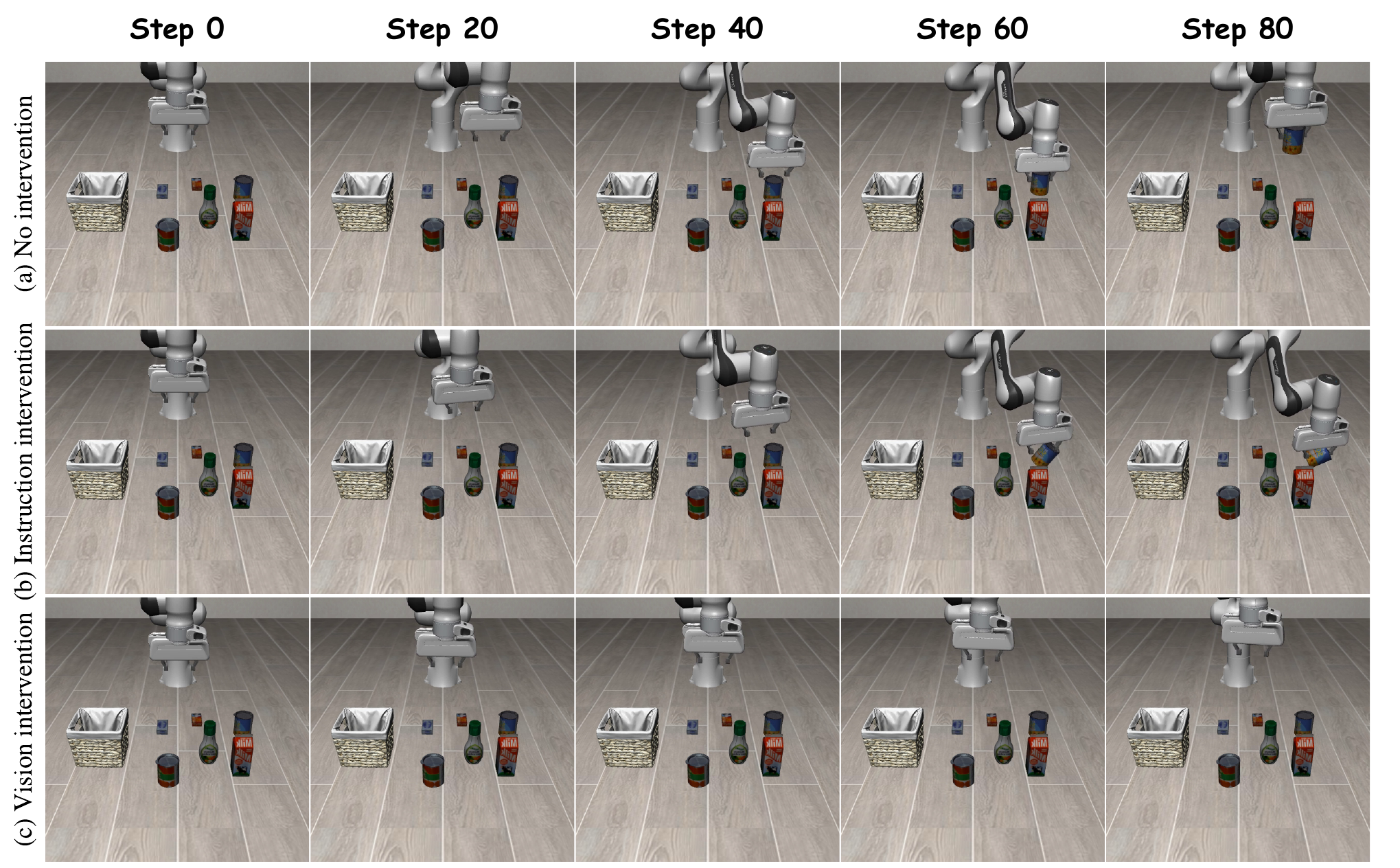}
    \caption{
    \textbf{Representative rollout sequences for $\pi_0$.}
    Under instruction intervention, the policy still reaches and grasps the correct object, although it does not complete the lift.
    Under vision intervention, the gripper moves slowly toward the right with substantially reduced motion magnitude.
    }
    \label{fig:obs2_pi0}
\end{figure*}

\begin{figure*}[t]
    \centering
    \includegraphics[width=\textwidth]{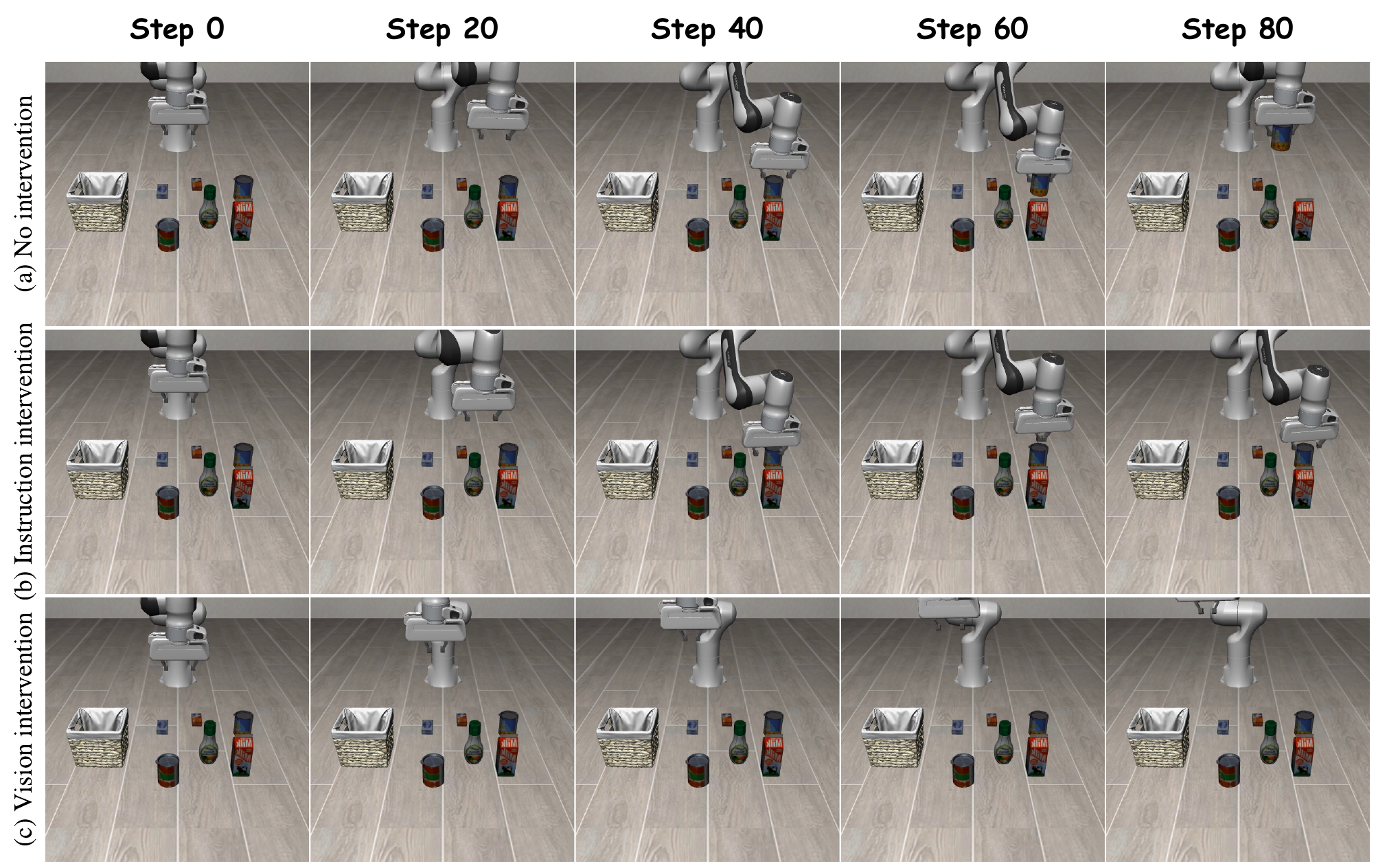}
    \caption{
    \textbf{Representative rollout sequences for $\pi_{0.5}$.}
    Under instruction intervention, the first grasp attempt is slightly misaligned, but the policy still approaches and interacts with the correct target.
    Under vision intervention, the policy instead exhibits erratic motion without a clear task-directed manipulation pattern.
    }
    \label{fig:obs2_pi05}
\end{figure*}

\begin{figure*}[t]
    \centering
    \includegraphics[width=\textwidth]{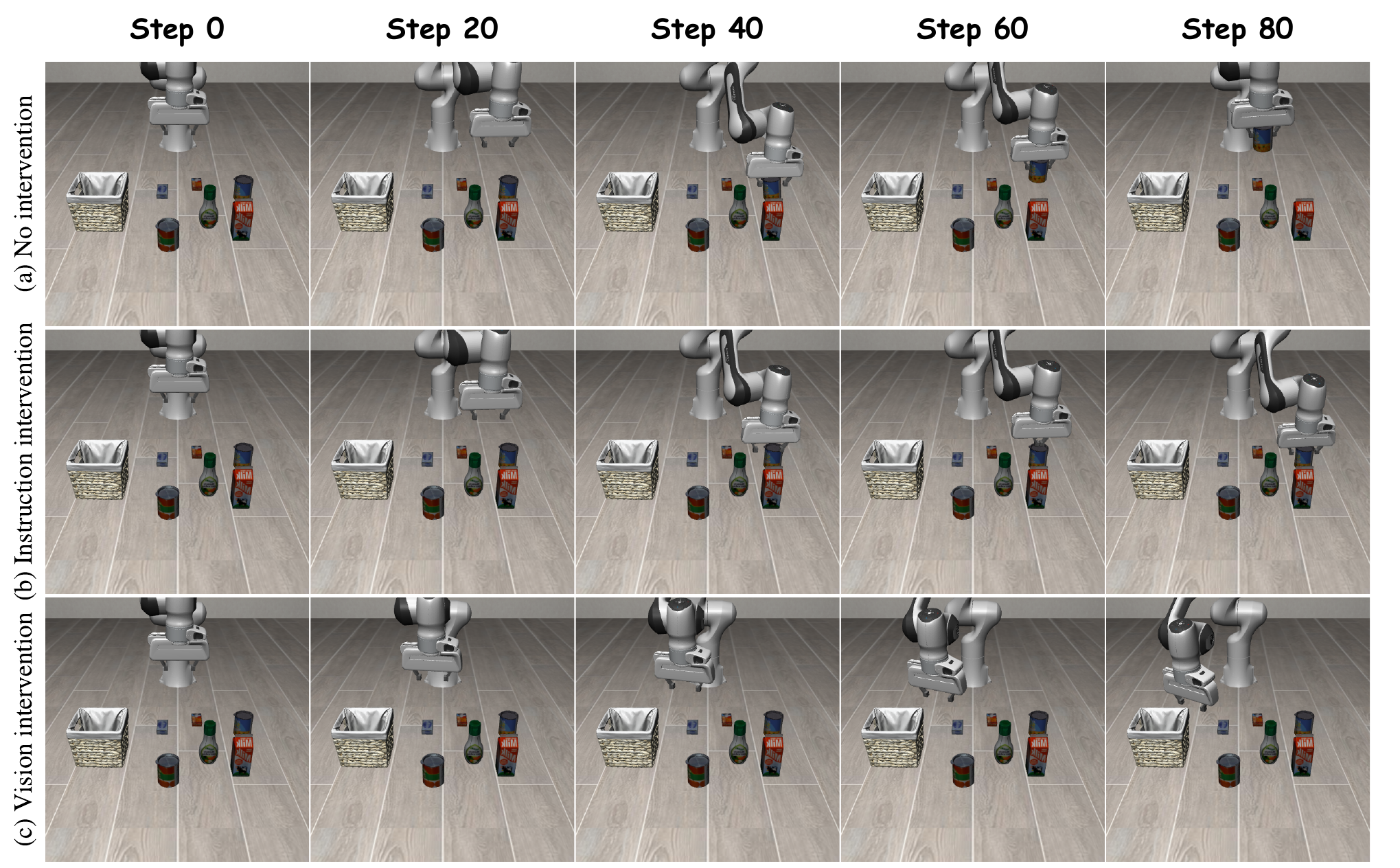}
    \caption{
    \textbf{Representative rollout sequences for Fast-WAM.}
    Under instruction intervention, the first grasp attempt is misaligned, but the policy continues to approach the correct target.
    Under vision intervention, the robot repeatedly moves toward the basket while rotating, substantially degrading the recognizable task-like manipulation behavior.
    }
    \label{fig:obs2_fastwam}
\end{figure*}

\begin{figure*}[t]
    \centering
    \includegraphics[width=\textwidth]{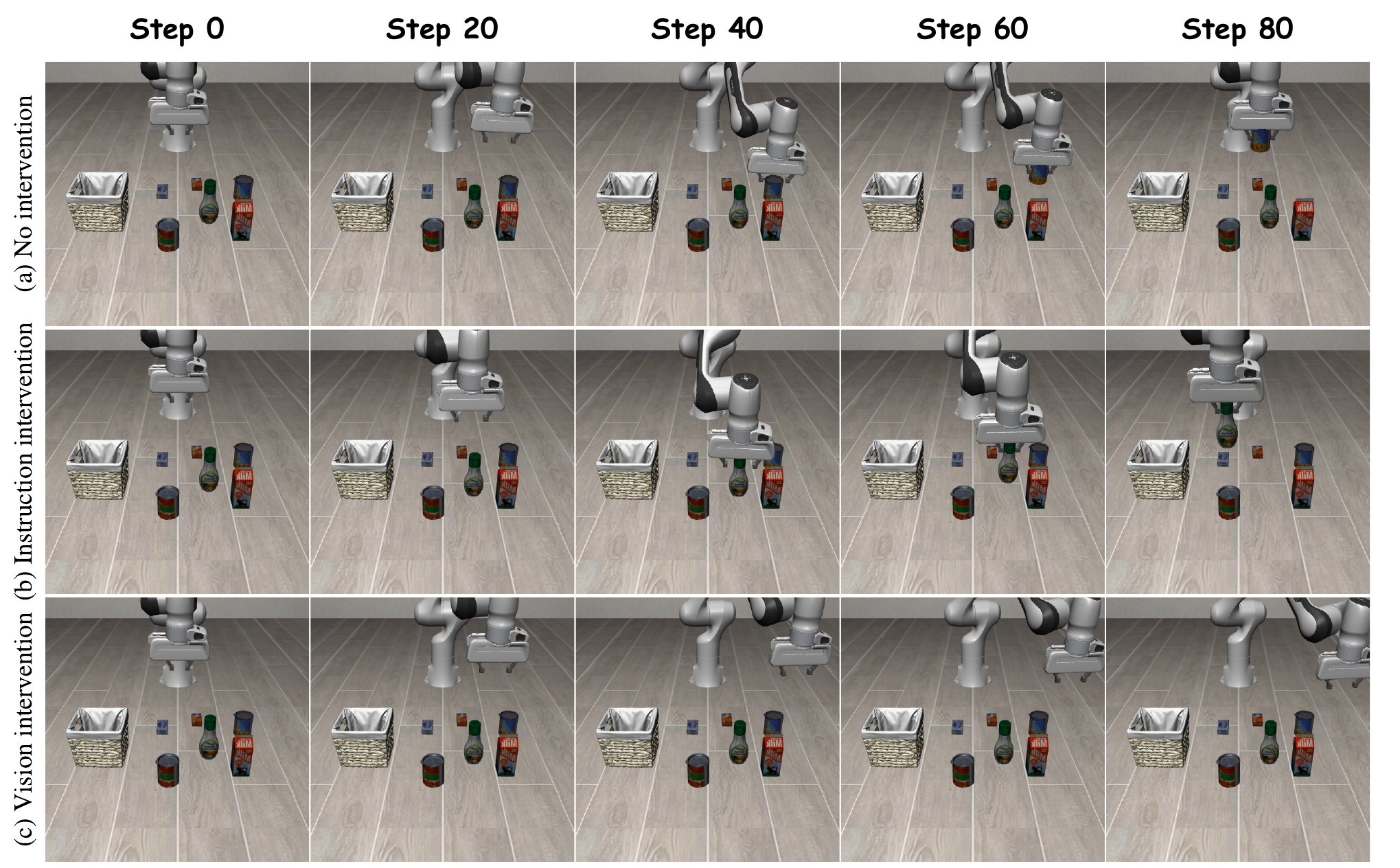}
    \caption{
    \textbf{Representative rollout sequences for LingBot-VA.}
    Under instruction intervention, the policy still executes a coherent manipulation behavior but grasps an object corresponding to a different task.
    Under vision intervention, the gripper instead drifts toward the right and slightly downward without exhibiting comparable task-like structure.
    }
    \label{fig:obs2_lingbot_va}
\end{figure*}

Below, we visualize representative rollouts for all five evaluated policies under three conditions: clean inference, instruction intervention, and vision intervention
(Figs.~\ref{fig:obs2_oft}, \ref{fig:obs2_pi0}, \ref{fig:obs2_pi05}, \ref{fig:obs2_fastwam}, and \ref{fig:obs2_lingbot_va}). We visualize each rollout using five frames sampled every 20 environment steps, at steps 0, 20, 40, 60, and 80. At 20 fps, these frames correspond to video timestamps
  0, 1, 2, 3, and 4 seconds, respectively.
These examples illustrate the contrast in behavioral task-likeness across input conditions.

For OpenVLA-OFT (Fig.~\ref{fig:obs2_oft}), the rollout under instruction intervention closely resembles the clean rollout, with the robot successfully reaching and grasping the correct object in both cases.
Under vision intervention, however, the gripper instead repeatedly translates toward the right without exhibiting a recognizable manipulation behavior.
For $\pi_0$ (Fig.~\ref{fig:obs2_pi0}), the instruction-intervened policy still reaches and grasps the correct object, although it does not successfully complete the lift.
In contrast, under vision intervention, the gripper moves only slowly toward the right.
Although this direction is broadly consistent with motion toward the target, the movement magnitude is substantially reduced compared with normal task execution.
For $\pi_{0.5}$ (Fig.~\ref{fig:obs2_pi05}), the first grasp attempt under instruction intervention is slightly misaligned, but the policy nevertheless approaches and continues interacting with the correct target.
Its behavior under vision intervention is qualitatively different, exhibiting erratic motion without a clear task-directed manipulation pattern.
Fast-WAM (Fig.~\ref{fig:obs2_fastwam}) shows a similar pattern under instruction intervention: despite an initially misaligned grasp, the robot continues to approach the correct target.
With vision intervention, however, it repeatedly moves toward the basket while simultaneously rotating, rather than carrying out a recognizable task-specific manipulation sequence.
Finally, for LingBot-VA (Fig.~\ref{fig:obs2_lingbot_va}), the instruction-intervened policy exhibits a particularly informative failure mode: it successfully grasps an object, but not the one specified by the original task, effectively carrying out the behavior of a different task.
Under vision intervention, the gripper instead initially drifts toward the right and slightly downward.

Across these examples, replacing the task instruction does not necessarily eliminate structured manipulation behavior: the policy may still approach and interact with meaningful objects, successfully grasp the intended object, or even execute a coherent but different task.
In contrast, replacing visual input substantially degrades this task-likeness, producing behaviors such as nearly unidirectional drift, greatly reduced motion, erratic movement, or repetitive translation and rotation.
These qualitative observations complement the success-rate results and provide direct behavioral evidence for the strong influence of visual priors on policy execution.

\FloatBarrier

\FloatBarrier
\subsubsection{Mid-Execution Stop Test after Training on the ``Remain Still'' Instruction}

\begin{figure}[t]
    \centering
    \begin{subfigure}[t]{0.49\linewidth}
        \centering
        \includegraphics[width=\linewidth]{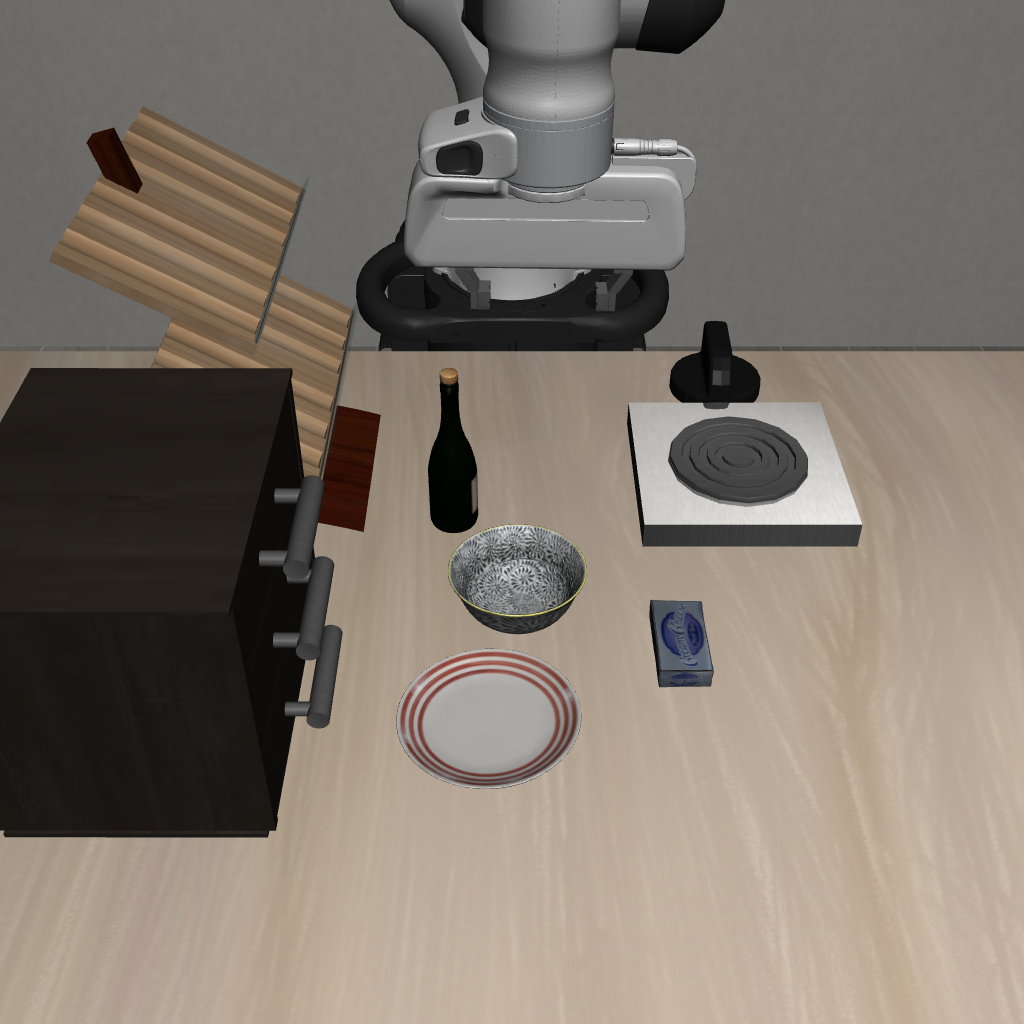}
        \caption{At the instruction switch.}
        \label{fig:obs2_rs_high}
    \end{subfigure}
    \hfill
    \begin{subfigure}[t]{0.49\linewidth}
        \centering
        \includegraphics[width=\linewidth]{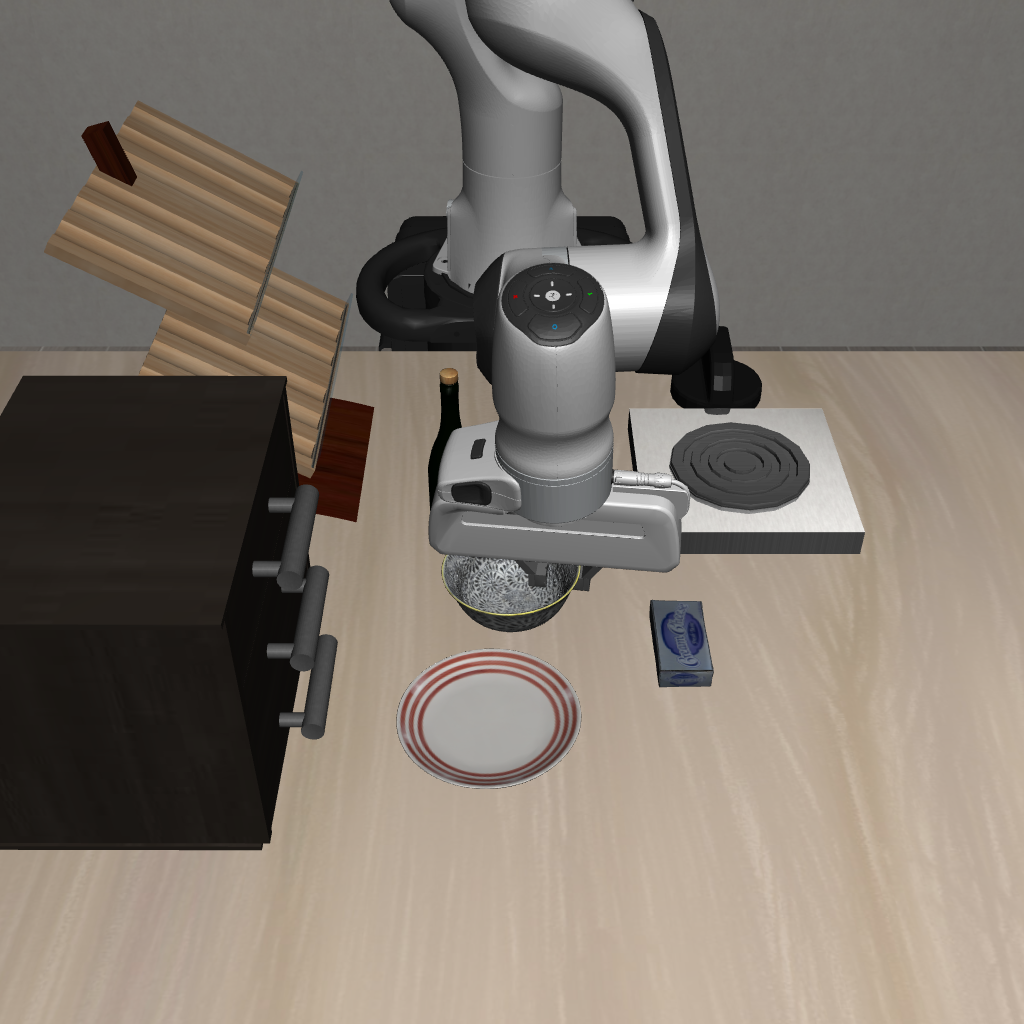}
        \caption{After 100 post-switch steps.}
        \label{fig:obs2_rs_final}
    \end{subfigure}

    \caption{
    Representative mid-execution stop failure after explicit training on the
    ``remain still'' instruction.
    The source instruction is ``put the bowl on the plate.'' 
    At the switching state, the instruction is changed to ``remain still''
    (left). Despite the updated instruction, the robot continues to move
    substantially over the following 100 environment steps (right), resulting
    in a cumulative end-effector path length of 30.74\,cm and a final
    displacement of 23.81\,cm from the switching state.
    }
    \label{fig:obs2_remain_still}
\end{figure}

\paragraph{Experimental setup.}
We further test whether OpenVLA-OFT can reliably halt an ongoing behavior when the ``remain still'' instruction is explicitly included in training. After adding explicit remain-still supervision, we evaluate online instruction switching from each of the ten LIBERO-Goal source tasks to ``remain still'' at different stages of execution, and measure how much the robot continues to move after the switch.

\paragraph{Training with explicit remain-still supervision.}
We augment the OpenVLA-OFT training data with 100 explicit remain-still trajectories. Each trajectory contains 20 environment steps and is initialized from a robot state sampled across the workspace covered by LIBERO-Goal demonstrations, with the sampled states distributed across the ten LIBERO-Goal task contexts. Throughout each trajectory, the six end-effector translation and rotation action dimensions are set to zero, while the gripper is kept in its initial open state. Thus, the training data explicitly associates the instruction ``remain still'' with zero commanded end-effector motion.

To retain supervision for the original LIBERO-Goal tasks during this additional training, we also include 50 original LIBERO-Goal demonstration trajectories, with five demonstrations sampled from each of the ten tasks. We finetune OpenVLA-OFT on this mixed dataset and use the resulting Step-55,000 checkpoint for evaluation.

\paragraph{Online mid-execution stop evaluation.}
We evaluate the trained policy using an online instruction-switching protocol. For each of the ten LIBERO-Goal tasks, the policy first executes the corresponding source instruction from the common initial state. We then switch the instruction to ``remain still'' at
\[
t_s \in \{5,10,15,20,25\},
\]
resulting in $10 \times 5 = 50$ switching rollouts in total. At the switching boundary, any unexecuted source-conditioned actions in the current action chunk are discarded, and the policy immediately replans under the ``remain still'' instruction. We then continue execution for 100 environment steps. OpenVLA-OFT executes an open-loop action chunk of eight environment actions before each new policy query. All movement statistics below are computed only over the post-switch portion of the rollout.

\paragraph{Evaluation metrics.}
Let $p_0$ denote the end-effector position at the switching state and $p_t$ its position after $t$ post-switch environment steps. We measure two complementary quantities. First, the cumulative end-effector path length over a horizon $H$ is
\[
L_H = \sum_{t=1}^{H} \|p_t-p_{t-1}\|_2,
\]
which measures the total amount of motion after the stop instruction is issued. Second, we measure the final displacement from the switching state,
\[
D_H = \|p_H-p_0\|_2.
\]
The latter distinguishes substantial displacement from small local oscillations around the switching state. A policy that reliably follows the ``remain still'' instruction should keep both quantities close to zero.

\paragraph{Results.}
Despite explicit remain-still supervision, the policy continues to move substantially after the instruction switch. Table~\ref{tab:obs2_remain_still} summarizes the post-switch motion over horizons of 20, 50, and 100 environment steps.

\begin{table}[t]
\centering
\caption{
Post-switch end-effector motion after switching to ``remain still''
following explicit remain-still training.
Results are aggregated over 50 rollouts across the ten LIBERO-Goal
source tasks and five switching steps.
}
\label{tab:obs2_remain_still}
\small
\setlength{\tabcolsep}{8pt}
\begin{tabular}{c rrr rrr}
\toprule
& \multicolumn{3}{c}{Cumulative Path Length $L_H$ (cm)}
& \multicolumn{3}{c}{Final Displacement $D_H$ (cm)} \\
\cmidrule(lr){2-4}
\cmidrule(lr){5-7}
Post-Switch Horizon & Mean & Median & Std. & Mean & Median & Std. \\
\midrule
20 steps  &  7.85 &  5.49 &  6.20 &  7.07 &  5.10 & 5.86 \\
50 steps  & 20.44 & 18.37 & 14.07 & 12.68 & 12.45 & 7.23 \\
100 steps & 41.29 & 31.66 & 21.48 & 23.20 & 22.26 & 6.34 \\
\bottomrule
\end{tabular}
\end{table}

The 20-step horizon is particularly informative because it matches the length of the remain-still trajectories used during training. Even over this same horizon, the end effector travels 7.85\,cm on average and ends 7.07\,cm away from the switching state. The motion continues to accumulate over longer execution: after 50 steps, the average cumulative path reaches 20.44\,cm, and after 100 steps it reaches 41.29\,cm, with an average final displacement of 23.20\,cm.

The large final displacement also shows that the measured motion cannot be explained solely by small action jitter around the switching state. At the 100-step horizon, every one of the 50 rollouts has both a cumulative path length and a final displacement greater than 10\,cm. The minimum cumulative path length across all rollouts is 11.13\,cm, and even the minimum final displacement is 10.04\,cm.

\paragraph{Qualitative example.}
Figure~\ref{fig:obs2_remain_still} shows a representative rollout whose source instruction is ``put the bowl on the plate.'' At the switching state, the robot has already progressed toward the bowl under the source instruction. The instruction is then changed to ``remain still.'' Nevertheless, over the subsequent 100 environment steps, the robot moves substantially away from the switching state rather than remaining stationary. For this rollout, the cumulative end-effector path length is 30.74\,cm and the final displacement from the switching state is 23.81\,cm. Among rollouts with bowl-related source instructions, we select this example because its cumulative path length and final displacement are close to the aggregate levels across all 50 rollouts, rather than because it exhibits unusually large motion.

\paragraph{Interpretation.}
These results show that explicit remain-still supervision alone does not make the policy reliably stop under online instruction switching. Even after the policy is directly trained to associate ``remain still'' with zero commanded end-effector motion, substantial movement persists when the instruction is issued from states that already reflect partial progress along another task.

Taken together with the modality-intervention results in Appendix~D.1.1, which show that disrupting visual input substantially degrades structured task-like behavior, these findings support the interpretation that visual priors contribute to behavioral persistence during execution. In particular, visually grounded task-progress cues may favor continuation of an ongoing manipulation pattern, making the policy less responsive to updated instructions and thereby contributing to failures under mid-execution instruction switching.

\FloatBarrier
\subsubsection{\rev{Visual Attention under Instruction Updates}}
\label{app:attention_persistence}
\begingroup

We examine whether a new instruction redirects the policy's visual attention toward the updated target. Using OpenVLA-OFT, we replace the instruction while holding the observation and robot state fixed. We inspect attention from the positions used by the action head to image patches in the final language-backbone layer (layer 31), averaged over heads and query positions.

\paragraph{The instruction changes, but attention can remain on the previous target.}
Figure~\ref{fig:attention_switch} illustrates this mismatch in both Cheese-to-Wine and Wine-to-Cheese updates, at source-execution steps 24 and 72, respectively. Although the new instruction specifies a different object, attention remains prominent around the gripper and the previous target, with weaker emphasis on the newly requested object. These examples show how an updated goal can coexist with visual attention still aligned with the preceding task. Heatmaps are normalized within each camera and use separate color scales, so their brightness should be interpreted spatially within each panel.

\paragraph{Attention becomes less responsive to instruction changes during execution.}
We extend the comparison to four LIBERO-Goal tasks: cream cheese into the bowl, wine bottle onto the rack, bowl onto the cabinet, and turning on the stove. From a shared initial configuration, we execute each source task for 24 steps and compare attention under its original instruction and each of the other three instructions, yielding 12 directed comparisons. We make the same instruction comparisons at the initial state. The change between two within-camera attention distributions $p$ and $q$ is measured by total variation, $\mathrm{TV}=\frac{1}{2}\sum_j|p_j-q_j|$; smaller values mean less redistribution when the instruction changes.

The result is consistent across the measured task pairs and both camera views (Table~\ref{tab:attention_instruction_pairs}). Changing the instruction produces 46.1\% less attention redistribution in the third-person view and 54.6\% less in the wrist view at step 24 than initially. Thus, after the robot has progressed under a source instruction, the same change in task language has a smaller effect on where the model attends.

The two analyses provide complementary evidence: Fig.~\ref{fig:attention_switch} shows attention on previously relevant regions under a new instruction, while the four-task comparison shows reduced spatial responsiveness to instruction changes. Initial maps also share scene hotspots; TV quantifies how their spatial attention distributions change with the instruction. Together, these results show that \emph{attention can remain aligned with the preceding task and become less responsive to an updated instruction}, consistent with the influence of visual priors on ongoing behavior. These probes use one initial configuration and one trajectory per source task to characterize same-state predictions.

\begin{table}[!htbp]
\centering
\small
\setlength{\tabcolsep}{8pt}
\caption{\rev{\textbf{Instruction-induced attention changes are smaller during execution.} Mean last-layer TV across all 12 directed task pairs. Each comparison changes only the instruction at a fixed state. The initial baseline contains six unique unordered pairs reused across directions.}}
\label{tab:attention_instruction_pairs}
\begin{tabular}{lrrrr}
\toprule
Camera view & Initial & Step 24 & Reduction & Pairs with lower TV \\
\midrule
Third-person & 0.1562 & 0.0842 & 46.1\% & 12/12 \\
Wrist & 0.1618 & 0.0735 & 54.6\% & 12/12 \\
\bottomrule
\end{tabular}
\end{table}

\endgroup
\FloatBarrier

\clearpage
\subsubsection{\rev{Which Visual Information Guides Early Manipulation?}}
\label{app:robot_state_visual_prior}
\begingroup

The modality experiments establish the importance of vision. We next ask which parts of the visual input most strongly guide the model's behavior. An image contains both the broader scene---the background, surrounding objects, and their arrangement---and cues about the robot's current configuration relative to an object. Distinguishing their roles helps clarify what visually suggests a familiar manipulation: the scene as a whole, or the local configuration from which the robot is about to act.

We examine this question with OpenVLA-OFT in \emph{put the bowl on the plate}, focusing on the initial approach, gripper closure, and lifting stages. Starting partway through approaching the bowl, we replace the task instruction with ``remain still'' to observe the behavior that persists without a task-specific language cue. Across 20 initial configurations, normal camera inputs (A) lead the robot to continue approaching the bowl, close its gripper, and lift the bowl in all 20 cases. This provides the behavior whose visual support we seek to identify.

We first remove the broader scene while preserving the robot and the bowl in both camera views (D; Fig.~\ref{fig:robot_state_visual_inputs}). The background, other objects, and even the destination plate are masked out, yet the bowl is still lifted in 17/20 cases. Much of the surrounding visual information can therefore be removed without eliminating the early manipulation behavior. This directs attention to the information that remains: the robot, the object, and their spatial relationship.

\begin{figure}[!htbp]
    \centering
    \includegraphics[width=\linewidth]{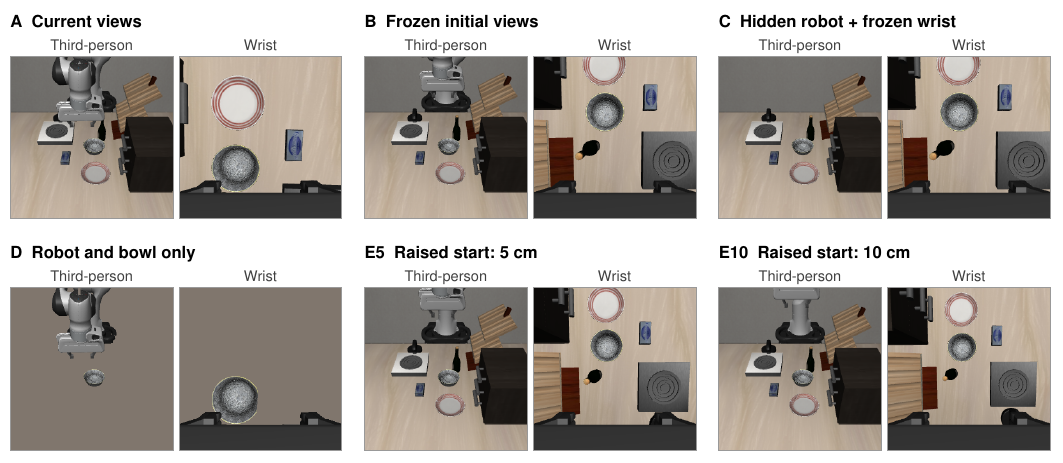}
    \caption{\rev{\textbf{What the policy sees when robot-state cues are preserved or altered.} Each panel pairs third-person and wrist inputs at rollout start. A uses current views; B freezes both views to the initial images; C hides the robot in the current third-person view and freezes the wrist input; D retains only the robot and bowl. A--D share a physical starting state. E5/E10 instead raise the default initial pose by 5/10\,cm and use normal images. Views are re-rendered from recorded starts (configuration 10), upright and before policy preprocessing.}}
    \label{fig:robot_state_visual_inputs}
\end{figure}

The current robot configuration enters the images through two channels: the gripper's relation to the bowl in the third-person view, and the wrist-camera view determined by the robot pose. We alter these cues by freezing both cameras to their initial images (B), or by hiding the robot in the current third-person view while freezing the wrist input (C). Both interventions eliminate bowl lifts and gripper-closing commands across all 20 configurations. The arm still moves substantially, but the movement no longer produces the object interaction observed in A/D. Together with D, this contrast suggests that visual cues tied to the current robot--object configuration play a particularly important role in sustaining the early manipulation behavior.

We further examine whether the behavior changes when the robot configuration changes while normal images remain available. Raising the default initial pose by 5 or 10\,cm (E5/E10) produces no bowl lifts. Instead, all 20 endpoints in each condition are nearest to the \emph{turn on the stove} demonstration point cloud under mean 30-nearest-neighbor distance. Thus, the absence of bowl interaction is accompanied by an endpoint association with another task's demonstrations. The nearest-task identity changes during execution: none of the raised starting positions is nearest to the stove point cloud, whereas all resulting endpoints are.

\begin{table}[!htbp]
    \centering
    \small
    \setlength{\tabcolsep}{6pt}
    \caption{\rev{\textbf{Early manipulation across 20 initial configurations.} A--E10 use ``remain still''; REF uses the original instruction. Approach: EEF-to-bowl distance decreases by $>5$\,cm. Lift: bowl-center height increases by $>2$\,cm at any step in the 60-step window, a proxy for the lifting stage. $\Delta d$: final minus initial EEF-to-bowl distance.}}
    \label{tab:robot_state_visual_prior}
    \begin{tabular}{lrrr}
        \toprule
        Condition & Approach & Bowl lift & Mean $\Delta d$ (cm) \\
        \midrule
        A: current views & 20/20 & 20/20 & $-14.17$ \\
        B: frozen initial views & 0/20 & 0/20 & $+12.65$ \\
        C: hidden robot, frozen wrist & 0/20 & 0/20 & $+11.15$ \\
        D: robot and bowl only & 16/20 & 17/20 & $-8.70$ \\
        E5: raised 5\,cm & 0/20 & 0/20 & $+4.03$ \\
        E10: raised 10\,cm & 0/20 & 0/20 & $-0.66$ \\
        \midrule
        REF: original instruction & 20/20 & 20/20 & $-14.12$ \\
        \bottomrule
    \end{tabular}
    \vspace{3pt}
    \begin{minipage}{0.97\linewidth}
        \footnotesize\rev{Wilson 95\% intervals for bowl-lift proportions: A, 83.9--100\%; D, 64.0--94.8\%; B/C, 0--16.1\%.}
    \end{minipage}
\end{table}

The pattern in Table~\ref{tab:robot_state_visual_prior} extends beyond the original example: among the 19 additional configurations, A produces 19 lifts, D produces 16, and B/C produce none. These results suggest that the local robot--object configuration is a major source of the visual cues supporting this behavior; much of the broader scene is unnecessary for the same early-stage interaction to emerge. This gives our visual trajectory-prior interpretation a concrete basis: cues about the robot's current state can support a familiar manipulation continuation even under ``remain still''. The conditions vary complementary aspects of the input and state: B/C alter visual feedback and visual--proprioceptive consistency, D masks surrounding scene information, and E changes the physical state. Together, the results support a prominent role for robot-state-related visual cues in sustaining the observed manipulation behavior.

\endgroup
\FloatBarrier

\subsection{Non-Target Task Completion in Failed Instruction Switching}
\label{app:wrong_task}

We further quantify how often failed instruction-switching episodes result in the completion of a non-target task.
We analyze the CALVIN-Custom online switching evaluations used in Observation~1, covering four switching ranges,
$t_s \in [3,9], [9,15], [15,21], [21,27]$.
Each policy is evaluated on 240 episodes per switching range, yielding 960 episodes in total.

We define a \emph{wrong-task completion} as an episode in which the target task is not completed, but the robot completes a different task.
This captures a strong form of trajectory hijacking in which the policy does not merely drift toward a non-target trajectory, but ultimately completes a non-target task.
It is therefore a conservative indicator of trajectory hijacking, since partial drift toward a non-target trajectory need not result in completion of that task.

\begin{table}[t]
\centering
\caption{
\textbf{Wrong-task completion as a percentage of all switching episodes.}
Each switching-range column reports the percentage of all 240 episodes in that range for which the target task fails but a non-target task is completed.
Overall aggregates all four switching ranges.
}
\label{tab:wrong_task_all}
\begin{tabular}{lccccc}
\toprule
Policy
& $[3,9]$
& $[9,15]$
& $[15,21]$
& $[21,27]$
& Overall \\
\midrule
OpenVLA-OFT & 9.2 & 12.9 & 18.8 & 20.8 & 15.4 \\
$\pi_0$     & 0.0 & 7.5  & 16.7 & 31.3 & 13.9 \\
$\pi_{0.5}$ & 3.3 & 12.9 & 12.9 & 16.7 & 11.5 \\
Fast-WAM    & 1.3 & 0.4  & 0.8  & 2.1  & 1.1 \\
LingBot-VA  & 0.8 & 2.9  & 5.4  & 0.0  & 2.3 \\
\bottomrule
\end{tabular}
\end{table}

\begin{table}[t]
\centering
\caption{
\textbf{Wrong-task completion as a percentage of target-task failures.}
Each switching-range column reports the percentage of target-task failures in that range for which the robot completes a non-target task.
Overall aggregates target-task failures across all four switching ranges.
}
\label{tab:wrong_task_failure}
\begin{tabular}{lccccc}
\toprule
Policy
& $[3,9]$
& $[9,15]$
& $[15,21]$
& $[21,27]$
& Overall \\
\midrule
OpenVLA-OFT & 11.8 & 14.0 & 20.8 & 22.6 & 17.5 \\
$\pi_0$     & 0.0  & 8.5  & 17.0 & 31.4 & 17.0 \\
$\pi_{0.5}$ & 9.5  & 15.2 & 13.7 & 16.7 & 14.6 \\
Fast-WAM    & 2.0  & 0.4  & 0.8  & 2.1  & 1.3 \\
LingBot-VA  & 2.5  & 4.4  & 5.8  & 0.0  & 3.1 \\
\bottomrule
\end{tabular}
\end{table}

Tables~\ref{tab:wrong_task_all} and~\ref{tab:wrong_task_failure} show that non-target task completion occurs across multiple policy architectures and switching ranges.
For OpenVLA-OFT, $\pi_0$, and $\pi_{0.5}$, wrong-task completion accounts for 17.5\%, 17.0\%, and 14.6\% of target-task failures overall, respectively.
Thus, a non-negligible fraction of switching failures do not simply end without completing a task; instead, the robot completes a task other than the switched target.

For several policies, wrong-task completion also becomes more frequent at later switching stages.
The trend is particularly pronounced for $\pi_0$, for which wrong-task completion rises from 0.0\% of target-task failures in $[3,9]$ to 31.4\% in $[21,27]$.
OpenVLA-OFT shows a similar increase from 11.8\% to 22.6\%.
This pattern is consistent with later switching states lying further along the source-task behavior, where redirecting execution toward a new target becomes more challenging.

The frequency of complete wrong-task execution varies substantially across policies, with Fast-WAM and LingBot-VA exhibiting this outcome less frequently.
Importantly, a low wrong-task completion rate does not imply that the remaining switching failures are unrelated to trajectory hijacking.
Our definition of trajectory hijacking also includes partial drift toward a non-target trajectory, which may disrupt target-task execution without culminating in completion of another task.
Overall, these results provide quantitative evidence that complete execution of a non-target task is a recurring failure pattern in mid-execution instruction switching, complementing the qualitative examples in Fig.~\ref{fig:obs3}.

\subsection{\rev{Where Do Task Distinctions Weaken during Action Generation?}}
\label{app:representation_analysis}
\begingroup

Observation~3 suggests that task distinctions can weaken as the gripper approaches an object. To locate where this attenuation occurs, we trace task-conditioned representations inside $\pi_{0.5}$, from the language backbone through the action expert to the output. We ask where inside the policy different tasks become less distinguishable as execution approaches the source object.

Our main comparison follows the same wine-source trajectory at two stages. At the early state ($t=10$), the gripper is still approaching the bottle and different instructions produce clearly distinct action directions. At the later state ($t=25$), the gripper is close to the bottle and the action directions become less distinguishable across instructions. At each state, we hold the visual observation, robot state, and initial action noise fixed while changing only the instruction. We then compare task separation between the two states within each representation space, using the relative separation $D$ defined below.

\paragraph{Task differences persist throughout the language backbone.}
Even when the gripper is close to the bottle and the action directions become less distinguishable, the language backbone continues to produce distinct representations for the four tasks. Figure~\ref{fig:representation_trace}a shows that these differences persist across all 18 layers. At the backbone output, the late state's object-token separation remains 95\% of its early-state value (Table~\ref{tab:representation_trace}).

The shared final newline token provides a complementary view. It has the same input embedding under all four instructions, but develops different contextual representations as the instructions are processed. Its late-state separation retains 90\% of the early-state value. Thus, the near-bottle state still preserves task-dependent language representations, even though the action-level distinctions have been weakened.

\begin{figure}[t]
  \centering
  \includegraphics[width=\linewidth]{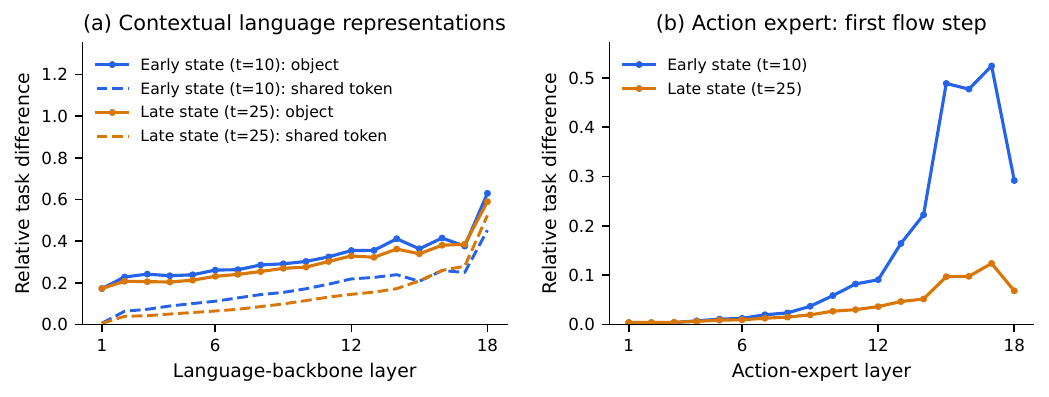}
\caption{\rev{\textbf{Task differences persist in the language backbone but are weaker in the action expert near the source object.} We compare early ($t=10$) and late ($t=25$) states on the same wine-source trajectory. (a) Task separation persists across the language backbone: solid lines show object-token averages and dashed lines the shared final text token. (b) At the first flow update, all instructions start from identical action noise. Task separation grows through the expert in both states, but much less strongly in the late state. Axes show relative separation $D$ (Eq.~\ref{eq:representation_separation}) with different scales. Curves precede final normalization; Table~\ref{tab:representation_trace} reports module outputs after normalization.}}
  \label{fig:representation_trace}
\end{figure}

\paragraph{Task differences become much weaker in the action expert.}
We next follow the same four instructions into the action expert. Here, the early and near-bottle states differ sharply. The early state develops clear task separation during action generation, whereas separation is much weaker in the near-bottle state (Fig.~\ref{fig:representation_trace}b). This contrast is already present at the first flow update, when all instructions begin from identical action noise, and persists through the subsequent updates. The difference remains visible at the final flow update (Table~\ref{tab:representation_trace}).

The attenuation continues from the action expert toward the output: task-conditioned flow velocities and first actions retain less separation in the late state. A second trajectory shows the same qualitative pattern: language object-token separation remains comparatively stable, while separation in the action-generation pathway becomes weaker. These results localize the loss of task distinction to action generation: task information persists throughout the language backbone, but is expressed less strongly in the action expert near the source object.

\begin{table}[t]
  \centering
  \small
  \caption{\rev{\textbf{Task separation along the language-to-action computation.} Values are $D$ for the early/late state pair in Fig.~\ref{fig:representation_trace}. Ratios compare states within each row. Language and expert measurements follow final normalization.}}
  \label{tab:representation_trace}
  \begin{tabular}{lrrr}
    \toprule
    Representation / output & Early ($t=10$) & Late ($t=25$) & Late / early \\
    \midrule
    Language: object tokens & 0.782 & 0.740 & 0.946 \\
    Language: shared final token & 0.551 & 0.497 & 0.903 \\
    Action expert: flow step 1 & 0.250 & 0.039 & 0.157 \\
    Action expert: flow step 10 & 0.737 & 0.141 & 0.191 \\
    \bottomrule
  \end{tabular}
\end{table}

\paragraph{Measurement and scope.}
We trace all 18 backbone layers and all 18 expert layers across ten flow updates, using eight states chosen from a 72-observation screen before inspecting hidden representations. The screen spans 12 source trajectories and two noise seeds; tracing uses one shared seed. Object-token vectors are averaged over each instruction's manipulated-object span. Expert vectors concatenate the first five action positions; velocity vectors concatenate their xyz components. Final actions use only the first xyz prediction. For four task-conditioned vectors $H=\{h_1,\ldots,h_4\}$, relative separation is
\begin{equation}
\label{eq:representation_separation}
D(H)=\frac{1}{6}\sum_{i<j}
\frac{\operatorname{RMS}(h_i-h_j)}
{\sqrt{\bigl(\operatorname{RMS}(h_i)^2+\operatorname{RMS}(h_j)^2\bigr)/2}},
\end{equation}
where $\operatorname{RMS}$ averages over vector coordinates. Smaller $D$ means weaker geometric separation. These are fixed-observation comparisons of task-conditioned predictions; the late/early ratios quantify how task-conditioned representations become less separated across execution states. The pattern motivates CAS: rather than requiring the base policy to produce fully separated actions, CAS uses the weakened residual distinction between target-conditioned and counterfactual actions and amplifies their difference into a more task-specific steering direction.

\paragraph{Implications for CAS and CAS-ST.}
These observations suggest that task distinctions can be weakened during action generation without being entirely lost. Importantly, even when task-conditioned actions are influenced by the current trajectory, their relative differences can remain informative (Fig.~\ref{fig:obs3}(d)). CAS builds on this insight: by contrasting the target-conditioned action with counterfactual actions from potentially interfering tasks, it explicitly amplifies the remaining task-dependent differences to strengthen redirection toward the switched target. Rather than modifying the policy's internal representations, CAS leverages the distinctions already reflected in its action predictions to counteract interfering trajectory priors. Beyond inference-time steering, CAS-guided self-training (CAS-ST) further enables the policy to learn more distinguishable task-conditioned actions from successful de-hijacking trajectories. Through iterative self-training, these distinctions can be progressively strengthened and internalized into the policy, as illustrated by the Effect of CAS-ST in Fig.~\ref{fig:overview}(c).

\endgroup
\FloatBarrier

\subsection{\rev{Local Positional Robustness of Trajectory Priors}}
\label{app:local_prior_robustness}
\begingroup
\raggedbottom

When the gripper is close to a drawer handle, demonstrations typically continue with contact and a pulling motion. A policy trained on these regularities can learn to favor that continuation from the current visual state. If this preference persists after the instruction changes, it contributes to trajectory hijacking. The same tendency could also help complete the intended interaction after redirection, provided it tolerates some variation in where the gripper arrives.

\begin{samepage}
Local displacements let us examine this possibility while keeping the robot in the same task-associated neighborhood. After the released OpenVLA-OFT policy takes 25 actions toward one of three LIBERO-Goal tasks, we shift the gripper by up to 3\,cm and let it replan without CAS. We use CAS's KNN distance to delimit the neighborhood, retaining physically valid shifts only when the source trajectory remains closest among the three task libraries. Keeping the original instruction tests whether the interaction remains recoverable; replacing it with ``remain still'' tests whether it persists when the instruction no longer requests it.
\par
\end{samepage}

\begin{table}[t]
  \centering
  \small
  \caption{\rev{\textbf{Local positional robustness and trajectory proximity.} Original-task completion (\%; 50 trials per cell) after random end-effector translations of magnitude $r$, with orientation preserved. Tasks are opening the middle drawer (0), placing cream cheese in the bowl (6), and turning on the stove (7). ``Remain still'' is the literal post-prefix instruction. KNN distances ($K=30$) are medians over original-instruction trials, in cm; bold marks the source task.}}
  \label{tab:local_prior_robustness}
  \setlength{\tabcolsep}{6pt}
  \renewcommand{\arraystretch}{1.05}
  \begin{tabular}{ccrrrrr}
    \toprule
    & & \multicolumn{2}{c}{Original-task completion (\%)} & \multicolumn{3}{c}{KNN distance (cm)} \\
    \cmidrule(lr){3-4}\cmidrule(lr){5-7}
    Task & $r$ (cm) & Original & Remain still & $d_0$ & $d_6$ & $d_7$ \\
    \midrule
    \multirow{4}{*}{0} & 0 & 94 & 98 & \textbf{1.11} & 6.49 & 11.80 \\
      & 1 & 100 & 96 & \textbf{1.11} & 6.47 & 11.57 \\
      & 2 & 92 & 98 & \textbf{1.18} & 7.26 & 11.99 \\
      & 3 & 100 & 92 & \textbf{1.30} & 6.40 & 11.68 \\
    \midrule
    \multirow{4}{*}{6} & 0 & 96 & 90 & 10.55 & \textbf{1.31} & 17.77 \\
      & 1 & 92 & 92 & 10.62 & \textbf{1.33} & 17.60 \\
      & 2 & 94 & 92 & 10.17 & \textbf{1.50} & 17.85 \\
      & 3 & 96 & 88 & 10.62 & \textbf{1.54} & 17.69 \\
    \midrule
    \multirow{4}{*}{7} & 0 & 100 & 100 & 18.38 & 16.10 & \textbf{1.24} \\
      & 1 & 100 & 100 & 18.43 & 15.86 & \textbf{1.40} \\
      & 2 & 100 & 100 & 18.50 & 15.99 & \textbf{1.53} \\
      & 3 & 100 & 100 & 18.57 & 16.30 & \textbf{1.96} \\
    \bottomrule
  \end{tabular}
\end{table}

The interaction remains highly recoverable: completion stays at 92--100\% under the original instruction across the tested radii (Table~\ref{tab:local_prior_robustness}). More strikingly, original-task completion remains at 88--100\% after switching to ``remain still''. For drawer opening, the policy completes 92\% of trials after a 3\,cm shift under ``remain still'', compared with 98\% without perturbation. It is failing to stop, yet still succeeding at the original interaction. This is consistent with trajectory hijacking, while showing that the persistent behavior survives a range of local positions rather than depending on one exact gripper pose.

The KNN measurements locate this behavior relative to demonstrations. Across the reported conditions, median distances to the source task remain within 1.11--1.96\,cm, while those to either alternative task are at least 6.40\,cm. This proximity to demonstrated states is the lightweight proxy for trajectory-prior strength used by CAS (Sec.~\ref{sec:inference_time_cas}): it indicates which task has local geometric support, without directly measuring the policy's internal prior. Here, KNN defines the neighborhood in which perturbations are sampled; the rollouts show that the behavior remains robust within it.

These results suggest that CAS can exploit the same behavioral persistence that makes instruction switching difficult. Its role is to redirect execution away from an interfering trajectory and toward states compatible with the intended task. Within that region, the underlying policy can use its learned behavior and visual feedback to refine positioning and complete the interaction. Thus, effective redirection may allow some error in the precise gripper position. This local tolerance offers an explanation for the broad effective range of $\lambda$ observed in Appendix~\ref{app:hyper}: different steering magnitudes may still reach states from which the policy can complete the intended task.

\FloatBarrier
\clearpage
\endgroup

\section{Additional Experiments and Analyses of Inference-Time CAS}

\subsection{Inference-Time Efficiency Analysis}
\label{app:efficiency}

We further analyze the inference-time efficiency of CAS on CALVIN-Custom-Switch using $\pi_0$ for $t_s \in [15,21]$. We compare the base policy, CAS, and PCD over the same 240 evaluation episodes on a single NVIDIA H100 GPU. All methods use the same maximum episode length of 180 environment actions.

\begin{table}[t]
\centering
\caption{
\textbf{Inference-time efficiency on CALVIN-Custom-Switch with $\pi_0$.}
We compare the base policy, CAS, and PCD over the same 240 episodes for $t_s \in [15,21]$ on a single NVIDIA H100 GPU, with a maximum episode length of 180 environment actions.
CAS is evaluated using serial counterfactual policy queries.
\emph{Avg. Inference Latency per Replan} sums the wall-clock time of all policy queries issued within one replan.
Values in parentheses denote the inference latency relative to the base policy.
}
\label{tab:inference_efficiency}
\begin{tabular}{lcc}
\toprule
Method & Avg. Episode Time & Avg. Inference Latency per Replan \\
\midrule
Base  & 14.14 s & 102.2 ms (1.00$\times$) \\
+ CAS & 12.13 s & 157.4 ms (1.54$\times$) \\
+ PCD & 38.15 s & 788.3 ms (7.71$\times$) \\
\bottomrule
\end{tabular}
\end{table}

\paragraph{Efficiency of CAS.}
CAS may query the policy under multiple task instructions at the same state, including the target instruction and the instructions of selected hijacking tasks. These instruction-conditioned action predictions are independent and can therefore be issued in a single batched policy call. In principle, such batching can substantially reduce the additional sequential policy-query latency introduced by counterfactual queries. In our current implementation, however, we use a simpler serial implementation and query the policy separately for different task instructions.

Even without such parallelization, CAS introduces relatively modest additional policy-query cost. As shown in Table~\ref{tab:inference_efficiency}, CAS has a lower average episode time, 12.13\,s compared with 14.14\,s for the base policy. Since all methods use the same maximum episode length of 180 environment actions, this lower average episode time is consistent with CAS more frequently completing the target task before reaching the episode horizon.

The reported per-replan timing measures the accumulated latency of all policy queries within each replan, including the base target-conditioned query and any additional counterfactual queries. We therefore report this metric as \emph{Avg. Inference Latency per Replan}. CAS increases the average inference latency per replan from 102.2\,ms for the base policy to 157.4\,ms, corresponding to only $1.54\times$ the base latency. In comparison, PCD has an average inference latency of 788.3\,ms per replan, or $7.71\times$ the base latency.

Across all CAS replans in this evaluation, CAS makes 1.60 policy queries on average, including the target-conditioned query. Equivalently, in addition to the single target-conditioned query required by the base policy, CAS introduces only 0.60 counterfactual policy queries per replan on average. This low query count results from the selective nature of CAS: rather than querying every non-target task, CAS only queries tasks whose trajectory distributions are sufficiently close to the current state and thus exert strong interfering priors.

\paragraph{Scaling to larger task sets.}
One potential concern is that, under a serial implementation, the policy-query cost of CAS could increase if many counterfactual tasks need to be queried. Importantly, however, the number of counterfactual policy queries is determined by the number of locally interfering trajectories rather than directly by the total number of tasks. Even as the overall task set grows, CAS only queries non-target tasks whose KNN distance to the current state falls below the selection threshold. Therefore, the number of counterfactual queries need not grow proportionally with the size of the task set.

In settings where many trajectories strongly overlap around the current state, the computational budget can be further controlled by reducing the KNN distance threshold $\delta$, thereby retaining only the strongest interfering trajectory priors. The selected counterfactual queries can also be evaluated in parallel through batched policy inference, as discussed above. Finally, the lightweight Source-only CAS variant introduced in Sec.~\ref{sec:inference_time_cas} uses only the known source task and avoids querying additional candidate tasks, while still providing substantial improvements (Sec.~\ref{sec:ablations}).

\subsection{Generality across Diverse Trajectory Regimes}
\label{app:diverse_trajectory_regimes}

The strength and structure of trajectory priors may depend on the object-pose distribution used to generate demonstrations.
When object poses vary only slightly around fixed nominal locations, demonstrations tend to form narrow and highly concentrated trajectory distributions.
As object-pose variation increases, the resulting trajectories become more dispersed, potentially changing the interference patterns among tasks.
A qualitatively different setting arises when object positions are sampled over a broad workspace region without fixed nominal centers, where the resulting task trajectories no longer follow the same object-centered spatial structure.

\begin{figure*}[t]
    \centering
    \begin{subfigure}[t]{0.24\textwidth}
        \centering
        \includegraphics[width=\linewidth]{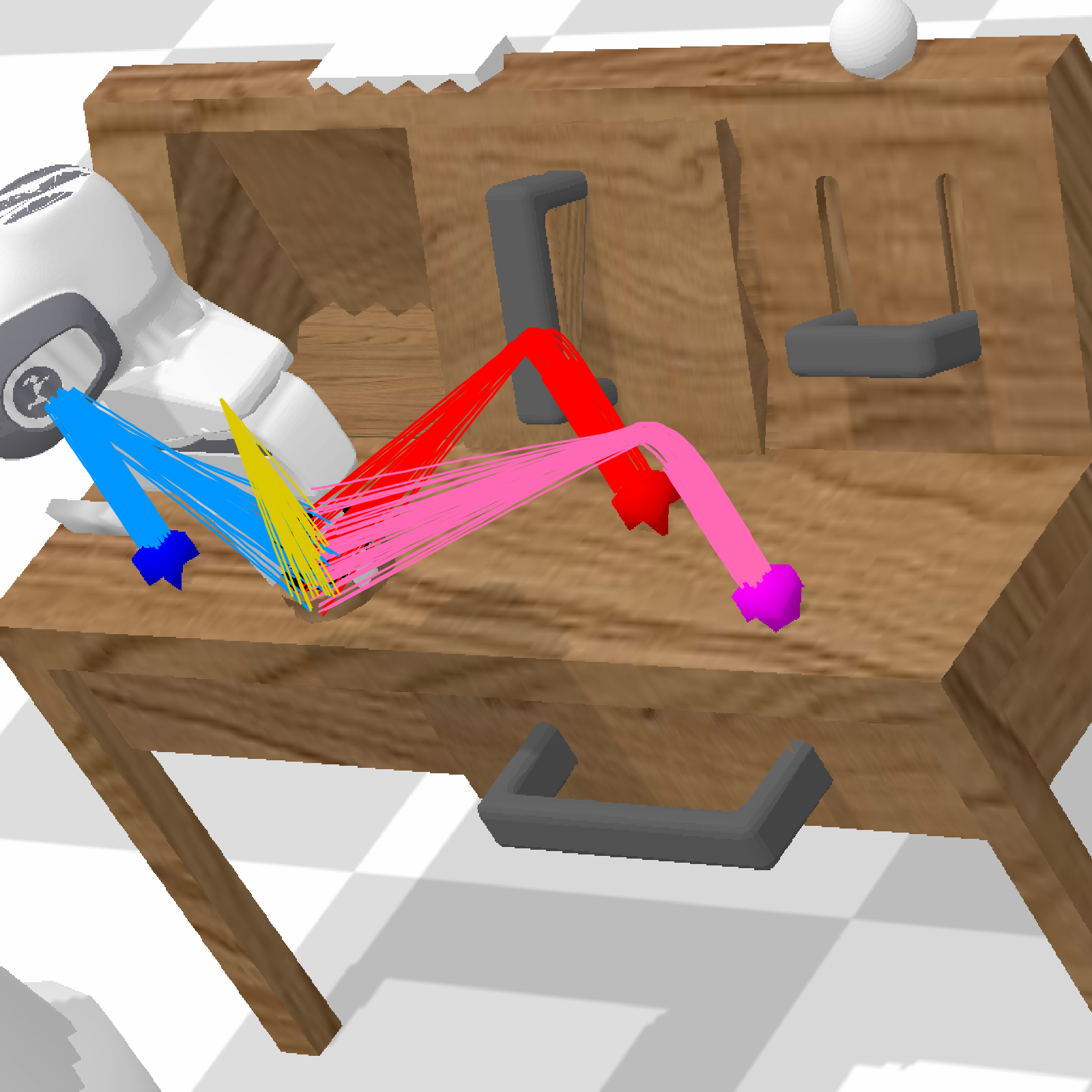}
        \caption{\textbf{Narrow}}
        \label{fig:traj_regime_narrow}
    \end{subfigure}
    \hfill
    \begin{subfigure}[t]{0.24\textwidth}
        \centering
        \includegraphics[width=\linewidth]{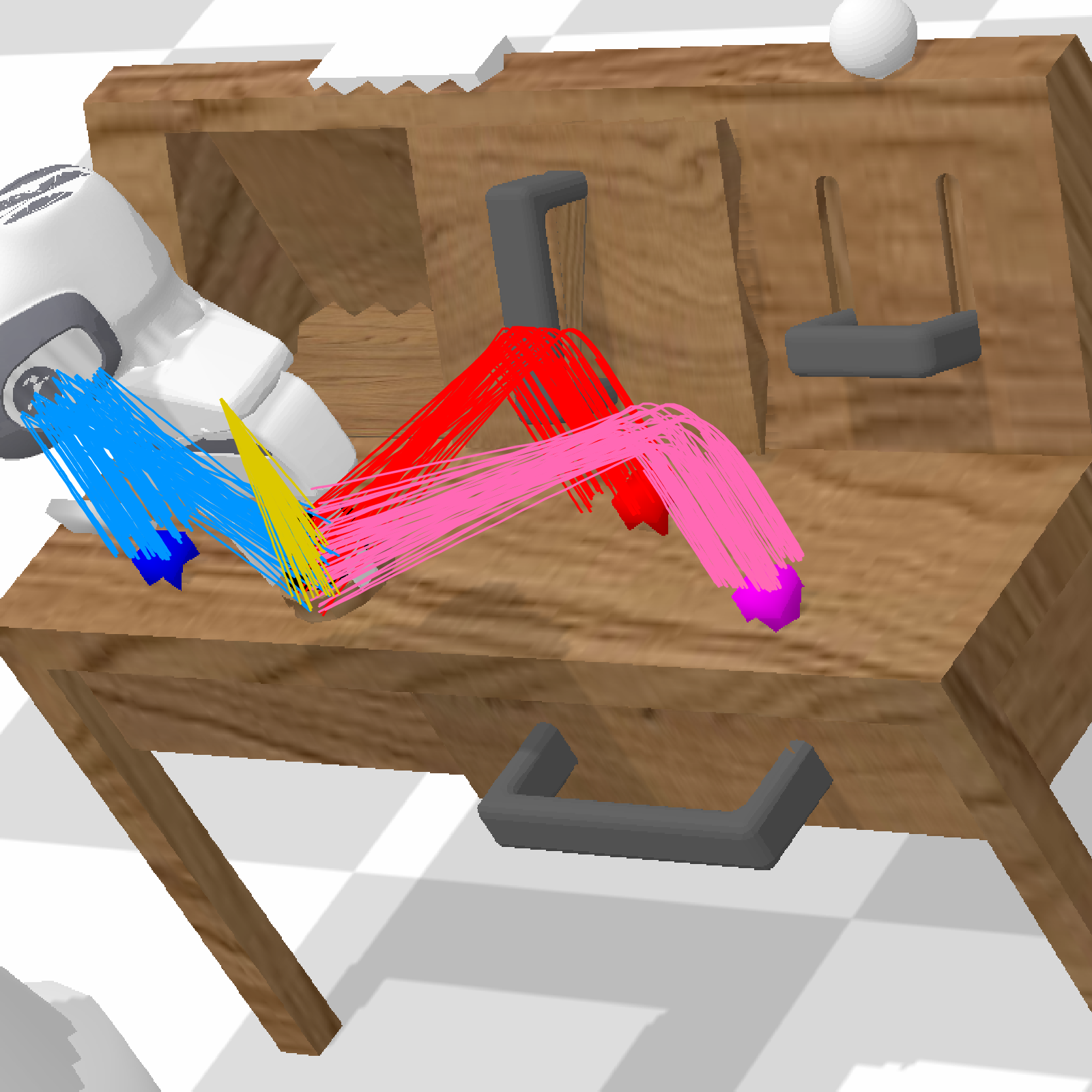}
        \caption{\textbf{Medium}}
        \label{fig:traj_regime_medium}
    \end{subfigure}
    \hfill
    \begin{subfigure}[t]{0.24\textwidth}
        \centering
        \includegraphics[width=\linewidth]{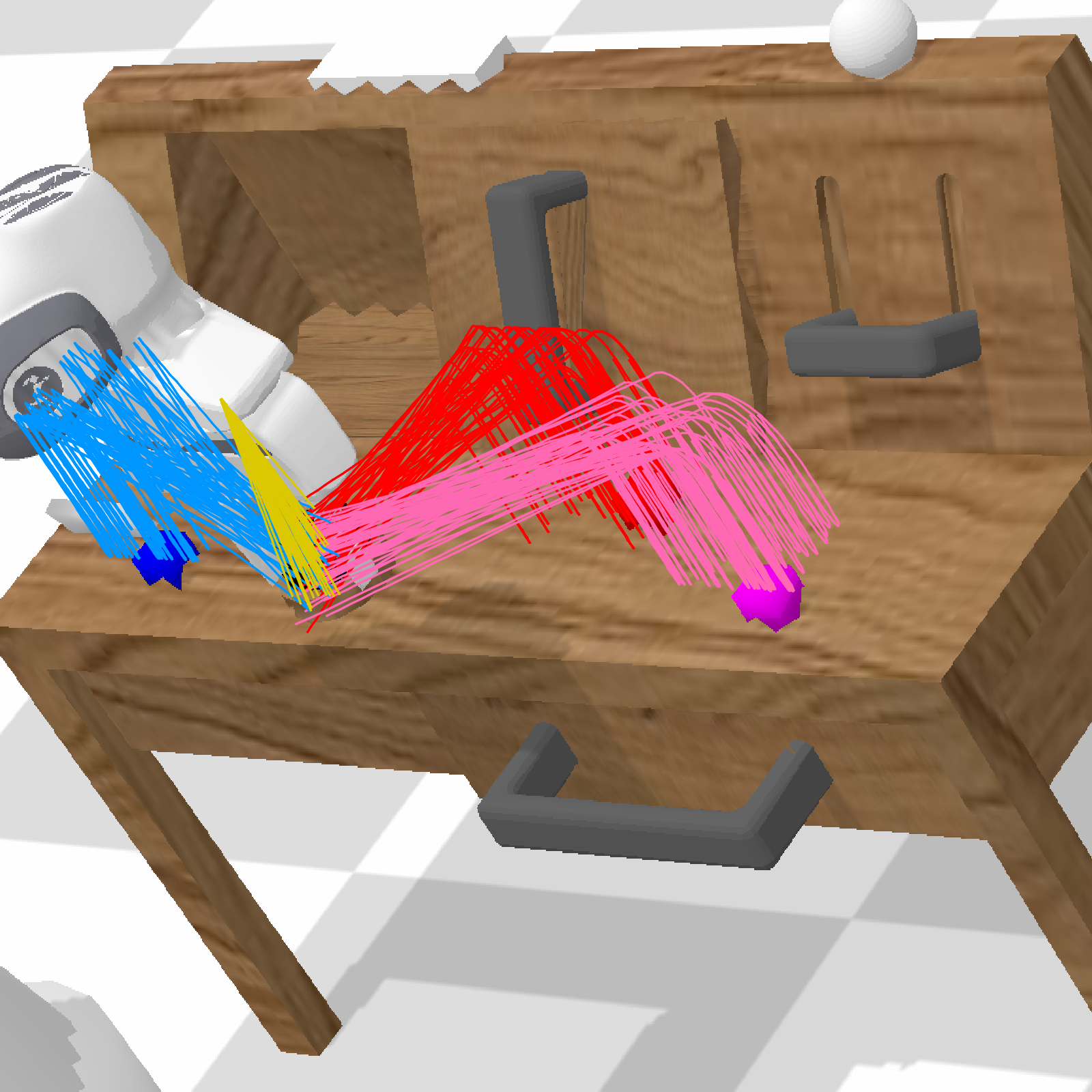}
        \caption{\textbf{Wide}}
        \label{fig:traj_regime_wide}
    \end{subfigure}
    \hfill
    \begin{subfigure}[t]{0.24\textwidth}
        \centering
        \includegraphics[width=\linewidth]{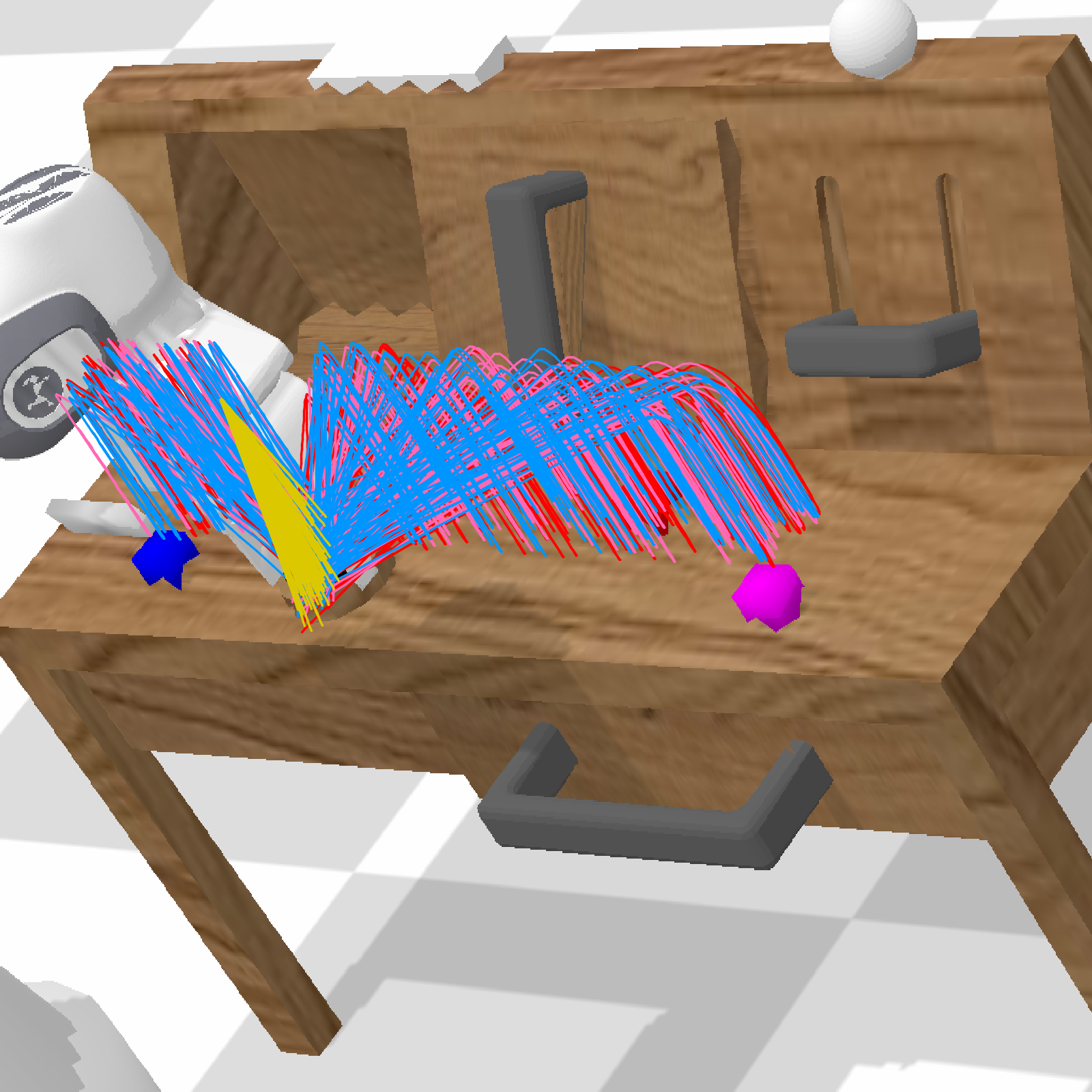}
        \caption{\textbf{Random}}
        \label{fig:traj_regime_random}
    \end{subfigure}
    \caption{
    Visualization of task trajectory distributions under different object-pose regimes in CALVIN-Custom-Switch.
    In \emph{narrow}, \emph{medium}, and \emph{wide}, object poses are sampled as perturbations around fixed nominal centers, with progressively larger perturbation ranges producing increasingly dispersed trajectories.
    The \emph{random} regime instead samples object positions over a broader workspace region without fixed nominal centers, resulting in a qualitatively different trajectory structure.
    }
    \label{fig:traj_regimes}
\end{figure*}

To evaluate whether CAS depends on a particular trajectory geometry, we consider two complementary variations of CALVIN-Custom-Switch.
First, we construct three concentration-controlled regimes, \emph{narrow}, \emph{medium}, and \emph{wide}, by progressively increasing the object-pose perturbation range around fixed nominal centers.
Second, we construct a \emph{random} regime in which object positions are sampled over a broader workspace region without fixed nominal centers.
For each regime, the object-pose distribution used for training is aligned with that used for evaluation, so these experiments isolate the effect of demonstration-trajectory structure rather than train--test distribution shift.
Fig.~\ref{fig:traj_regimes} visualizes the resulting trajectory distributions.

\subsubsection{Varying Trajectory Concentration}
\label{app:trajectory_concentration}

We first study whether CAS depends on narrow and highly concentrated trajectory distributions.
The \emph{narrow}, \emph{medium}, and \emph{wide} settings share the same nominal object locations but use progressively larger object-pose perturbation ranges.
The \emph{narrow} setting corresponds to CALVIN-Custom-Switch-Fixed used in the main experiments and uses the smallest perturbation range.
The \emph{medium} and \emph{wide} settings introduce increasingly larger object-pose variation, producing progressively more dispersed task trajectories, as illustrated in Fig.~\ref{fig:traj_regimes}~(a)--(c).

\begin{table}[t]
\centering
\caption{
Success rates (\%) of $\pi_0$ on CALVIN-Custom-Switch under different levels of trajectory concentration.
The \emph{narrow} setting corresponds to CALVIN-Custom-Switch-Fixed used in the main experiments.
}
\label{tab:trajectory_concentration}
\setlength{\tabcolsep}{7pt}
\renewcommand{\arraystretch}{1.12}
\begin{adjustbox}{max width=\linewidth,center}
\begin{tabular}{llccc}
\toprule
\multirow{2}{*}{Switch range}
& \multirow{2}{*}{Method}
& \multicolumn{3}{c}{Trajectory concentration} \\
\cmidrule(lr){3-5}
& & Narrow & Medium & Wide \\
\midrule
\multirow{2}{*}{$[9,15]$}
& $\pi_0$
& 22.1 & 12.9 & 12.1 \\
& $\pi_0$+CAS
& \textbf{73.8} & \textbf{62.9} & \textbf{65.8} \\
\midrule
\multirow{2}{*}{$[15,21]$}
& $\pi_0$
& 6.7 & 2.1 & 5.4 \\
& $\pi_0$+CAS
& \textbf{45.4} & \textbf{40.8} & \textbf{43.8} \\
\midrule
\multirow{2}{*}{$[21,27]$}
& $\pi_0$
& 0.0 & 0.0 & 0.0 \\
& $\pi_0$+CAS
& \textbf{20.4} & \textbf{8.8} & \textbf{10.4} \\
\bottomrule
\end{tabular}
\end{adjustbox}
\vspace{-0.5em}
\end{table}

Table~\ref{tab:trajectory_concentration} reports the results across the three trajectory-concentration regimes.
CAS consistently improves switching success across all concentration levels and switching ranges.
For early switches in the $[9,15]$ range, CAS improves success from $22.1\%$ to $73.8\%$ in the narrow setting, from $12.9\%$ to $62.9\%$ in the medium setting, and from $12.1\%$ to $65.8\%$ in the wide setting.
The gains remain substantial for later and more challenging switches.
In the $[15,21]$ range, CAS improves success by $38.7$, $38.7$, and $38.4$ percentage points in the narrow, medium, and wide settings, respectively.
In the latest $[21,27]$ range, the base policy fails across all three regimes with $0.0\%$ success, whereas CAS still achieves $20.4\%$, $8.8\%$, and $10.4\%$ success.

These results show that the effectiveness of CAS is not restricted to highly concentrated trajectory distributions.
CAS remains effective as demonstrations become increasingly dispersed, suggesting that its benefit does not depend on a particular level of trajectory concentration.

\subsubsection{Randomized Object Placements}
\label{app:random_object_placement}

We next consider a qualitatively different trajectory regime in which object positions are sampled over a broader workspace region without fixed nominal centers.
Unlike the concentration-controlled settings above, which preserve task-specific nominal object locations while varying only the perturbation range, the \emph{random} regime substantially changes the spatial organization of the resulting task trajectories, as illustrated in Fig.~\ref{fig:traj_regimes}~(d).
This setting provides a stronger test of whether CAS depends on object-centered spatial regularities or a particular trajectory geometry.

\begin{table}[t]
\centering
\caption{
Success rates (\%) under the random trajectory regime on CALVIN-Custom-Switch-Random.
Object positions are randomized over a broader workspace region rather than sampled as perturbations around nominal centers.
}
\label{tab:random_object_placement}
\setlength{\tabcolsep}{5.5pt}
\begin{adjustbox}{max width=0.72\linewidth,center}
\begin{tabular}{llcc}
\toprule
Switch range & Policy & Base & CAS \\
\midrule
\multirow{5}{*}{$[9,15]$}
& OFT          & 1.7  & \textbf{4.6}  \\
& $\pi_0$      & 33.3 & \textbf{60.0} \\
& $\pi_{0.5}$  & 44.6 & \textbf{65.4} \\
& Fast-WAM     & 38.7 & \textbf{52.9} \\
& LingBot-VA   & 33.7 & \textbf{34.5} \\
\midrule
\multirow{5}{*}{$[15,21]$}
& OFT          & 1.3  & \textbf{2.1}  \\
& $\pi_0$      & 8.8  & \textbf{34.2} \\
& $\pi_{0.5}$  & 22.5 & \textbf{50.4} \\
& Fast-WAM     & 19.1 & \textbf{38.7} \\
& LingBot-VA   & 10.4 & \textbf{13.7} \\
\midrule
\multirow{5}{*}{$[21,27]$}
& OFT          & 1.3  & \textbf{1.7}  \\
& $\pi_0$      & 2.1  & \textbf{14.2} \\
& $\pi_{0.5}$  & 6.3  & \textbf{37.1} \\
& Fast-WAM     & 4.5  & \textbf{22.0} \\
& LingBot-VA   & 1.2  & \textbf{7.0}  \\
\bottomrule
\end{tabular}
\end{adjustbox}
\end{table}

Table~\ref{tab:random_object_placement} reports the results under the random trajectory regime across five policy architectures.
CAS improves switching success over the corresponding base policy for every evaluated policy and switching range.

For $\pi_0$, CAS improves success from $33.3\%$ to $60.0\%$ in the $[9,15]$ range, from $8.8\%$ to $34.2\%$ in the $[15,21]$ range, and from $2.1\%$ to $14.2\%$ in the $[21,27]$ range.
For $\pi_{0.5}$, the corresponding improvements across the three ranges are from $44.6\%$ to $65.4\%$, from $22.5\%$ to $50.4\%$, and from $6.3\%$ to $37.1\%$, respectively.
CAS also consistently improves Fast-WAM, increasing success from $38.7\%$ to $52.9\%$, from $19.1\%$ to $38.7\%$, and from $4.5\%$ to $22.0\%$ across the three progressively later switching ranges.
For LingBot-VA, the gain is modest in the $[9,15]$ range, from $33.7\%$ to $34.5\%$, but becomes more pronounced in the later ranges, improving from $10.4\%$ to $13.7\%$ in $[15,21]$ and from $1.2\%$ to $7.0\%$ in $[21,27]$.
Although the absolute improvements for OFT are smaller, CAS still improves over the corresponding baseline in all three switching ranges.

These results show that CAS does not rely on task trajectories being organized around fixed nominal object locations.
Even when object placements are broadly randomized and the resulting trajectory geometry changes substantially, CAS consistently improves switching performance across multiple policy architectures.
Together with the trajectory-concentration study above, these findings demonstrate that CAS remains effective across both increasingly dispersed trajectory distributions and qualitatively different randomized trajectory structures.

\subsection{Robustness to Scene Appearance Shifts}
\label{app:scene_appearance_shift}

Visual trajectory priors may be affected by the visual consistency between training and evaluation environments.
In our main experiments, the scene appearance at evaluation time matches that used during training.
To test whether CAS remains effective under visual appearance shifts, we construct a separate scene-appearance-shift evaluation using $\pi_0$.

\begin{figure}[t]
    \centering
    \begin{subfigure}[b]{0.24\linewidth}
        \centering
        \includegraphics[width=\linewidth]{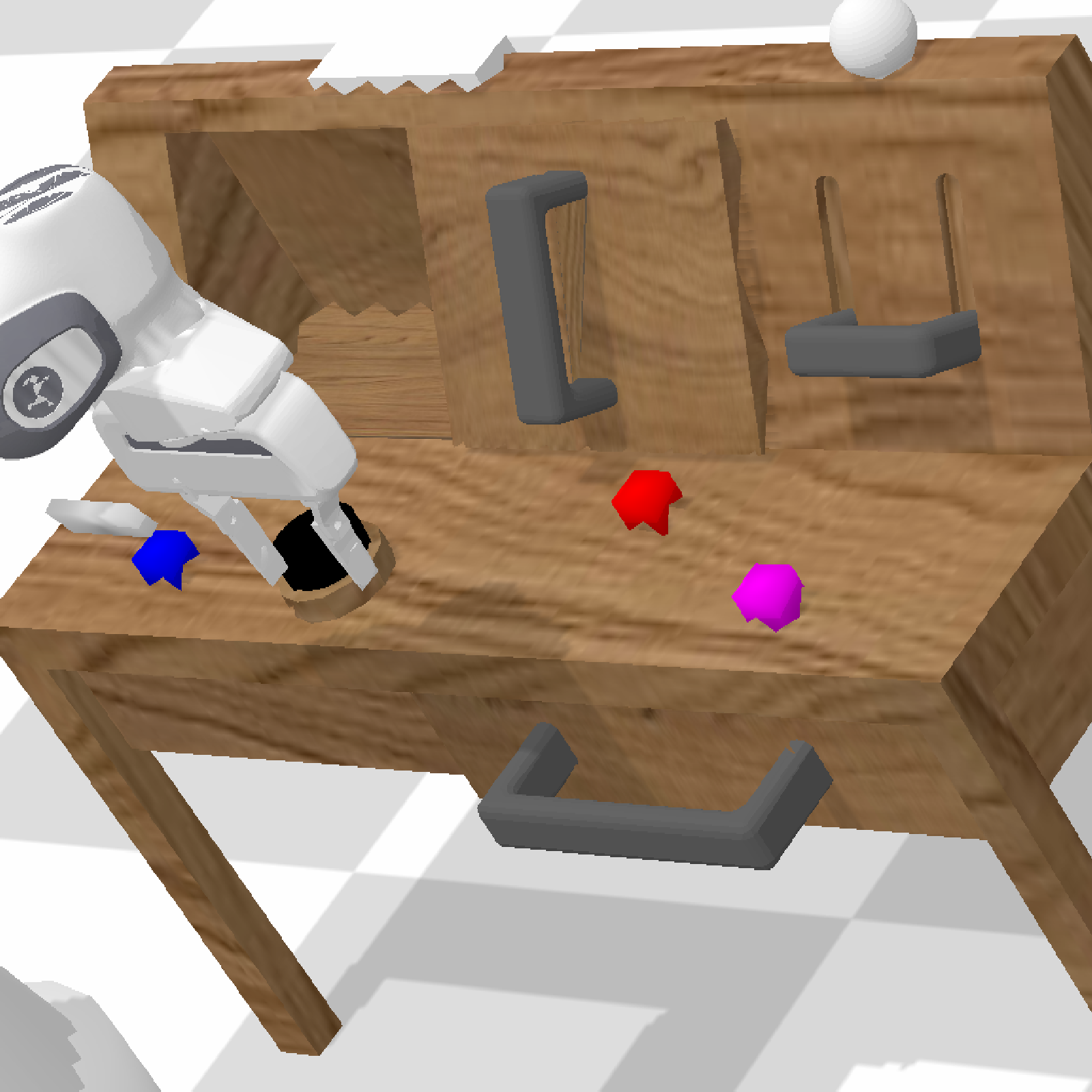}
        \caption{D0 (original)}
        \label{fig:scene_D0}
    \end{subfigure}
    \hfill
    \begin{subfigure}[b]{0.24\linewidth}
        \centering
        \includegraphics[width=\linewidth]{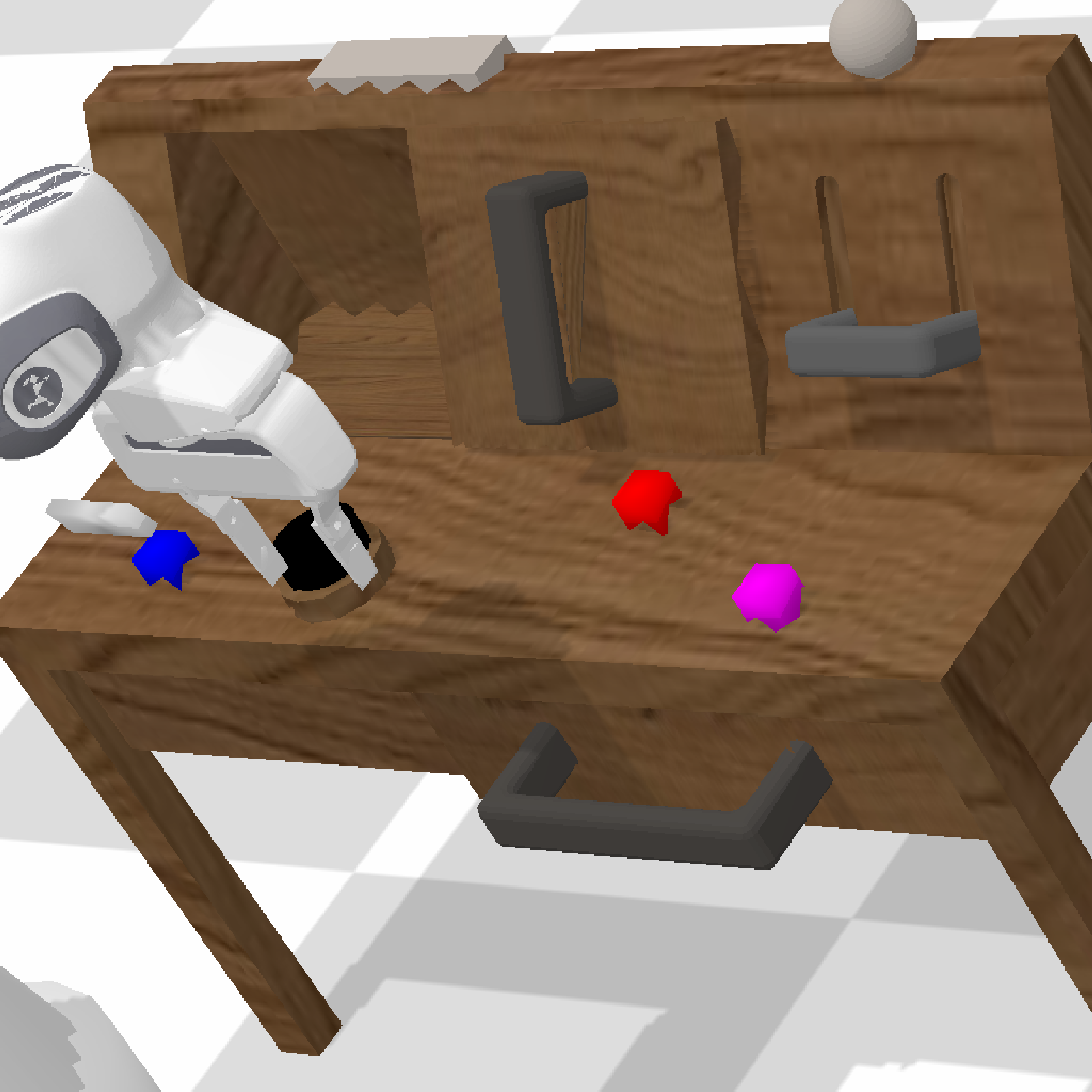}
        \caption{D1 (mild)}
        \label{fig:scene_D1}
    \end{subfigure}
    \hfill
    \begin{subfigure}[b]{0.24\linewidth}
        \centering
        \includegraphics[width=\linewidth]{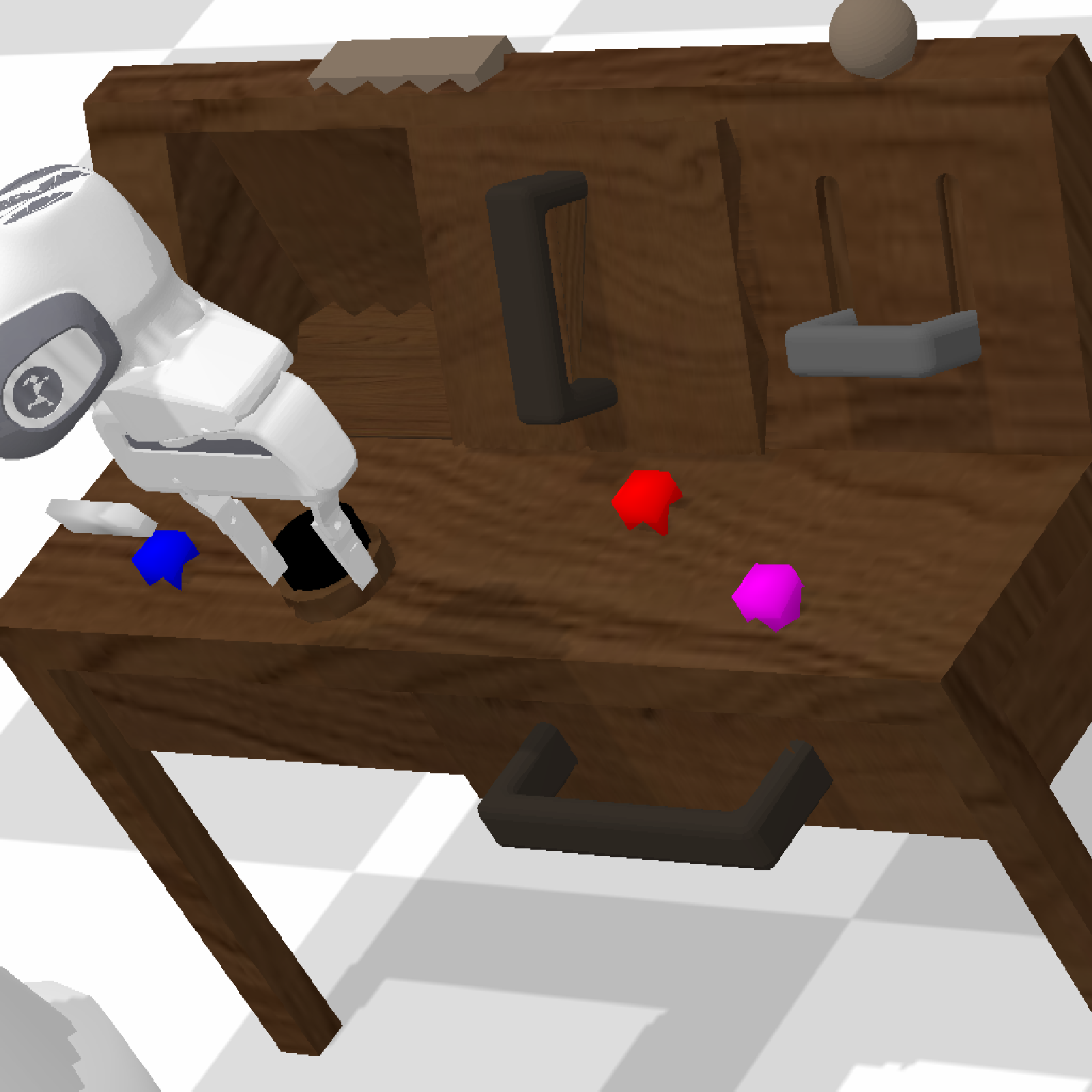}
        \caption{D2 (moderate)}
        \label{fig:scene_D2}
    \end{subfigure}
    \hfill
    \begin{subfigure}[b]{0.24\linewidth}
        \centering
        \includegraphics[width=\linewidth]{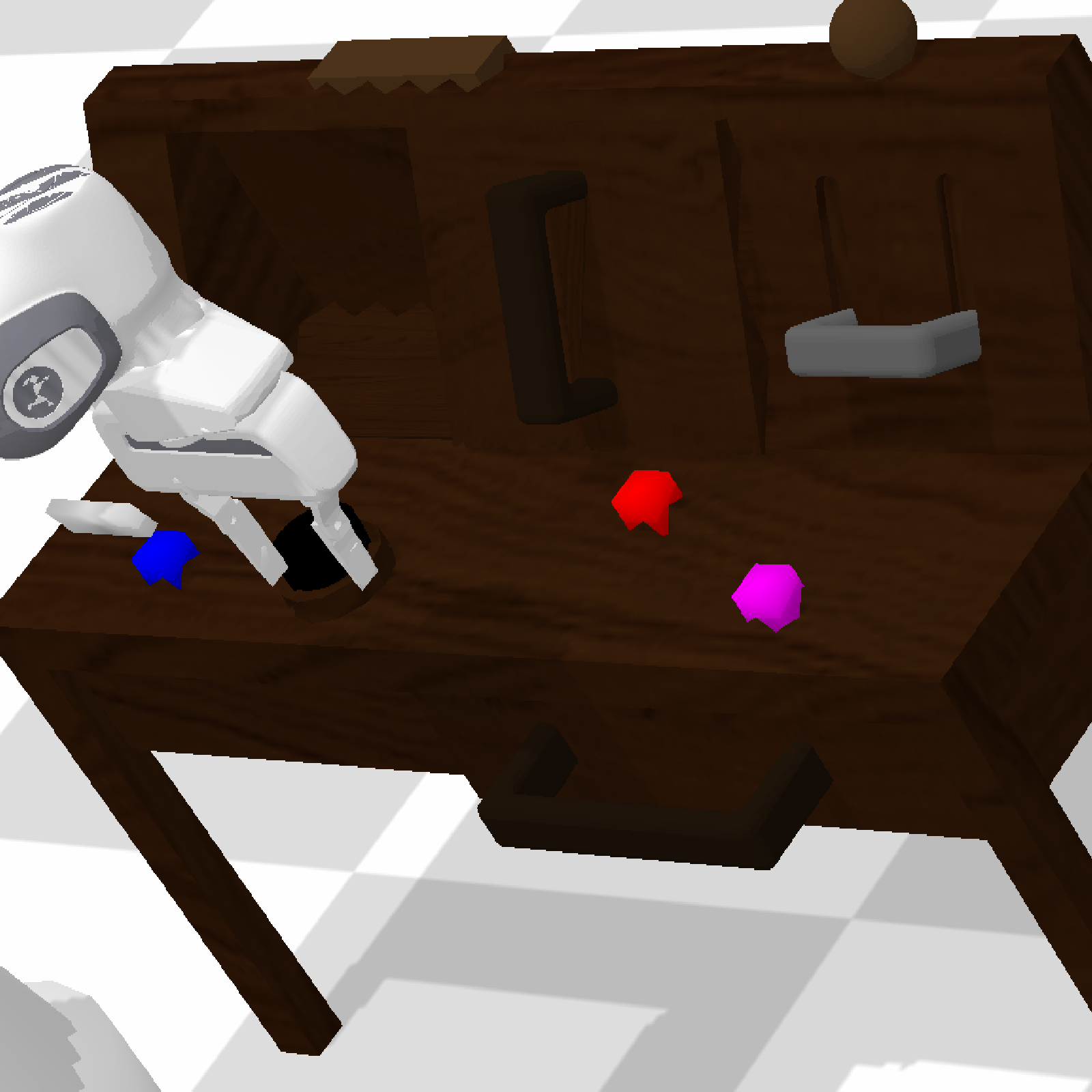}
        \caption{D3 (strong)}
        \label{fig:scene_D3}
    \end{subfigure}
    \caption{
    Scene appearance variants used for evaluating robustness to visual train--test mismatch.
    D0 is the original CALVIN-Custom-Switch scene used for training in our main experiments, while D1--D3 progressively darken the background appearance.
    The task semantics, object layout, and scene geometry remain unchanged across variants.
    }
    \label{fig:scene_shift_variants}
\end{figure}

As shown in Fig.~\ref{fig:scene_shift_variants}, we create four appearance variants based on the CALVIN-Custom-Switch scene.
We denote the original training scene as \emph{D0} and the progressively darkened variants as \emph{D1}--\emph{D3}.
All four variants preserve the same task semantics, object layout, and scene geometry; only the appearance of the wood-textured background surfaces is changed.
The policy is trained only on D0 and evaluated on all four variants.

To isolate the effect of scene appearance, we construct a separate set of state-matched switching episodes shared across D0--D3.
Switching states are sampled at normalized reference progress
$u \sim \mathcal{U}[0.4,0.6]$, corresponding to nominal reference steps $[12,18]$ on the same 30-step reference scale used in CALVIN-Custom-Switch.
For each scene variant, we evaluate all 12 ordered off-diagonal source--target task pairs with 50 episodes per pair, yielding 600 switching episodes per variant.
The same underlying robot states, object configurations, source--target assignments, and sampled reference-progress values are used across D0--D3; only the scene appearance differs.
Success is determined solely by completion of the target task, and we report the switch-only success rate over these 600 episodes.

\begin{table}[t]
\centering
\caption{
Switch-only success rates (\%) under scene appearance shifts for $\pi_0$.
Each variant contains 600 episodes covering 12 ordered off-diagonal source--target pairs with 50 repetitions per pair.
}
\label{tab:scene_shift_results}
\setlength{\tabcolsep}{7pt}
\begin{tabular}{lcc}
\toprule
Scene variant & $\pi_0$ & $\pi_0+\mathrm{CAS}$ \\
\midrule
D0 (original) & 13.2 & \textbf{55.0} \\
D1 (mild)     & 13.7 & \textbf{55.8} \\
D2 (moderate) & 14.5 & \textbf{55.2} \\
D3 (strong)   & 13.7 & \textbf{53.8} \\
\bottomrule
\end{tabular}
\end{table}

Table~\ref{tab:scene_shift_results} reports the results.
The base $\pi_0$ policy exhibits similar success rates across D0--D3, ranging from 13.2\% to 14.5\%.
CAS substantially improves switching performance in every scene variant, with success rates remaining highly stable between 53.8\% and 55.8\% across appearance-shift strengths.
Notably, CAS achieves 55.0\% on the original D0 scene and 53.8\% even under the strongest D3 shift.

These results show that the effectiveness of CAS does not rely on exact visual matching between the training and evaluation scenes.
Because the physical states, task assignments, and reference-progress samples are matched across D0--D3, the consistent improvements indicate that counterfactual action steering remains effective even when the visual appearance of the environment differs from that seen during training.

\subsection{Robustness Across Evaluation Random Seeds}
\label{app:random_seed}
\begin{table}[t]
\centering
\caption{
Robustness to inference random seeds on LIBERO-Goal-Switch.
We evaluate $\pi_{0.5}$ at $t_s=25$ over five random seeds.
Each seed contains the same 90 ordered source--target task pairs.
CAS consistently improves over the base policy across all evaluated seeds.
Seed 7 corresponds to the setting used in the main evaluation.
}
\label{tab:seed_robustness}
\begin{tabular}{lcccccc}
\toprule
Method & Seed 0 & Seed 1 & Seed 2 & Seed 3 & Seed 7 & Mean $\pm$ Std. \\
\midrule
Base
& 35.6 & 27.8 & 36.7 & 35.6 & 34.4 & 34.0 $\pm$ 3.6 \\
+ CAS
& 37.8 & 46.7 & 47.8 & 45.6 & 44.4 & \textbf{44.4 $\pm$ 3.9} \\
\bottomrule
\end{tabular}
\end{table}

We further evaluate whether the improvement from CAS is robust to randomness during policy inference. We focus on $\pi_{0.5}$ on LIBERO-Goal-Switch at the challenging switching step $t_s=25$. In addition to seed 7 used in the main evaluation, we repeat the same evaluation with four additional random seeds, $\{0,1,2,3\}$. For each seed, we evaluate the same 90 ordered source--target task pairs, while keeping all other evaluation and CAS settings unchanged.

As shown in Table~\ref{tab:seed_robustness}, CAS consistently outperforms the corresponding base policy across all five random seeds. Averaged over seeds $\{0,1,2,3,7\}$, the base policy achieves a success rate of $34.0\%$, while CAS achieves $44.4\%$, corresponding to an average improvement of $10.4$ percentage points. The improvement is observed for every evaluated seed, indicating that the benefit of CAS is not specific to the particular random seed used in the main evaluation.

\rev{
\subsection{Robustness to Initial States}
\label{app:initial_states}

\paragraph{Rationale for using a shared initial state.}
The main LIBERO-Goal-Switch evaluation starts every episode from one shared initial state (Appendix~\ref{app:libero_goal_switch_details}).
This provides a controlled comparison: all evaluations begin from the same robot configuration and object arrangement, which removes initial-state variation and focuses the evaluation on source--target instruction switching and the proposed steering method.
Diversity in switching scenarios is provided by the 90 ordered source--target pairs and four switching steps, since each switching state $V_{t_s}$ is generated online by executing the source task for $t_s$ steps, producing different intermediate configurations across source tasks and switching steps.
Moreover, official LIBERO-Goal initial states share the same scene layout and differ primarily in small placement perturbations.
Nevertheless, varying the initial state offers an additional dimension of evaluation diversity and a more direct assessment of robustness to initial-state selection, which we evaluate below.
}

\begin{table}[t]
\centering
\caption{
Target-task success rates (\%) of $\pi_0$ on LIBERO-Goal-Switch with a different, randomly drawn official initial state for each switching pair.
Within each pair, Base and CAS start from the same initial state with the same model seed.
}
\label{tab:random_init}
\begin{tabular}{lcccc}
\toprule
Method & $t_s=10$ & $t_s=15$ & $t_s=20$ & $t_s=25$ \\
\midrule
Base  & 64.4 & 35.6 & 24.4 & 15.6 \\
+ CAS & \textbf{72.2} & \textbf{51.1} & \textbf{34.4} & \textbf{23.3} \\
\bottomrule
\end{tabular}
\end{table}

\rev{
\paragraph{Evaluation with a different initial state for every switching pair.}
To verify that the conclusions do not depend on the shared state, we re-evaluate $\pi_0$ with a different initial state for every switching pair. From all 500 official LIBERO-Goal initial states (10 tasks $\times$ 50), excluding the shared state used in the main evaluation, we draw 90 states uniformly without replacement with a fixed seed and assign one to each ordered source--target pair, so every state is used exactly once. Within each pair, Base and CAS start from the same initial state with the same model seed. All other evaluation and CAS settings follow the main evaluation.

As shown in Table~\ref{tab:random_init}, CAS consistently improves switching success at every switching step under the randomized-state protocol. Averaged across switching steps, CAS improves the success rate from $35.0\%$ to $45.3\%$ ($+10.3$ percentage points), closely matching the $+10.6$-point improvement observed under the shared-state protocol in Table~\ref{tab:main_results}. The base policy also exhibits similar average performance under the two protocols ($35.0\%$ vs.\ $36.9\%$). These results indicate that the effectiveness of CAS is not specific to the shared initial state used in the main evaluation.
}

\subsection{Additional Ablation Analysis}
\label{app:abla}
\subsubsection{Steering-Weight Sweep for CAS Variants.}
\label{app:design}
\begin{figure}[t]
    \centering
    \includegraphics[width=0.7\linewidth]
    {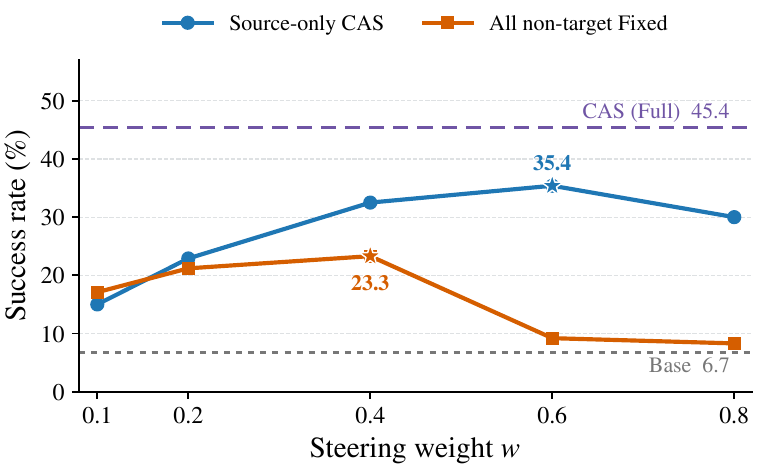}
     \caption{
    \textbf{Steering-weight sweep for CAS variants.}
    We vary the fixed steering weight
    $w \in \{0.1, 0.2, 0.4, 0.6, 0.8\}$
    for Source-only CAS and All non-target Fixed.
    Every tested setting of both variants outperforms the base policy.
    Source-only CAS achieves its best Switch SR of $35.4\%$
    at $w=0.6$, while All non-target Fixed peaks at $23.3\%$
    at $w=0.4$.
    Dashed horizontal lines show the base policy and full CAS for reference.
    }
    \label{fig:design_weight_ablation}
\end{figure}

We sweep the fixed steering weight for the two fixed-weight variants
in Table~\ref{tab:cas_design_ablation}.
Specifically, we evaluate
$w \in \{0.1, 0.2, 0.4, 0.6, 0.8\}$
for Source-only CAS and All non-target Fixed.
The complete results are shown in
Fig.~\ref{fig:design_weight_ablation}.

Remarkably, \emph{every tested steering weight for both variants
outperforms the base policy}.
This consistent improvement across the entire sweep demonstrates that
counterfactual steering provides a robust corrective signal rather than
relying on a narrowly tuned steering strength.
Source-only CAS achieves its best Switch SR of $35.4\%$ at
$w=0.6$, while All non-target Fixed peaks at $23.3\%$ at $w=0.4$;
these best-performing settings are reported in
Table~\ref{tab:cas_design_ablation}.

The sweep further reveals different sensitivity to the steering strength.
Source-only CAS remains consistently effective across a broad
range of $w$, further demonstrating the robustness of the lightweight
source-only strategy.
In contrast, All non-target Fixed degrades substantially as $w$
becomes large, consistent with over-steering when the same correction
strength is indiscriminately applied to multiple weakly relevant
counterfactual tasks.

\subsubsection{Hyperparameter Ablations}
\label{app:hyper}

We further study the effect of the three key hyperparameters in CAS: the number of nearest neighbors $K$, the KNN distance threshold $\delta$, and the steering strength $\lambda$. Specifically, $K$ controls the local neighborhood used to estimate the distance between the current end-effector position and each task trajectory distribution, while $\delta$ determines which non-target trajectories are selected as potential hijacking tasks. The steering strength $\lambda$ controls the overall magnitude of the counterfactual correction applied to the target-conditioned action. When $\lambda=0$, the counterfactual correction vanishes and CAS reduces exactly to the base policy without action steering.

Unless otherwise specified, we use $K=30$, $\delta=0.1$, and $\lambda=0.03$ as the default setting, which achieves a Switch SR of 45.4$\%$. Starting from this configuration, we vary one hyperparameter at a time while keeping the other two fixed.

\begin{figure*}[t]
    \centering
    \includegraphics[width=\textwidth]{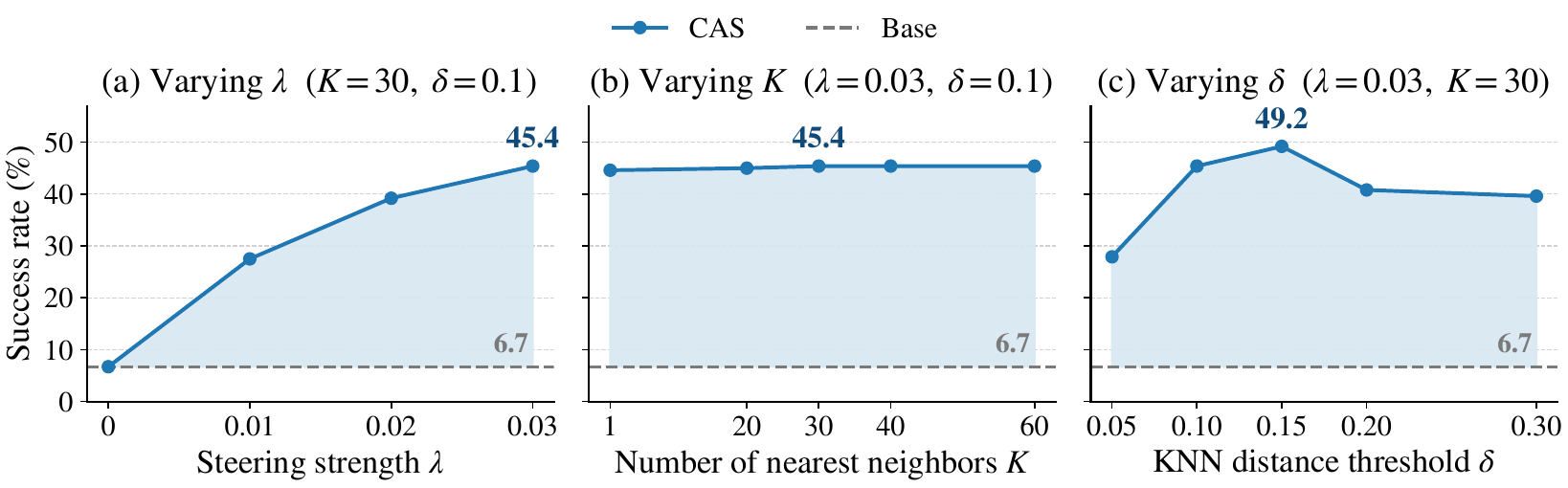}
    \caption{
        Hyperparameter ablations of CAS on CALVIN-Custom-Switch for $t_s \in [15,21]$.
        Starting from the default setting $K=30$, $\delta=0.1$, and $\lambda=0.03$, we vary one hyperparameter at a time while keeping the other two fixed:
        (a) steering strength $\lambda$, (b) number of nearest neighbors $K$, and (c) KNN distance threshold $\delta$.
        The dashed line denotes the base policy without CAS.
    }
    \label{fig:hyp3}
\end{figure*}

As shown in Fig.~\ref{fig:hyp3}, CAS maintains substantial improvements over the base policy across a broad range of hyperparameter settings. Increasing $\lambda$ from zero progressively strengthens counterfactual steering and leads to substantial gains in switching performance. In contrast, the performance remains highly stable across different choices of $K$, indicating that CAS does not require a precise neighborhood size for estimating local trajectory proximity. Varying $\delta$ changes how selectively CAS includes non-target trajectories as potential hijacking tasks, yet CAS remains substantially better than the base policy across all tested values. The best tested $\delta$ achieves a Switch SR of 49.2$\%$. Overall, these results suggest that CAS is effective across a reasonably broad range of choices for $K$, $\delta$, and $\lambda$, rather than relying on a narrowly tuned hyperparameter configuration.

\begin{figure}[t]
    \centering
    \includegraphics[width=\textwidth]{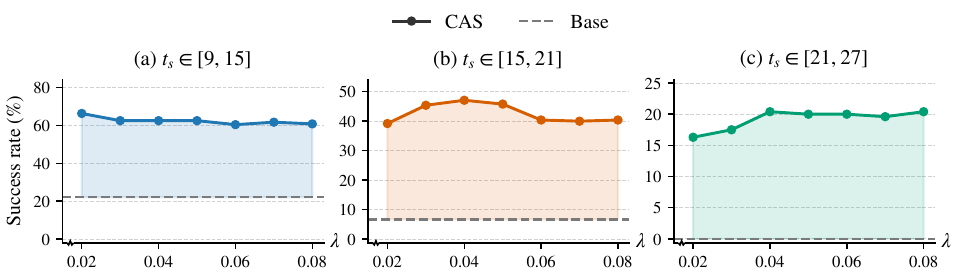}
    \caption{Effect of the steering strength $\lambda$ on CAS with $\pi_0$ on CALVIN-Custom-Switch ($K=30$, $\delta=0.1$) in the three switching ranges (a) $t_s\in[9,15]$, (b) $[15,21]$, and (c) $[21,27]$. Base denotes the base policy without CAS. Over a broad range of $\lambda$, CAS clearly improves over the base policy in all three switching ranges.}
    \label{fig:lambda_sweep}
\end{figure}

\rev{\paragraph{Steering strength across switching ranges.}
We further extend the $\lambda$ sweep from the $[15,21]$ range to all three switching ranges, keeping $K=30$ and $\delta=0.1$ fixed. As shown in Fig.~\ref{fig:lambda_sweep}, CAS clearly improves over the base policy across a broad range of $\lambda$ in all three switching ranges.}

\rev{
\subsubsection{Comparison of Alternative Action Steering Strategies}
\label{app:action_steering}

To examine whether the benefits of CAS can be explained by simple action amplification or arbitrary counterfactual corrections, we compare several alternative action steering strategies using $\pi_0$ on LIBERO-Goal-Switch and CALVIN-Custom-Switch (Table~\ref{tab:alternative_action_steering}). Let $a_0$ denote the target-conditioned action, $a_r$ the action conditioned on a randomly sampled non-target instruction, and $a_{\mathrm{CAS}}$ the full CAS output at the same observation. All transformations below apply to the six motion dimensions, while the gripper command remains that of the target-conditioned action.
}

\begin{table}[t]
\centering
\caption{
Comparison of full CAS with alternative action correction strategies on LIBERO-Goal-Switch and CALVIN-Custom-Switch.
We evaluate direct scaling of the target-conditioned action with different scaling factors $\alpha$, random counterfactual selection with adaptive KNN-based or fixed steering weights, reversed counterfactual correction, and final-action norm matching.
We also compare against random counterfactual action interpolation with $\beta=0.5$.
All methods use the $\pi_0$ policy architecture and the corresponding benchmark evaluation protocol, with the switching step set to $t_s=20$ on LIBERO and $t_s\in[15,21]$ on CALVIN.
We report target-task success rates (\%); higher is better.
}
\label{tab:alternative_action_steering}
\small
\setlength{\tabcolsep}{4pt}
\begin{adjustbox}{max width=\linewidth}
\begin{tabular}{lrr}
\toprule
Method & LIBERO, $t_s=20$ & CALVIN, $[15,21]$ \\
\midrule
Base policy & 30.0 & 6.7 \\
Full CAS & \textbf{37.8} & \textbf{45.4} \\
\midrule
Target-action scaling ($\alpha=1.5$) & 20.0 & 2.9 \\
Target-action scaling ($\alpha=2$) & 21.1 & 1.3 \\
Target-action scaling ($\alpha=3$) & 18.9 & 0.4 \\
Random counterfactuals (KNN-based weighting) & 25.6 & 22.9 \\
Random counterfactuals (fixed weighting, $w=0.6$) & 22.2 & 22.1 \\
CAS with reversed correction direction & 14.4 & 0.0 \\
CAS with final-action norm matching & 40.0 & 42.5 \\
Random counterfactual action interpolation ($\beta=0.5$) & 8.9 & 2.9 \\
\bottomrule
\end{tabular}
\end{adjustbox}
\end{table}

\rev{
\paragraph{Target-action scaling and random counterfactuals.}
Target-action scaling applies $a=\alpha a_0$, testing whether strengthening the target action alone is sufficient. Across $\alpha\in\{1.5,2,3\}$, success rates are 18.9--21.1\% on LIBERO and 0.4--2.9\% on CALVIN. For random counterfactuals, we uniformly sample one non-target instruction at each replanning step and apply $a=a_0+w(a_0-a_r)$. The weight is either computed from the sampled task's KNN trajectory distance or fixed at $w=0.6$. Neither random-counterfactual variant applies the distance-threshold gate $\delta$: we sample from all non-target tasks to evaluate random selection without distance-based filtering. KNN distances are used only to determine the steering weight in the adaptive variant. These variants achieve 25.6\%/22.9\% and 22.2\%/22.1\% on LIBERO/CALVIN, respectively. Neither simple scaling nor arbitrary counterfactual selection reproduces the performance observed with CAS, supporting the value of selecting state-relevant counterfactual actions.

The improvement of random counterfactuals with KNN-based weighting over the base policy on CALVIN (22.9\% vs.\ 6.7\%) may partly reflect its small task set: CALVIN-Custom-Switch contains only four tasks, leaving just three candidates after excluding the target task. Random selection therefore has a relatively high chance of choosing a task that provides a useful correction, which may explain the improvement even without targeted task selection. In contrast, on LIBERO, random counterfactuals with KNN-based weighting reduce success from 30.0\% to 25.6\%, suggesting that randomly selected counterfactual actions can introduce interference rather than provide useful steering.

\paragraph{Correction direction and final-action norm.}
To test the direction of the correction, we apply $a=a_0-(a_{\mathrm{CAS}}-a_0)$, reversing the CAS correction while preserving its magnitude at a given observation. Success drops to 14.4\% on LIBERO and 0.0\% on CALVIN. To separately examine final-action magnitude, we preserve the CAS output direction and rescale it as $a=(\|a_0\|_2/\|a_{\mathrm{CAS}}\|_2)a_{\mathrm{CAS}}$, with zero norms handled separately. This matches the joint six-dimensional motion-command norm before native controller processing. The resulting success rates are 40.0\% and 42.5\%. The norm-matched variant outperforms full CAS on LIBERO but underperforms it on CALVIN, suggesting that the effect of action magnitude depends on the benchmark. Its performance on both benchmarks indicates that increasing the final action norm alone is insufficient to explain the benefits of CAS.

\paragraph{Random counterfactual action interpolation.}
As a simple interpolation baseline, we apply $a=(1-\beta)a_0+\beta a_r$ with $\beta=0.5$, using the same uniform non-target instruction sampling rule. This yields only 8.9\% success on LIBERO and 2.9\% on CALVIN. Together, these comparisons support the importance of the content and direction of the counterfactual correction beyond merely amplifying the target action or mixing it with an arbitrary non-target action.
}

\section{Additional Experiments and Analyses of CAS-Guided Self-Training}
\label{app:ST}
\subsection{Qualitative Analysis of Higher-Quality CAS-guided Rollouts}
\label{app:better_ST}

\begin{figure*}[t]
    \centering

    \begin{subfigure}[t]{\textwidth}
        \centering
        \caption{Without CAS: Suboptimal interaction causes the plate to tilt}
        \includegraphics[width=\textwidth]{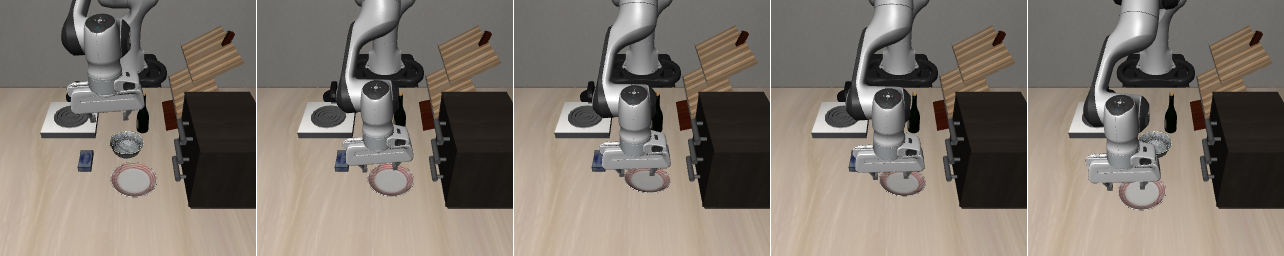}
    \end{subfigure}

    \vspace{2mm}

    \begin{subfigure}[t]{\textwidth}
        \centering
        \caption{With CAS: Improved redirection enables a smoother push}
        \includegraphics[width=\textwidth]{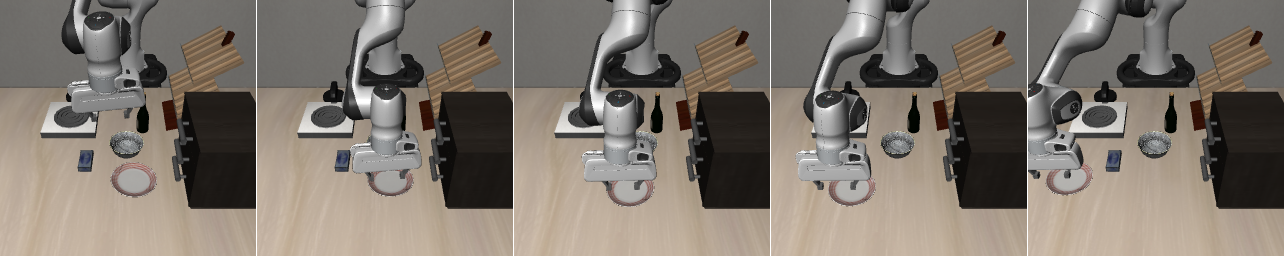}
    \end{subfigure}

    \caption{
    \textbf{Successful rollouts for a switch from \textit{put the bowl on the
    stove} to \textit{push the plate to the front of the stove}.}
    Both rollouts ultimately complete the target task. Without CAS, the policy
    remains more affected by the source-task trajectory prior and approaches
    the plate from a less suitable interaction position. As a result, one side
    of the plate lifts during the push, making the manipulation less smooth.
    With CAS, the gripper redirects toward a more suitable position for
    interacting with the plate and pushes it more smoothly toward the target
    location. Thus, although both rollouts succeed, CAS yields a cleaner
    post-switch manipulation trajectory.
    }
    \label{fig:better_ST_case1}
\end{figure*}

\begin{figure*}[t]
    \centering

    \begin{subfigure}[t]{\textwidth}
        \centering
        \caption{Without CAS: Insufficient redirection leads to a missed first grasp}
        \includegraphics[width=\textwidth]{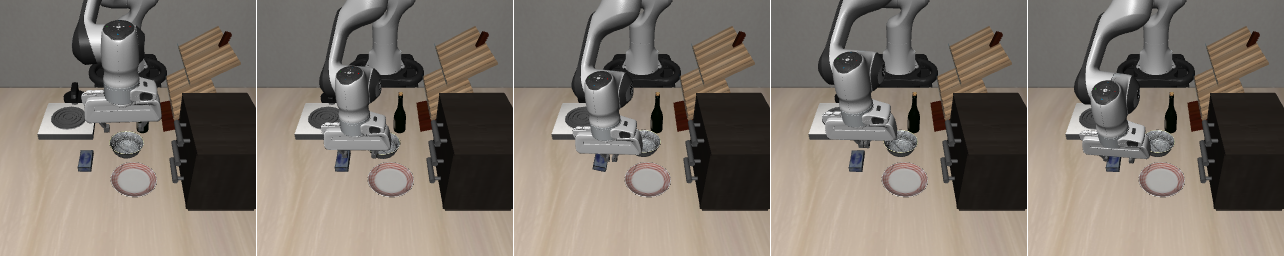}
    \end{subfigure}

    \vspace{2mm}

    \begin{subfigure}[t]{\textwidth}
        \centering
        \caption{With CAS: Improved redirection enables a successful first grasp}
        \includegraphics[width=\textwidth]{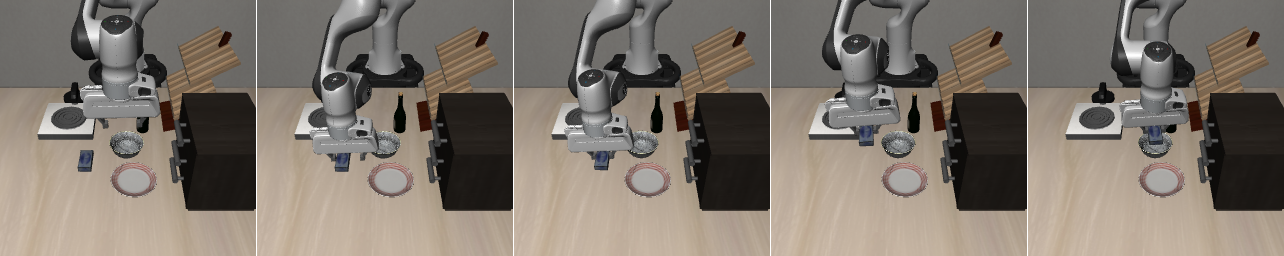}
    \end{subfigure}

    \caption{
    \textbf{Successful rollouts for a switch from \textit{push the plate to the
    front of the stove} to \textit{put the cream cheese in the bowl}.}
    Both rollouts eventually complete the target task. Without CAS, the initial
    post-switch redirection toward the cream cheese is insufficient, causing the
    first grasp attempt to miss. The policy must then re-approach the object and
    perform additional corrective motions before succeeding. With CAS, the
    gripper redirects more effectively toward the cream cheese and succeeds on
    the first grasp attempt, yielding a cleaner trajectory with fewer
    unnecessary intermediate actions.
    }
    \label{fig:better_ST_case2}
\end{figure*}

\begin{figure*}[t]
    \centering

    \begin{subfigure}[t]{\textwidth}
        \centering
        \caption{Without CAS: Insufficient redirection leads to an awkward front-edge grasp}
        \includegraphics[width=\textwidth]{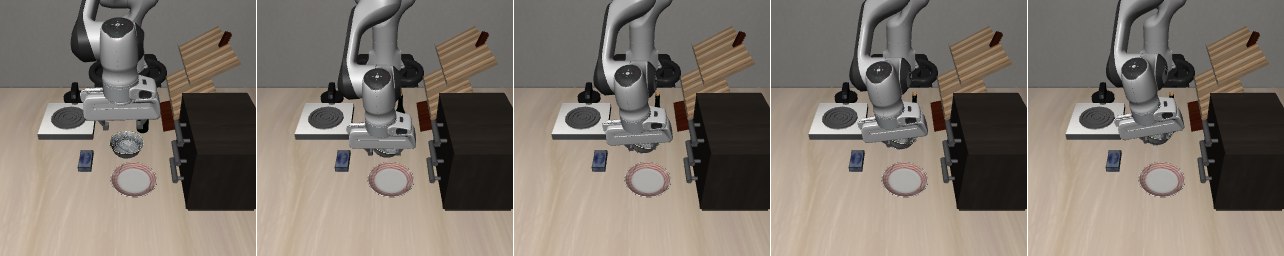}
    \end{subfigure}

    \vspace{2mm}

    \begin{subfigure}[t]{\textwidth}
        \centering
        \caption{With CAS: Improved redirection enables a stable side grasp}
        \includegraphics[width=\textwidth]{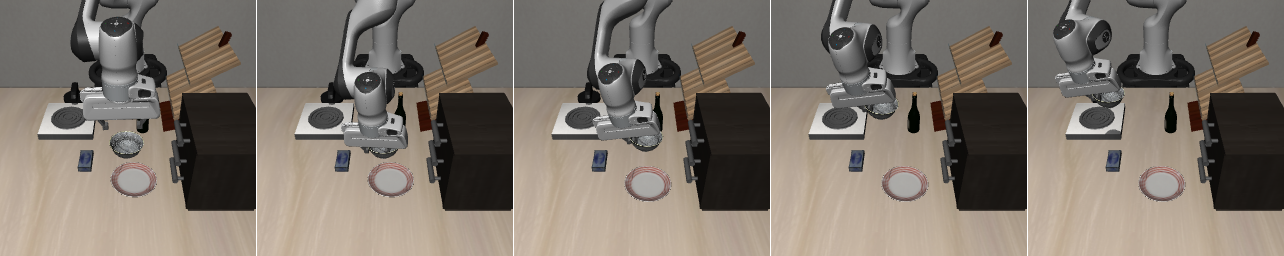}
    \end{subfigure}

    \caption{
    \textbf{Successful rollouts for a switch from \textit{push the plate to the
    front of the stove} to \textit{put the bowl on the stove}.}
    Both rollouts ultimately succeed. Without CAS, the policy does not redirect
    sufficiently toward a favorable grasp configuration and awkwardly catches
    the front edge of the bowl before lifting it. With CAS, the gripper is
    redirected toward a more suitable grasp pose and stably grasps the side of
    the bowl, resulting in more direct progress toward the updated task.
    Despite identical binary success outcomes, the two rollouts therefore
    exhibit clearly different post-switch manipulation quality.
    }
    \label{fig:better_ST_case3}
\end{figure*}
To better understand why successful CAS-guided rollouts provide more effective
supervision for self-training, we qualitatively compare successful rollouts
collected with and without CAS during the one-round self-training data
collection on LIBERO-Goal-Switch with $\pi_{0.5}$ at $t_s=25$. We deliberately
focus on cases where both rollouts ultimately complete the switched target task.
These examples therefore reveal differences in the quality of successful
trajectories that are not captured by the binary task-success metric.

Figures~\ref{fig:better_ST_case1}--\ref{fig:better_ST_case3} show three
representative examples. In each figure, the first displayed frame corresponds
to the instruction-switching state, followed by several key frames from the
post-switch rollout. The frames shown with and without CAS are temporally
aligned for direct comparison; the last displayed frame is not necessarily the
final frame of the episode.

\paragraph{Case analysis.}
Figure~\ref{fig:better_ST_case1} shows a switch from
\textit{put the bowl on the stove} to
\textit{push the plate to the front of the stove}. Although both rollouts
eventually complete the target task, their post-switch behaviors differ
substantially. Without CAS, the gripper does not sufficiently redirect toward a
suitable interaction configuration for the plate. It consequently contacts and
pushes the plate from an unfavorable position, causing one side of the plate to
tilt upward and making the subsequent manipulation less smooth. With CAS, the
counterfactual correction redirects the gripper toward a more suitable
interaction position, leading to a smoother and more direct push toward the
target location. Thus, even though both rollouts are successful, the CAS-guided
rollout provides a cleaner transition from the switching state toward the
updated task.

Figure~\ref{fig:better_ST_case2} shows a switch from
\textit{push the plate to the front of the stove} to
\textit{put the cream cheese in the bowl}. Without CAS, the gripper does not
redirect sufficiently toward the cream cheese. This results in a misaligned
first grasp attempt, after which the policy must re-approach the object and
perform additional corrective motions before eventually succeeding. With CAS,
the gripper redirects more directly toward the cream cheese and completes the
grasp on the first attempt. The resulting trajectory therefore avoids the
failed intermediate attempt and unnecessary corrective motions observed without
CAS.

Figure~\ref{fig:better_ST_case3} shows another switch from
\textit{push the plate to the front of the stove}, this time to
\textit{put the bowl on the stove}. Without CAS, the gripper reaches the bowl
from an awkward pose, catches its front edge, and lifts it using an undesirable
grasp. With CAS, the gripper reaches a more suitable target-task grasp pose,
allowing it to stably grasp the side of the bowl. The temporally aligned frames
also show more direct progress toward the updated task in the CAS-guided
rollout, consistent with its more appropriate post-switch redirection and grasp
configuration.

\paragraph{Implications for self-training.}
Across all three examples, the policy ultimately completes the switched target
task both with and without CAS. However, successful rollouts collected without CAS can still exhibit signs of trajectory hijacking, such as insufficient redirection, suboptimal interaction configurations, failed intermediate attempts, or unnecessary corrective motions.
CAS mitigates these issues and produces more effective de-hijacking behavior by providing sufficient post-switch redirection toward the updated target task, even in episodes that would eventually succeed without CAS.
These examples highlight an important distinction between \emph{task success}
and \emph{trajectory quality}. Binary task success treats the paired rollouts
above as equivalent, despite substantial differences in the post-switch
behaviors they provide as supervision. In particular, CAS-guided successful
rollouts provide cleaner examples of how the policy should redirect from a
state influenced by a non-target trajectory prior toward the updated target
task. This qualitative analysis therefore provides complementary evidence for
the higher training value of CAS-guided rollouts, consistent with the
self-training results in Table~\ref{tab:ST_main}, where CAS-ST consistently
outperforms Vanilla-ST with the same number of retained trajectories.

\subsection{Instruction-Conditioned Action Visualization}
\label{app:action_vis}
\begin{figure}[t]
    \centering
    \includegraphics[width=0.4\linewidth]{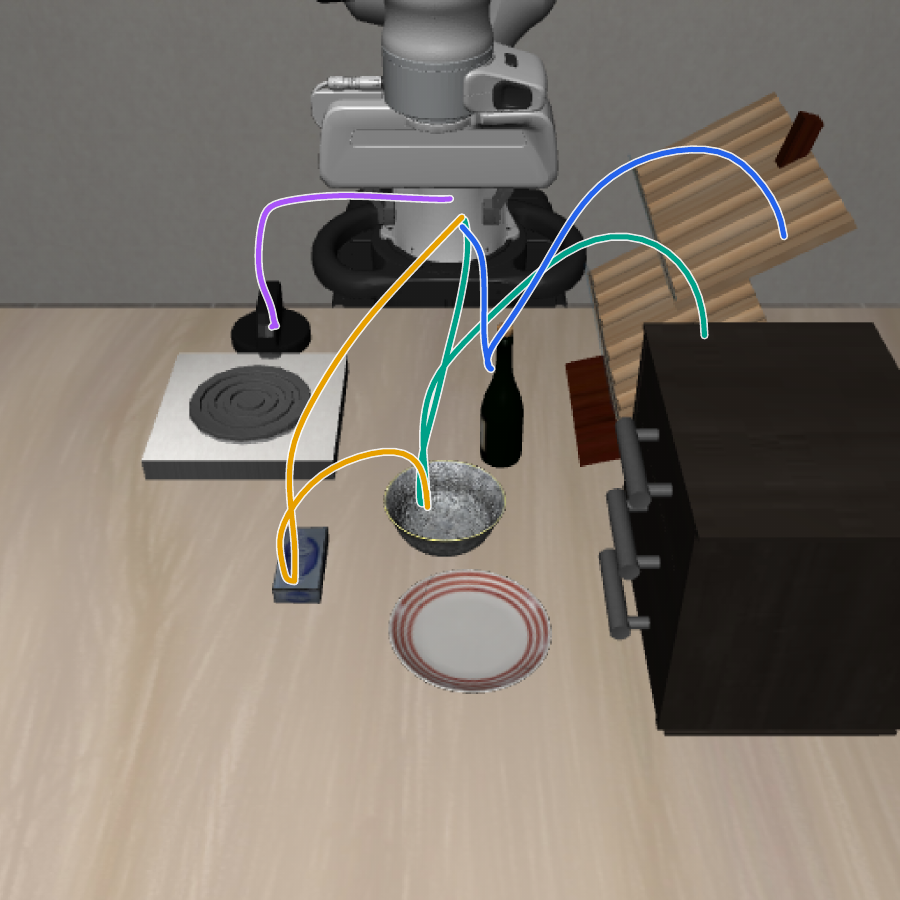}
    \caption{
    Official demonstration trajectories for the four LIBERO-Goal tasks used in our action visualization:
    \textcolor[HTML]{2563EB}{\textit{put the wine bottle on top of the cabinet}},
    \textcolor[HTML]{00A087}{\textit{put the bowl on top of the cabinet}},
    \textcolor[HTML]{E69F00}{\textit{put the cream cheese in the bowl}}, and
    \textcolor[HTML]{A855F7}{\textit{turn on the stove}}.
    The trajectory colors are consistent with those used in the action visualizations and provide visual references for the corresponding task-consistent motion directions.
    }
    \label{fig:nav}
\end{figure}

To further analyze the effect of self-training on instruction-conditioned action predictions, we compare instruction-conditioned actions before and after two rounds of CAS-ST, as in Fig.~\ref{fig:obs3}~(d). Specifically, we query the original and two-round CAS-ST checkpoints from the same robot states and visualize the actions predicted under different LIBERO-Goal task instructions. The robot states are sampled from intermediate states encountered during task-execution rollouts. To avoid clutter from displaying all ten tasks simultaneously, we select the same four tasks as in Fig.~\ref{fig:obs3}~(c,d): \textit{put the wine bottle on top of the cabinet}, \textit{put the bowl on top of the cabinet}, \textit{put the cream cheese in the bowl}, and \textit{turn on the stove}. To provide a visual reference for the task-consistent motion directions, we additionally show one official demonstration trajectory for each selected task in Fig.~\ref{fig:nav}.

These trajectories are intended to provide only a coarse reference for interpreting the action arrows below. In particular, the visualized arrows correspond to 3D end-effector translational actions projected onto the 2D camera plane, and thus motion components perpendicular to the image plane are not visible. In addition, task-consistent actions need not point directly toward the target object, since grasping may first require moving the gripper above the object before descending to grasp it. We therefore focus primarily on the coarse task-dependent motion directions. Despite these limitations, the contrasts between the two checkpoints are clear in the examples below: after two rounds of CAS-guided self-training, the predicted actions are generally better aligned with their corresponding task directions, while spurious overlap among actions for different task instructions is substantially reduced.

\begin{figure}[t]
    \centering
    \begin{minipage}[t]{0.4\linewidth}
        \centering
        \includegraphics[width=\linewidth]{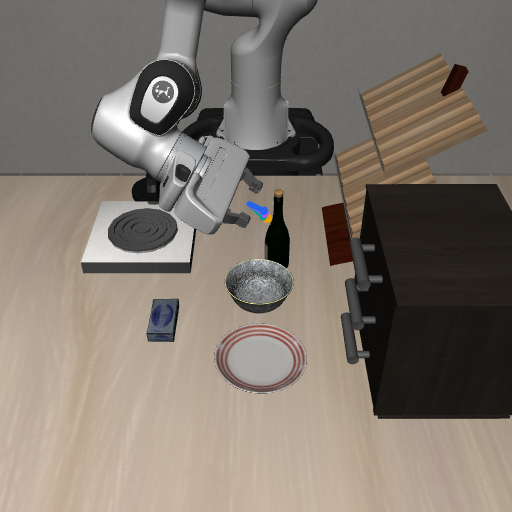}
        \vspace{-1mm}

        {\small (a) Base checkpoint}
    \end{minipage}
    \hspace{0.05\linewidth}
    \begin{minipage}[t]{0.4\linewidth}
        \centering
        \includegraphics[width=\linewidth]{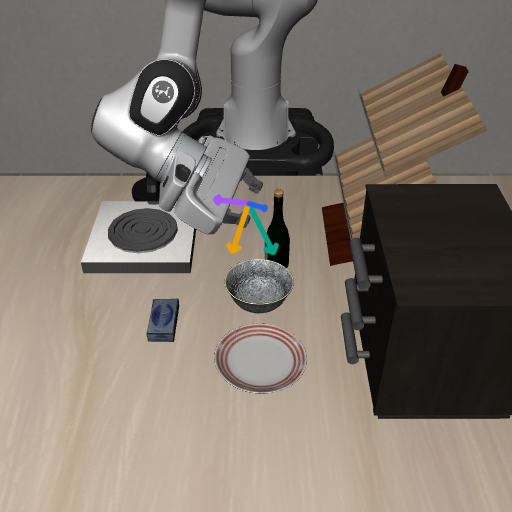}
        \vspace{-1mm}

        {\small (b) Two-round CAS-ST}
    \end{minipage}
    \caption{
    Instruction-conditioned actions for the same robot state before and after two rounds of CAS-guided self-training.
    Colors correspond to
    \textcolor[HTML]{2563EB}{\textit{wine bottle}},
    \textcolor[HTML]{00A087}{\textit{bowl}},
    \textcolor[HTML]{E69F00}{\textit{cream cheese}}, and
    \textcolor[HTML]{A855F7}{\textit{stove}} tasks.
    Under the base checkpoint, the four actions nearly overlap and point toward the wine-bottle region.
    After self-training, the actions become clearly separated and point broadly toward their corresponding task-relevant regions.
    }
    \label{fig:action_vis_case1}
\end{figure}

\begin{figure}[t]
    \centering
    \begin{minipage}[t]{0.4\linewidth}
        \centering
        \includegraphics[width=\linewidth]{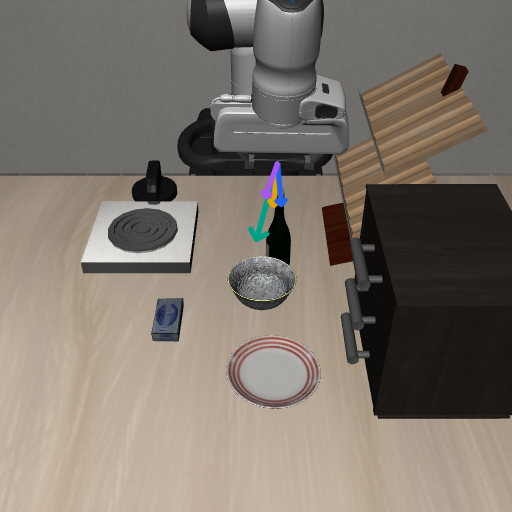}
        \vspace{-1mm}

        {\small (a) Base checkpoint}
    \end{minipage}
    \hspace{0.05\linewidth}
    \begin{minipage}[t]{0.4\linewidth}
        \centering
        \includegraphics[width=\linewidth]{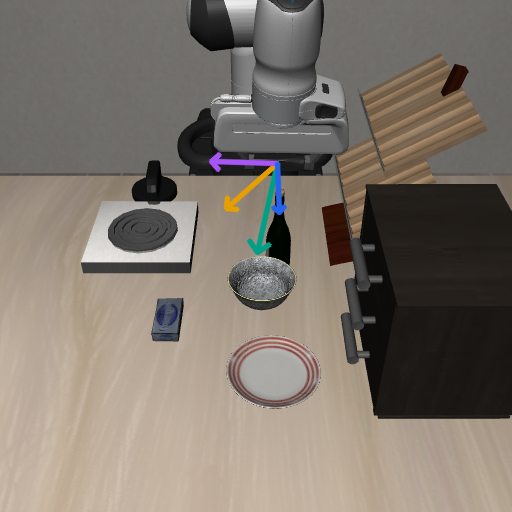}
        \vspace{-1mm}

        {\small (b) Two-round CAS-ST}
    \end{minipage}
    \caption{
    Another example of instruction-conditioned action predictions at the same robot state.
    Colors correspond to
    \textcolor[HTML]{2563EB}{\textit{wine bottle}},
    \textcolor[HTML]{00A087}{\textit{bowl}},
    \textcolor[HTML]{E69F00}{\textit{cream cheese}}, and
    \textcolor[HTML]{A855F7}{\textit{stove}} tasks.
    The base checkpoint predicts very similar actions across all four instructions, whereas the two-round CAS-ST checkpoint produces clearly separated, task-dependent action directions.
    }
    \label{fig:action_vis_case2}
\end{figure}

\begin{figure}[t]
    \centering
    \begin{minipage}[t]{0.4\linewidth}
        \centering
        \includegraphics[width=\linewidth]{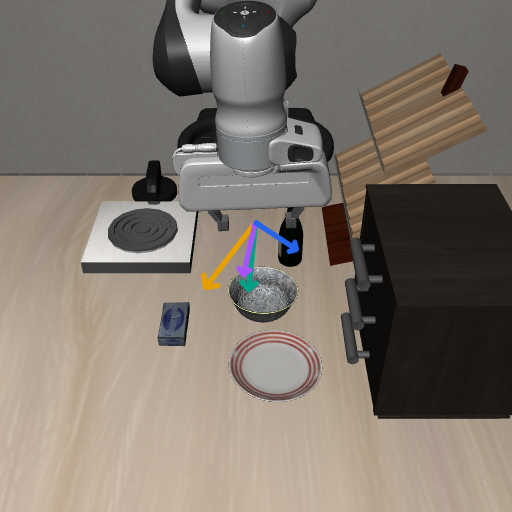}
        \vspace{-1mm}

        {\small (a) Base checkpoint}
    \end{minipage}
    \hspace{0.05\linewidth}
    \begin{minipage}[t]{0.4\linewidth}
        \centering
        \includegraphics[width=\linewidth]{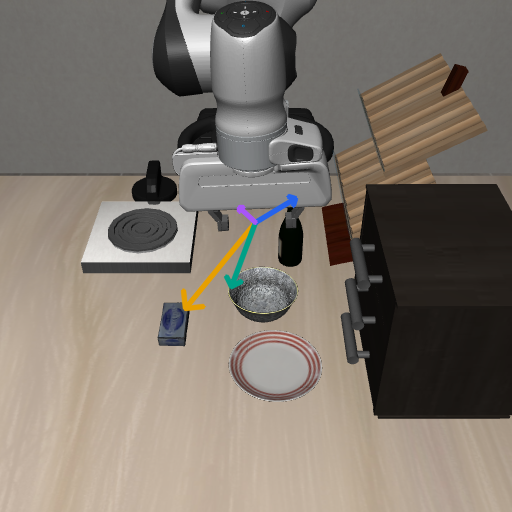}
        \vspace{-1mm}

        {\small (b) Two-round CAS-ST}
    \end{minipage}
    \caption{
    Instruction-conditioned actions in a state where the base checkpoint already exhibits partial task-dependent separation.
    Colors correspond to
    \textcolor[HTML]{2563EB}{\textit{wine bottle}},
    \textcolor[HTML]{00A087}{\textit{bowl}},
    \textcolor[HTML]{E69F00}{\textit{cream cheese}}, and
    \textcolor[HTML]{A855F7}{\textit{stove}} tasks.
    While some base actions are task-consistent, others remain misaligned.
    After two rounds of CAS-ST, all four actions become more clearly separated and better aligned with their corresponding task directions.
    }
    \label{fig:action_vis_case3}
\end{figure}

\begin{figure}[t]
    \centering
    \begin{minipage}[t]{0.4\linewidth}
        \centering
        \includegraphics[width=\linewidth]{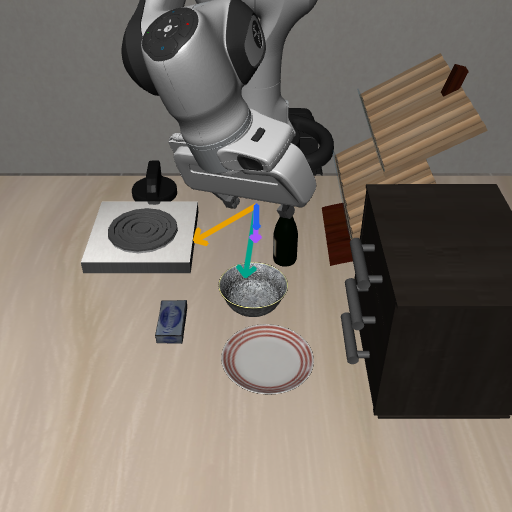}
        \vspace{-1mm}

        {\small (a) Base checkpoint}
    \end{minipage}
    \hspace{0.05\linewidth}
    \begin{minipage}[t]{0.4\linewidth}
        \centering
        \includegraphics[width=\linewidth]{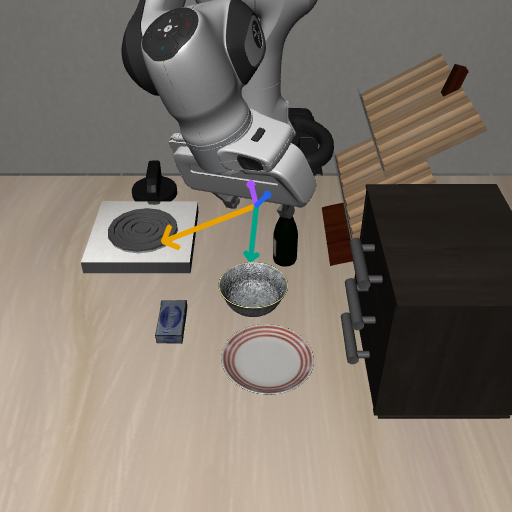}
        \vspace{-1mm}

        {\small (b) Two-round CAS-ST}
    \end{minipage}
    \caption{
    Instruction-conditioned actions in a state where the base checkpoint confuses a subset of task instructions.
    Colors correspond to
    \textcolor[HTML]{2563EB}{\textit{wine bottle}},
    \textcolor[HTML]{00A087}{\textit{bowl}},
    \textcolor[HTML]{E69F00}{\textit{cream cheese}}, and
    \textcolor[HTML]{A855F7}{\textit{stove}} tasks.
    In the base checkpoint, the wine-bottle and stove actions are biased toward the bowl region.
    After two rounds of CAS-ST, these actions shift toward directions more consistent with their respective tasks.
    }
    \label{fig:action_vis_case4}
\end{figure}

\paragraph{Case 1.}
As shown in Fig.~\ref{fig:action_vis_case1}, the base checkpoint predicts nearly overlapping actions for all four task instructions, with all arrows pointing broadly toward the wine-bottle region. After two rounds of CAS-ST, the actions become clearly separated: the purple action points toward the stove knob, the blue action toward the wine bottle, the orange action toward the cream-cheese region to the left of the bowl, and the green action toward the bowl. The resulting directions are substantially more task-dependent and better aligned with the corresponding tasks.

\paragraph{Case 2.}
In Fig.~\ref{fig:action_vis_case2}, the actions predicted for the four task instructions are very similar under the base checkpoint and all point approximately toward the wine bottle. After two rounds of CAS-ST, the actions become clearly separated: the purple, blue, orange, and green actions point broadly toward the stove knob, wine bottle, cream cheese, and bowl, respectively.

\paragraph{Case 3.}
In Fig.~\ref{fig:action_vis_case3}, the base checkpoint already exhibits some task-dependent action separation. The orange and green actions are broadly consistent with the cream-cheese and bowl tasks, respectively. However, the purple action, which should move toward the stove-knob region, instead points toward the bowl, while the blue action is poorly aligned with the wine-bottle direction. After two rounds of CAS-ST, all four actions become more clearly separated and better aligned with their corresponding task directions.

\paragraph{Case 4.}
As shown in Fig.~\ref{fig:action_vis_case4}, the orange and green actions predicted by the base checkpoint are broadly consistent with the cream-cheese and bowl tasks, respectively. In contrast, both the blue and purple actions are biased toward the bowl region rather than toward the wine bottle and stove. After two rounds of CAS-ST, the blue action redirects toward the wine bottle, while the purple action shifts toward the rear-left region, which is more consistent with the stove task.
\paragraph{Summary.}
Across these examples, CAS-guided self-training leads to clearer task-dependent action separation and generally better alignment with the corresponding task directions. Notably, even in challenging cases where the base checkpoint predicts highly similar actions across different instructions, these overlaps are substantially reduced after two rounds of CAS-ST.

\subsection{Self-Training Performance under Different Training Budgets}
\label{app:self_training_budget}

We further evaluate Vanilla-ST and CAS-ST under different finetuning budgets on CALVIN-Custom-Switch using Fast-WAM. The self-training data construction and finetuning setup follow Appendix~\ref{app:self_training_details}; here, we vary only the number of finetuning steps.

As shown in Table~\ref{tab:fastwam_st_budget}, CAS-ST consistently outperforms Vanilla-ST across all evaluated training budgets. Notably, CAS-ST reaches 42.9\% success with only 4k finetuning steps, already substantially outperforming Vanilla-ST trained for 20k steps (21.3\%). This further demonstrates the higher training value and finetuning efficiency of CAS-guided rollouts.

\begin{table}[t]
\centering
\caption{Self-training performance (\%) of Fast-WAM on CALVIN-Custom-Switch under different finetuning budgets. CAS-ST with only 4k finetuning steps already substantially outperforms Vanilla-ST trained for 20k steps.}
\label{tab:fastwam_st_budget}
\setlength{\tabcolsep}{12pt}
\begin{tabular}{@{}ccc@{}}
\toprule
Finetuning Steps & Vanilla-ST & CAS-ST \\
\midrule
4k  & 17.5 & 42.9 \\
8k  & 18.8 & 48.8 \\
12k & 20.8 & 51.3 \\
16k & 19.6 & 53.8 \\
20k & 21.3 & 50.0 \\
\bottomrule
\end{tabular}
\end{table}

\subsection{Pair-Level Analysis of CAS-Guided Self-Training}
\label{app:self_training_pairwise}

We further examine the CALVIN-Custom-Switch self-training results at the
individual source--target pair level.
We follow the self-training setup described in
Appendix~\ref{app:self_training_details}, using the $[9,15]$ switching range
for both data collection and evaluation.
Here, we compare the pair-level difference in the number of retained
self-training trajectories with the corresponding evaluation success rate.

\paragraph{CAS-ST can perform better with fewer training trajectories.}
Table~\ref{tab:calvin_pairwise_fewer} highlights two source--target pairs for
which CAS-ST uses fewer self-training trajectories than Vanilla-ST, while
achieving higher evaluation success.
For Red $\rightarrow$ Button, CAS-ST uses four fewer trajectories while
improving the success rate from 70.0\% to 100.0\%.
Similarly, for Button $\rightarrow$ Red, CAS-ST uses seven fewer trajectories
while improving the success rate from 65.0\% to 90.0\%.
Thus, for these pairs, the higher CAS-ST performance is observed despite
using fewer retained self-training trajectories.

\begin{table}[t]
    \centering
    \caption{
    Pair-level cases where CAS-ST uses fewer retained self-training
    trajectories than Vanilla-ST but achieves higher evaluation success.
    $\Delta$ Traj. denotes the difference in retained trajectories
    (CAS-ST $-$ Vanilla-ST), and $\Delta$ Succ. denotes the success-rate
    difference in percentage points.
    }
    \label{tab:calvin_pairwise_fewer}
    \small
    \renewcommand{\arraystretch}{1.08}
    \begin{tabular}{l r @{\hspace{1.5em}} r r r}
        \toprule
        & \multicolumn{1}{c}{$\Delta$ Traj.}
        & \multicolumn{3}{c}{Success Rate (\%)} \\
        \cmidrule(lr){2-2}
        \cmidrule(lr){3-5}
        Source $\rightarrow$ Target
        & CAS-ST $-$ Vanilla-ST
        & Vanilla-ST & CAS-ST & $\Delta$ Succ. (pp) \\
        \midrule
        Red $\rightarrow$ Button
        & $-4$
        & 70.0 & 100.0 & $+30.0$ \\
        Button $\rightarrow$ Red
        & $-7$
        & 65.0 & 90.0 & $+25.0$ \\
        \bottomrule
    \end{tabular}
\end{table}

\paragraph{Pairs with more CAS-ST trajectories show no regression.}
Table~\ref{tab:calvin_pairwise_more} reports all seven source--target pairs
for which CAS-ST contains more retained self-training trajectories than
Vanilla-ST.
CAS-ST achieves a higher success rate on six of these pairs and matches
Vanilla-ST on the remaining pair; none shows lower performance.
The additional number of retained trajectories ranges from three to eleven
across individual pairs.
For example, Red $\rightarrow$ Blue has six additional CAS-ST trajectories
and improves from 0.0\% to 45.0\%, while Pink $\rightarrow$ Button also has
six additional trajectories and improves from 10.0\% to 95.0\%.
Across all seven pairs, CAS-ST contains 49 more retained trajectories, with a
pooled success rate of 42.9\% compared with 3.6\% for Vanilla-ST.

\begin{table}[t]
    \centering
    \caption{
    All source--target pairs where CAS-ST has more retained self-training
    trajectories than Vanilla-ST.
    CAS-ST improves six pairs and ties one, with no observed regression.
    $\Delta$ Traj. denotes the difference in retained trajectories
    (CAS-ST $-$ Vanilla-ST), and $\Delta$ Succ. denotes the success-rate
    difference in percentage points.
    }
    \label{tab:calvin_pairwise_more}
    \small
    \renewcommand{\arraystretch}{1.08}
    \begin{tabular}{l r @{\hspace{1.5em}} r r r}
        \toprule
        & \multicolumn{1}{c}{$\Delta$ Traj.}
        & \multicolumn{3}{c}{Success Rate (\%)} \\
        \cmidrule(lr){2-2}
        \cmidrule(lr){3-5}
        Source $\rightarrow$ Target
        & CAS-ST $-$ Vanilla-ST
        & Vanilla-ST & CAS-ST & $\Delta$ Succ. (pp) \\
        \midrule
        Red $\rightarrow$ Pink
        & $+8$
        & 0.0 & 20.0 & $+20.0$ \\
        Red $\rightarrow$ Blue
        & $+6$
        & 0.0 & 45.0 & $+45.0$ \\
        Pink $\rightarrow$ Red
        & $+3$
        & 15.0 & 60.0 & $+45.0$ \\
        Pink $\rightarrow$ Blue
        & $+8$
        & 0.0 & 75.0 & $+75.0$ \\
        Pink $\rightarrow$ Button
        & $+6$
        & 10.0 & 95.0 & $+85.0$ \\
        Blue $\rightarrow$ Pink
        & $+7$
        & 0.0 & 0.0 & $0.0$ \\
        Button $\rightarrow$ Pink
        & $+11$
        & 0.0 & 5.0 & $+5.0$ \\
        \midrule
        Overall
        & $+49$
        & 3.6 & 42.9 & $+39.3$ \\
        \bottomrule
    \end{tabular}
\end{table}

Overall, this pair-level analysis provides a complementary view of the
aggregate self-training results.
For some task pairs, CAS-ST achieves higher success despite using fewer
training trajectories; meanwhile, among all pairs where CAS-ST uses more
trajectories, we observe either improved or unchanged pair-level performance.

\subsection{Further Generalization to Later Switches}
\label{app:self_training_later_switches}

We further evaluate the extrapolation of CAS-guided self-training to later switching settings on LIBERO-Goal-Switch. Following the self-training setup in Appendix~\ref{app:self_training_details}, all self-training trajectories are collected at $t_s=25$. In addition to the unseen later switch at $t_s=30$ reported in Table~\ref{tab:ST_main}, we further evaluate the same finetuned policies at $t_s=35$, extending the evaluation further beyond the switching setting used for data collection.

As shown in Table~\ref{tab:self_training_later_switches}, CAS-ST consistently outperforms Vanilla-ST at both unseen later switching steps for both $\pi_{0.5}$ and Fast-WAM. At $t_s=35$, CAS-ST achieves 41.1\% and 22.2\% success for $\pi_{0.5}$ and Fast-WAM, respectively, exceeding Vanilla-ST by 6.7 and 6.6 percentage points. These additional results show that CAS-ST continues to outperform Vanilla-ST at a later unseen switching step.

\begin{table}[t]
\centering
\caption{
Self-training performance (\%) on LIBERO-Goal-Switch at unseen later switching steps.
All self-training trajectories are collected at $t_s=25$.
The $t_s=30$ results are also reported in Table~\ref{tab:ST_main}, while $t_s=35$ provides an additional evaluation farther beyond the self-training regime.
}
\label{tab:self_training_later_switches}
\setlength{\tabcolsep}{8pt}
\begin{tabular}{@{}lcccc@{}}
\toprule
& \multicolumn{2}{c}{$\pi_{0.5}$} & \multicolumn{2}{c}{Fast-WAM} \\
\cmidrule(lr){2-3} \cmidrule(lr){4-5}
Method & $t_s=30$ & $t_s=35$ & $t_s=30$ & $t_s=35$ \\
\midrule
Base       & 27.8 & 25.5 & 16.7 & 14.4 \\
Vanilla-ST & 32.2 & 34.4 & 21.1 & 15.6 \\
CAS-ST     & 40.0 & 41.1 & 27.8 & 22.2 \\
\bottomrule
\end{tabular}
\end{table}

\subsection{Self-Training on CALVIN-Custom-Switch-Random}
\label{app:calvin_random_self_training}

We further examine CAS-guided self-training on
\emph{CALVIN-Custom-Switch-Random}, introduced in
Appendix~\ref{app:calvin_random_details}, where object placements are broadly
randomized without fixed nominal locations. As discussed above, such
randomization substantially changes the spatial organization of task trajectories and may consequently weaken the task-associated trajectory priors induced at particular states. We therefore
investigate whether the advantage of CAS-guided self-training persists under
this randomized object-placement setting.

As shown in Table~\ref{tab:calvin_random_self_training}, inference-time CAS substantially improves the success rate from $8.8\%$ to $34.2\%$. More importantly, CAS-ST reaches $27.9\%$ success without using CAS at inference time, while Vanilla-ST provides no improvement over the base policy ($8.3\%$ vs.\ $8.8\%$). These results show that CAS-guided rollouts remain effective self-training data even under fully randomized trajectory distributions, and provide substantially more useful supervision than vanilla successful rollouts in this setting.

\begin{table}[t]
\centering
\caption{
Self-training results on CALVIN-Custom-Switch-Random in the $[15,21]$ switching-step range.
CAS-ST substantially improves $\pi_0$, while Vanilla-ST does not provide a meaningful gain.
}
\label{tab:calvin_random_self_training}
\setlength{\tabcolsep}{6pt}
\begin{adjustbox}{max width=\linewidth,center}
\begin{tabular}{@{}lc@{}}
\toprule
Method & Success Rate (\%) \\
\midrule
Base       & 8.8 \\
+ CAS      & \textbf{34.2} \\
Vanilla-ST & 8.3 \\
CAS-ST     & 27.9 \\
\bottomrule
\end{tabular}
\end{adjustbox}
\end{table}

\FloatBarrier
\rev{
\subsection{Self-Training under Matched Tasks and Switching States}
\label{app:matched_self_training}

Matching the number of successful trajectories alone does not control task
composition or starting-state coverage. We therefore conduct a paired
self-training experiment with $\pi_{0.5}$ on LIBERO-Goal-Switch, matching the
switching states, source--target task pairs, and trajectory counts per pair.

For each of the ten tasks, we execute the base policy for 25 actions from ten
distinct initial states, then switch each resulting prefix to the other nine
target tasks, yielding 900 switching cases. Vanilla and CAS branch from the
same saved simulator state, with identical initial observations and policy
seeds. We retain 223 jointly successful pairs, with initial target-action
predictions aligned before CAS is applied. We use their complete post-switch trajectories, labeled with the
target instructions. Both training sets thus contain exactly the same
switching cases and task-pair counts.

Each set is combined with the same 500 official LIBERO-Goal demonstrations
(Table~\ref{tab:matched_st_data}). Both policies are fully fine-tuned from the
same official \texttt{pi05\_libero} checkpoint for 10,000 steps, with global
batch size 32 and seed 42.

\begin{table}[htbp]
\centering
\caption{Training data for the paired self-training comparison.
All selected trajectories and frames are retained without truncation or downsampling.}
\label{tab:matched_st_data}
\small
\setlength{\tabcolsep}{5pt}
\begin{tabular}{lrrrr}
\toprule
Training group & \makecell{Post-switch\\trajectories} & \makecell{Official\\demonstrations} & \makecell{Post-switch\\frames} & \makecell{Total dataset\\frames} \\
\midrule
Vanilla-ST & 223 & 500 & 27,495 & 91,223 \\
CAS-ST & 223 & 500 & 23,887 & 87,615 \\
\bottomrule
\end{tabular}
\end{table}

Both policies are evaluated without test-time CAS on the same 90 off-diagonal
task pairs at each switching time. As shown in Table~\ref{tab:matched_st_results},
CAS-ST outperforms Vanilla-ST at all four settings, including the later
switches not used for data collection. Overall success increases from
34.2\% to 41.9\%, a gain of 7.8 percentage points.

\begin{table}[htbp]
\centering
\caption{Success rates after 10,000 fine-tuning steps, without
test-time CAS. Each switching time has 90 episodes per method; Overall pools
all 360 episodes. Gains are computed from unrounded rates.}
\label{tab:matched_st_results}
\small
\begin{tabular}{lrrr}
\toprule
Switching time & Vanilla-ST (\%) & CAS-ST (\%) & Gain (percentage points) \\
\midrule
$t_s=25$ & 44.4 & \textbf{51.1} & +6.7 \\
$t_s=30$ & 31.1 & \textbf{44.4} & +13.3 \\
$t_s=35$ & 32.2 & \textbf{38.9} & +6.7 \\
$t_s=40$ & 28.9 & \textbf{33.3} & +4.4 \\
Overall & 34.2 & \textbf{41.9} & +7.8 \\
\bottomrule
\end{tabular}
\end{table}

CAS-ST achieves these gains with 13.1\% fewer post-switch frames and 4.0\%
fewer total dataset frames. The shorter successful trajectories are consistent
with the qualitative comparisons in Appendix~\ref{app:better_ST}: CAS redirects
the robot more decisively toward the updated target and more suitable
interaction poses, avoiding missed grasp attempts and unnecessary corrective
motions. These behaviors provide cleaner supervision for post-switch recovery,
even when both methods ultimately succeed. The improvement therefore persists
under matched task composition and switching-state coverage.
}
\FloatBarrier

\section{Beyond Instruction Switching: Trajectory-Biased Initial States}
\label{app:trajectory_biased_initial_states}

CAS is introduced in this work to address trajectory hijacking under mid-execution instruction switching.
However, interfering trajectory priors may also arise when the robot is initialized from a configuration that is already strongly biased toward particular task trajectories.
This raises a natural question: can CAS remain effective when the robot starts from a trajectory-biased initial state that is not induced by instruction switching?
To study this complementary setting, we construct trajectory-biased initial states and evaluate whether CAS can improve target-task execution from these challenging initial configurations.
As an auxiliary analysis beyond our primary instruction-switching setting, we focus on the VLA policies OpenVLA-OFT, $\pi_0$, and $\pi_{0.5}$.

\subsection{Construction of Trajectory-Biased Initial States}
\label{app:biased_init_construction}

We construct trajectory-biased initial states on LIBERO-Goal. For each task $i$, we aggregate the end-effector positions from its official demonstration trajectories into a task-specific trajectory point cloud $D_i$, following the notation in Sec.~\ref{sec:inference_time_cas}.
Given a candidate robot configuration with end-effector position $\mathbf{p}\in\mathbb{R}^3$, we measure its proximity to task $i$ using the average distance to its $K$ nearest demonstration points:
\[
d_i(\mathbf{p})
=
\frac{1}{K}
\sum_{\mathbf{q}\in \mathrm{KNN}(\mathbf{p},D_i)}
\|\mathbf{p}-\mathbf{q}\|_2.
\]
A smaller $d_i(\mathbf{p})$ indicates that the current end-effector position lies closer to the demonstrated trajectory distribution of task $i$.
We sort the distances to the ten task trajectories as
\[
d_{(1)}(\mathbf{p})
\leq
d_{(2)}(\mathbf{p})
\leq
\cdots
\leq
d_{(10)}(\mathbf{p}).
\]

To generate candidate states, we perturb the robot arm around its default configuration and allow the simulator to settle.
We retain a candidate state only if
\[
d_{(10)}(\mathbf{p}) < \tau,
\qquad
d_{(4)}(\mathbf{p}) - d_{(1)}(\mathbf{p}) > \delta_{\mathrm{gap}},
\]
where we set $\tau=0.12$ and $\delta_{\mathrm{gap}}=0.03$.
The first condition keeps the end-effector within a region well covered by the demonstrated task trajectories, ensuring that its proximity to different trajectory distributions remains meaningful.
The second condition enforces a pronounced imbalance among these trajectory proximities, such that the current end-effector position is substantially closer to a small subset of task trajectories than to the others.
We use $d_{(4)}$ rather than $d_{(2)}$ because several task trajectories can partially overlap in the workspace.
Together, these criteria construct initial states in which a particular subset of trajectory priors is already dominant before any task execution begins, creating challenging cases for following a potentially conflicting target instruction.

Following this procedure, we construct 42 trajectory-biased initial states. For each initial state, we evaluate all ten LIBERO-Goal target instructions, yielding 420 evaluation episodes for each policy--method configuration. The same fixed set of 42 initial states and target-instruction assignments is shared across all evaluated methods.

\subsection{Inference-Time Evaluation}
\label{app:biased_init_inference}

\paragraph{Main results.}

\begin{table}[t]
\centering
\caption{
Success rates (\%) from trajectory-biased initial states on LIBERO-Goal.
CAS consistently improves all evaluated VLA policies and outperforms PCD where applicable.
}
\label{tab:biased_init_inference}
\setlength{\tabcolsep}{6pt}
\begin{tabular}{llc}
\toprule
Policy & Method & Success Rate (\%) \\
\midrule
\multirow{2}{*}{OFT}
& Base  & 25.5 \\
& + CAS & \textbf{39.5} \\
\midrule
\multirow{3}{*}{$\pi_0$}
& Base  & 30.2 \\
& + PCD & 31.2 \\
& + CAS & \textbf{40.5} \\
\midrule
\multirow{3}{*}{$\pi_{0.5}$}
& Base  & 74.0 \\
& + PCD & 76.2 \\
& + CAS & \textbf{78.6} \\
\bottomrule
\end{tabular}
\end{table}

We first evaluate whether CAS can directly improve target-task execution from trajectory-biased initial states at inference time.
All methods are evaluated on the same 420 state--target episodes constructed from the 42 initial states described above. As shown in Table~\ref{tab:biased_init_inference}, CAS consistently improves all three evaluated VLA policies.
For OpenVLA-OFT, CAS increases the success rate from 25.5\% to 39.5\%; for $\pi_0$, from 30.2\% to 40.5\%; and for $\pi_{0.5}$, from 74.0\% to 78.6\%.
For $\pi_0$ and $\pi_{0.5}$, we additionally compare against the training-free PCD baseline, which provides only modest improvements over the corresponding base policies, while CAS achieves consistently stronger performance.
These results show that CAS remains effective when trajectory bias is already present in the initial robot configuration, even without any preceding source-task execution or instruction switch.

\paragraph{Robustness across checkpoints and observation settings.}
\label{app:additional_oft_checkpoint_view_eval}

We further examine whether the benefit of CAS under trajectory-biased initial states is specific to a particular OpenVLA-OFT checkpoint or observation configuration.
In addition to the official LIBERO-Goal OFT checkpoint used above, we finetune an OFT model ourselves on the same task suite and evaluate CAS on this reproduced checkpoint.
We also consider a more constrained single-view setting in which the wrist camera is removed and the policy receives only the third-person observation.

\begin{table}[t]
\centering
\caption{
Success rates (\%) from trajectory-biased initial states under different OpenVLA-OFT checkpoints and observation configurations.
}
\label{tab:libero_goal_additional_oft}
\setlength{\tabcolsep}{5pt}
\begin{tabular}{llcc}
\toprule
Checkpoint & Observation & Base & + CAS \\
\midrule
Official LIBERO-Goal OFT
& Third-person + wrist
& 25.5 & 39.5 \\
Our reproduced OFT
& Third-person + wrist
& 40.0 & 51.9 \\
Our reproduced OFT
& Third-person only
& 32.9 & 41.2 \\
\bottomrule
\end{tabular}
\end{table}

As shown in Table~\ref{tab:libero_goal_additional_oft}, CAS consistently improves target-task success across all evaluated settings.
On the official OFT checkpoint, CAS improves the success rate from 25.5\% to 39.5\%.
On our reproduced OFT checkpoint, CAS improves performance from 40.0\% to 51.9\%.
Even in the single-view setting without the wrist camera, CAS increases the success rate from 32.9\% to 41.2\%.
These results suggest that the benefit of CAS under trajectory-biased initial states is not specific to a single checkpoint or camera configuration.

\subsection{CAS-Guided Self-Training}
\label{app:biased_init_self_training}

\paragraph{Main results.}

\begin{table}[t]
\centering
\caption{
Self-training performance (\%) on the 42 trajectory-biased initial states on LIBERO-Goal.
CAS-ST consistently outperforms Vanilla-ST across all three evaluated VLA policies.
}
\label{tab:biased_init_self_training}
\setlength{\tabcolsep}{7pt}
\begin{tabular}{lccc}
\toprule
Method & OFT & $\pi_0$ & $\pi_{0.5}$ \\
\midrule
Base
& 25.5 & 30.2 & 74.0 \\
+ Vanilla-ST
& 33.1 & 49.8 & 78.8 \\
+ CAS-ST
& \textbf{48.8} & \textbf{53.1} & \textbf{80.7} \\
\bottomrule
\end{tabular}
\end{table}

We next examine whether CAS-guided trajectories can provide more effective supervision for improving the underlying policy under trajectory-biased initial states.
We compare CAS-guided self-training (CAS-ST) with vanilla self-training (Vanilla-ST).
Both methods finetune the same base policy using successful trajectories collected from the same distribution of trajectory-biased initial states, with collection states kept separate from the 42 states used for evaluation. Vanilla-ST and CAS-ST use exactly the same collection episodes, with matched initial robot states, object configurations, and target instructions. The only difference during data collection is whether CAS is enabled. Since CAS yields more successful rollouts, we uniformly downsample the CAS success pool to match the number of retained Vanilla-ST trajectories.
Both resulting policies are evaluated without CAS at inference time, so the comparison reflects what is learned from the collected trajectories rather than any inference-time steering effect.

As shown in Table~\ref{tab:biased_init_self_training}, CAS-ST consistently outperforms Vanilla-ST across all three evaluated VLA policies.
For OpenVLA-OFT, CAS-ST improves the success rate from 25.5\% to 48.8\%, compared with 33.1\% for Vanilla-ST.
For $\pi_0$, CAS-ST reaches 53.1\%, compared with 49.8\% for Vanilla-ST, while for $\pi_{0.5}$ it achieves 80.7\%, compared with 78.8\%.
Consistent with our findings under mid-execution instruction switching, these results further show that successful CAS-guided trajectories provide more effective self-training data than successful trajectories collected by the base policy alone, even when the trajectory bias is already present in the initial robot state.

\paragraph{Finetuning efficiency.}
\label{app:biased_init_self_training_steps}

\begin{table}[t]
\centering
\caption{
Self-training performance (\%) on the 42 trajectory-biased initial states under different finetuning budgets.
CAS-ST is substantially more finetuning-efficient: with only $10$k finetuning steps, it already outperforms Vanilla-ST trained for $50$k steps.
}
\label{tab:biased_init_self_training_steps}
\setlength{\tabcolsep}{6pt}
\begin{adjustbox}{max width=\linewidth,center}
\begin{tabular}{@{}lcc@{}}
\toprule
Method & Finetuning Steps & Success Rate (\%) \\
\midrule
Base & -- & 25.5 \\
\midrule
+ Vanilla-ST & $10$k & 27.9 \\
+ Vanilla-ST & $30$k & 34.1 \\
+ Vanilla-ST & $50$k & 33.1 \\
\midrule
+ CAS-ST & $10$k & 41.7 \\
+ CAS-ST & $30$k & 47.9 \\
+ CAS-ST & $50$k & \textbf{48.8} \\
\bottomrule
\end{tabular}
\end{adjustbox}
\end{table}

We further examine whether the advantage of CAS-guided self-training persists across different finetuning budgets.
All experiments use OpenVLA-OFT as the base policy and follow the same self-training protocol described above; Vanilla-ST and CAS-ST differ only in whether the successful training trajectories are collected with or without CAS guidance.

As shown in Table~\ref{tab:biased_init_self_training_steps}, CAS-ST consistently outperforms Vanilla-ST across all evaluated finetuning budgets.
With only $10$k finetuning steps, CAS-ST already reaches 41.7\% success, substantially exceeding the 33.1\% achieved by Vanilla-ST after $50$k steps.
Increasing the finetuning budget further improves CAS-ST to 47.9\% at $30$k steps and 48.8\% at $50$k steps, whereas Vanilla-ST provides substantially smaller gains and does not improve further from $30$k to $50$k steps.
Consistent with our findings under mid-execution instruction switching, these results show that CAS-guided trajectories provide more effective supervision and enable substantially more efficient policy improvement.
The same trend therefore holds when the challenging trajectory bias is already present in the robot's initial configuration rather than being induced by a preceding source-task execution.

\paragraph{Iterative self-training and standard-task performance.}

Since CAS-guided self-training can be applied iteratively, we further perform a second round of CAS-ST with OpenVLA-OFT.
For this second round, we generate a new set of trajectory-biased initial states using the same construction procedure described above, while ensuring that these states are disjoint from the 42 states used for evaluation.
We collect successful CAS-guided trajectories from these newly generated states and mix them with the original official LIBERO-Goal demonstration trajectories for further finetuning.

The second round provides an additional improvement on the trajectory-biased initial-state evaluation, increasing the success rate from 48.8\% after one round of CAS-ST to 51.7\% after two rounds.
This result shows that additional CAS-guided recovery data can continue to improve the policy beyond a single round of self-training.

Importantly, the resulting policy also maintains strong performance under the standard task setting.
On the standard LIBERO-Goal benchmark, it achieves \textbf{98.6\%} success, exceeding the \textbf{97.9\%} success rate reported for the official OpenVLA-OFT checkpoint.

\section{Experimental Details}
\label{app:experimental_details}

\subsection{Mid-Execution Instruction Switching Evaluation Details}
\label{app:obs1_details}

\paragraph{Switch-boundary handling.}
We define the switching step $t_s$ at the environment-action level.
At the switch, we discard any unexecuted source-conditioned actions in the
current action chunk, and the policy replans from the current observation under
the target instruction. For LingBot-VA, we additionally clear its cached
temporal state to eliminate explicit carryover from the source phase.

\paragraph{LIBERO-Goal online evaluation.}
We evaluate fully online instruction switching with
$t_s\in\{0,5,10,15,20,25\}$. At $t_s=0$, the policy executes the target
instruction directly from the common initial state. For $t_s>0$, it first
executes the source instruction for exactly $t_s$ environment actions and then
switches to the target instruction.

For each $t_s$, we evaluate all 90 ordered pairs of distinct source and target
tasks and report the mean target-task success rate. The initial state,
simulator stabilization, 300-action target-phase budget, and success criterion
follow Appendix~\ref{app:libero_goal_switch_details}.

\paragraph{CALVIN-Custom online evaluation.}
The CALVIN-Custom results in Fig.~\ref{fig:obs1} use an online switching
protocol, unlike the state-matched CALVIN-Custom-Switch protocol used for the
main method evaluation. We use the same four tasks as in
CALVIN-Custom-Switch.

For each nonzero switching range, we use 20 held-out initial configurations
and all 12 ordered source--target task pairs, yielding 240 episodes per range.
We consider four progressively later switching ranges:
$t_s\in[3,9]$, $[9,15]$, $[15,21]$, and $[21,27]$.
These ranges correspond to progress intervals
$[0.1,0.3]$, $[0.3,0.5]$, $[0.5,0.7]$, and $[0.7,0.9]$
on a 30-step reference scale. The reference scale determines only the
switching time; each policy executes its own source-conditioned prefix online.

After the switch, the policy has up to 180 environment actions to complete the
target task. For $t_s=0$, we directly execute each target task from 50 held-out
initial configurations, yielding 200 episodes in total.

\paragraph{Inference settings.}
All results use the base policies without CAS or other post-switch
interventions. On LIBERO-Goal, we follow each policy's standard evaluation
setup. Table~\ref{tab:obs1_inference} summarizes the prediction horizons and
the number of actions executed per policy query on both LIBERO-Goal and
CALVIN-Custom. These settings are fixed across switching times.

\begin{table}[t]
    \centering
    \caption{
    Prediction horizons and actions executed per policy query in the
    Observation~1 online evaluations. For LingBot-VA, $12/16$ denotes 12 actions
    in the first chunk after a reset and 16 actions in subsequent streaming chunks.
    }
    \label{tab:obs1_inference}
    \small
    \setlength{\tabcolsep}{5pt}
    \begin{tabular}{@{}lccc@{}}
        \toprule
        Policy & Predicted horizon & LIBERO actions/query & CALVIN actions/query \\
        \midrule
        OpenVLA-OFT & 8       & 8       & 8 \\
        $\pi_0$     & 50      & 5       & 5 \\
        $\pi_{0.5}$ & 10      & 5       & 5 \\
        Fast-WAM    & 32      & 10      & 5 \\
        LingBot-VA  & $12/16$ & $12/16$ & $12/16$ \\
        \bottomrule
    \end{tabular}
\end{table}

\subsection{Real-World Experiment Details}
\label{real_world_exp_details}

We provide additional implementation details for the real-robot experiments reported in Table~\ref{tab:real_world}. The experiments are conducted on an AIRBOT Play 6-DoF robot arm equipped with a parallel-jaw gripper. We evaluate the same mid-execution instruction-switching setting as in the main text, where the robot is first instructed to lift one colored block and the instruction is then switched to lifting the other block. We test both switching directions, blue-to-red and red-to-blue, at switching steps $t_s=20$ and $t_s=30$, with 20 trials for each setting. For a fair comparison, the base policy and CAS use matched object placements.

The deployed policy is a $\pi_0$ flow-matching policy initialized from the $\pi_0$ base checkpoint and finetuned on our real-world block-lifting demonstrations. We train a single checkpoint jointly on two tasks, \emph{pick and lift the red block} and \emph{pick and lift the blue block}. For each task, we collect 50 successful demonstration trajectories using a leader-follower teleoperation setup. The policy is finetuned for 60k gradient steps with a batch size of 32, using two NVIDIA H100 GPUs. The policy takes two RGB observations as input, including a static third-person camera view and a wrist camera view. Since the real-robot policy follows the standard $\pi_0$ observation interface, unused camera inputs are masked out. 

Unlike the simulation experiments, where policies output delta end-effector actions, the real-robot policy predicts delta joint actions for the six arm joints, together with a gripper command. We reconstruct absolute joint targets by adding the predicted joint deltas to the current joint configuration. CAS is applied only to the arm-action dimensions, while the gripper command is directly taken from the target-conditioned action. This follows the same principle as in simulation, where the gripper dimension is excluded from counterfactual steering. The gripper command is then discretized into open or close using a fixed threshold.

For CAS, hijacking tasks are selected using the same KNN-based trajectory-distance criterion as in simulation. We use the KNN distance threshold $\delta=0.03$ for all real-world experiments. The final CAS-corrected joint action is sent to the robot controller for execution.

\paragraph{Rationale for task selection.}
We deliberately choose two object-specific block-lifting tasks to provide a controlled evaluation of mid-execution instruction switching, rather than to benchmark general manipulation complexity. First, block lifting is a relatively simple and well-defined manipulation skill that can be reliably learned from demonstrations. This helps minimize failures arising from insufficient competence in the underlying tasks, allowing us to focus more directly on the difficulty of redirecting ongoing behavior after an instruction change. Second, the tasks have clear and objective success criteria: the robot must successfully lift the block specified by the updated instruction. Compared with more complex or extended tasks, such as cloth folding or room cleaning, this enables unambiguous evaluation of instruction-switching success.

Third, the two tasks involve blocks with identical geometry but different colors, and their instructions differ only in the target object. This design keeps the required manipulation skill and object geometry consistent while testing whether the policy can distinguish between closely related instructions and redirect its actions toward the newly specified target. Finally, the experimental setup allows careful control of physical conditions: \textbf{for each corresponding Base--CAS comparison, we match object positions and orientations as closely as possible, while varying object placements across trials.} In contrast, more complex manipulation tasks, such as cloth folding, involve deformable objects whose configurations are difficult to reproduce consistently between the corresponding Base and CAS evaluations. Such variations can introduce additional confounding factors, making it harder to attribute performance differences specifically to the instruction-switching intervention. Our use of rigid blocks therefore enables more controlled and directly comparable evaluations of CAS's effectiveness.

Together with bidirectional instruction switching and evaluation at two execution stages, these design choices provide a focused physical validation of CAS while minimizing confounding factors unrelated to instruction switching.

\subsection{Self-Training Details}
\label{app:self_training_details}

\paragraph{Controlled comparison.}
Our main one-round self-training comparison disentangles two benefits of CAS: a higher yield of successful switching rollouts and, more importantly, greater training value of the resulting trajectories. \emph{Base} denotes the original task-finetuned checkpoint. Vanilla-ST finetunes Base on successful switching rollouts collected without CAS, whereas CAS-ST finetunes the same Base checkpoint on successful CAS-guided rollouts. Both finetuned policies are evaluated without CAS, so the comparison reflects what is learned from the collected trajectories rather than inference-time steering.

For the main one-round self-training comparison, we roll out the policies in settings that follow the same task-pair and switching protocol as LIBERO-Goal-Switch and CALVIN-Custom-Switch, while keeping the collection episodes disjoint from those used for evaluation. We retain only rollouts that successfully complete the switched target task.

Under the same collection settings, CAS consistently yields a larger pool of successful trajectories than the Base policy. To compare the training value of these trajectories under the same data budget, we retain all successful Vanilla rollouts and uniformly sample an equal-sized subset from the CAS success pool using seed 0. Vanilla-ST and CAS-ST are therefore trained on the same number of self-training trajectories with matched finetuning settings.

\paragraph{Self-training data collection.}
For LIBERO-Goal-Switch, we collect self-training trajectories at $t_s=25$ using 450 distinct official LIBERO-Goal initial states. We cover all 90 ordered source--target task pairs defined in Appendix~\ref{app:libero_goal_switch_details}, with five initial states assigned to each pair. Vanilla and CAS use identical episode--state pairings. For $\pi_{0.5}$, we use environment seed 7, while Fast-WAM uses inference seeds 42 and 43.

For CALVIN-Custom-Switch, we generate collection states following the switching-state construction in Appendix~\ref{app:calvin_custom_switch_details}, using seed 42. We collect trajectories in the $[9,15]$ switching range, with 40 states for each of the 12 ordered source--target task pairs, yielding 480 rollout attempts per policy. Vanilla and CAS use the same saved intermediate states.

For both benchmarks, only the post-switch portion of each retained successful trajectory is used for self-training, and all retained samples are conditioned on the target instruction $I_{\mathrm{tgt}}$. Table~\ref{tab:st_data_summary} summarizes the collection settings and resulting training-data composition.

\begin{table}[!htbp]
\centering
\caption{Self-training data collection and dataset composition for the main one-round results. Vanilla-ST and CAS-ST use matched numbers of retained self-training trajectories.}
\label{tab:st_data_summary}
\small
\setlength{\tabcolsep}{1.0pt}
\renewcommand{\arraystretch}{1.08}
\begin{tabular*}{\textwidth}{@{\extracolsep{\fill}}lcccc@{}}
\toprule
& \multicolumn{2}{c}{LIBERO-Goal-Switch} & \multicolumn{2}{c}{CALVIN-Custom-Switch} \\
\cmidrule(lr){2-3}\cmidrule(lr){4-5}
Setting & $\pi_{0.5}$ & Fast-WAM & $\pi_{0.5}$ & Fast-WAM \\
\midrule
Collection switching setting & $t_s=25$ & $t_s=25$ & $[9,15]$ & $[9,15]$ \\
Collection rollout attempts (each) & 450 & $2\!\times\!450$ & 480 & 480 \\
Retained self-training trajectories (each) & 148 & 221 & 104 & 80 \\
Official demonstrations & 500 & 500 & -- & -- \\
Total training trajectories (each) & 648 & 721 & 104 & 80 \\
Reported checkpoint & 10k & 10k & 20k & 20k \\
\bottomrule
\end{tabular*}
\end{table}

\rev{
\paragraph{Switching-state distributions in the original self-training data.}
\label{app:st_switching_state_distribution}
We analyze all 148 retained Vanilla trajectories and 148 retained CAS
trajectories from the original $\pi_{0.5}$ LIBERO-Goal-Switch training sets
in Table~\ref{tab:st_data_summary}. Each trajectory contributes one
proprioceptive observation immediately after the 25 source actions and before
the first target action: end-effector position, orientation, and the two
gripper joint positions. These observations are verified against the first
states in the training data. This analysis concerns the original datasets,
rather than the 223-pair experiment in Appendix~\ref{app:matched_self_training}.

Within each source task, we use an energy two-sample test to compare the
distributions. Orientations are converted from axis-angle vectors to rotation
matrices. Each of the three feature blocks is divided by its pooled RMS
distance over distinct observation pairs, independently of method labels,
and the normalized blocks receive equal weight. The statistic is twice the
mean cross-group Euclidean distance minus the two mean within-group
distances; the latter include zero self-distances. The overall statistic
averages the ten source-task statistics equally.

Significance is assessed with 9,999 label permutations. The 102 cases retained
by both methods are paired by collection repeat, source task, target task,
and episode-initial-state hash; labels are exchanged within these pairs.
Labels of the remaining observations are shuffled within each source,
preserving group counts. The shared episode initialization does not require
the two independently executed prefixes to reach identical switching states.
An additional source--target-stratified test restricts the remaining label
permutations to the same target task. Supplementary position-only and
position-plus-orientation tests use 99,999 permutations. For each feature
choice, source-specific tests use Benjamini--Hochberg correction across the
ten sources.

\begin{table}[!htbp]
\centering
\caption{Switching-point proprioceptive distribution tests on the
original $\pi_{0.5}$ self-training data (148 trajectories per method).
Reported $p$-values use the aggregate statistic across ten source tasks.}
\label{tab:st_switching_state_distribution}
\small
\setlength{\tabcolsep}{6pt}
\begin{tabular}{lcc}
\toprule
Features & \makecell{Source-stratified\\$p$} & \makecell{Source--target-stratified\\$p$} \\
\midrule
Position, orientation, and gripper & 0.612 & 0.910 \\
Position and orientation & 0.344 & 0.746 \\
Position only & 0.056 & 0.312 \\
\bottomrule
\end{tabular}
\end{table}

Table~\ref{tab:st_switching_state_distribution} shows no statistically
detectable overall difference in switching-point proprioceptive
distributions. The position-only source-level analysis identifies a localized
difference for opening the middle drawer ($p=0.00403$, corrected $q=0.0403$):
the group centroids differ by 0.81\,cm, compared with a pooled within-group
RMS pairwise distance of 4.53\,cm. The paired experiment in
Appendix~\ref{app:matched_self_training} directly controls switching states
and task composition when comparing the training value of the trajectories.
}

\rev{
\paragraph{Action statistics and trajectory lengths.}
\label{app:st_action_analysis}
We analyze the complete successful post-switch trajectories used for
$\pi_{0.5}$ self-training on LIBERO-Goal: the original 148 trajectories per
method and the 223 strictly paired trajectories per method from
Appendix~\ref{app:matched_self_training}. Official demonstrations are excluded
from this analysis. All 742 recorded action sequences are verified against
the corresponding training data.

For each trajectory, we compute the mean L2 magnitude of the three-dimensional
translational and rotational commands and the mean L2 difference between
consecutive commands in each block. We then average each statistic equally
across trajectories, so longer trajectories do not receive greater weight.
Values are in normalized controller-input units, before internal controller
clipping and scaling. Adjacent-command differences capture changes in both
direction and magnitude.

\begin{table}[!htbp]
\centering
\caption{Action statistics of successful post-switch trajectories.
Each entry is an equal-weight mean over trajectories; the original and paired
datasets contain 148 and 223 trajectories per method, respectively.}
\label{tab:st_action_statistics}
\small
\setlength{\tabcolsep}{4pt}
\begin{tabular*}{\textwidth}{@{\extracolsep{\fill}}lrrrr@{}}
\toprule
& \multicolumn{2}{c}{Original} & \multicolumn{2}{c}{Paired} \\
\cmidrule(lr){2-3}\cmidrule(lr){4-5}
Action statistic & Vanilla & CAS & Vanilla & CAS \\
\midrule
Translational magnitude & 0.6688 & 0.7598 & 0.6635 & 0.7673 \\
Rotational magnitude & 0.07653 & 0.09326 & 0.07670 & 0.09114 \\
Adjacent translational-command change & 0.1050 & 0.1437 & 0.1016 & 0.1418 \\
Adjacent rotational-command change & 0.01684 & 0.02434 & 0.01621 & 0.02323 \\
\bottomrule
\end{tabular*}
\end{table}

Table~\ref{tab:st_action_statistics} shows consistent patterns across both
datasets. Relative to Vanilla, CAS increases mean translational magnitude by
13.6\%/15.6\% and rotational magnitude by 21.9\%/18.8\% in the original/paired
datasets. Adjacent translational- and rotational-command changes increase by
36.9\%/39.6\% and 44.5\%/43.3\%, respectively. These larger changes may reflect
more pronounced corrective adjustments and redirection under CAS; they
describe commanded actions rather than physical motion smoothness.

To measure the steering intervention directly, we compare each CAS command
with its corresponding unsteered target prediction from the same policy
query. Averaging first within trajectories and then across trajectories,
CAS changes the translational command direction by $10.55^\circ$/$10.13^\circ$
in the original/paired datasets, while preserving the corresponding gripper
command exactly. This directly shows that CAS modifies direction as well as
magnitude, supporting the interpretation that larger adjacent-command
changes may partly reflect corrective redirection.

\begin{table}[!htbp]
\centering
\caption{Mean successful post-switch trajectory length, measured
in executed target-action steps. All retained trajectories are used in full,
without truncation or downsampling.}
\label{tab:st_trajectory_lengths}
\small
\setlength{\tabcolsep}{8pt}
\begin{tabular}{lrrrr}
\toprule
Dataset & Trajectories per method & Vanilla & CAS & Reduction \\
\midrule
Original & 148 & 130.0 & 119.1 & 8.4\% \\
Paired & 223 & 123.3 & 107.1 & 13.1\% \\
\bottomrule
\end{tabular}
\end{table}

CAS also completes successful post-switch segments in fewer steps on average
(Table~\ref{tab:st_trajectory_lengths}). In the paired dataset, CAS is shorter
in 173 of 223 cases (77.6\%), longer in 46, and tied in four. The qualitative
comparisons in Appendix~\ref{app:better_ST} illustrate more suitable contact
and grasp poses, fewer failed grasp attempts, and more direct recovery.
These behaviors are consistent with the shorter successful trajectories.

Post-switch action and state distributions are expected to differ because
CAS modifies the base policy's actions and thereby changes subsequent state
visitation. We therefore analyze switching-point states to assess the
starting conditions before intervention, and use the paired experiment to
hold these states and task composition fixed while evaluating the training
value of the resulting successful trajectories.
}

\paragraph{CAS rollout settings.}
For CAS-guided data collection in the main one-round experiments, we use $K=30$ for the trajectory distance $d_i(p_t)$ and set the hijacking-task threshold to $\delta=0.1$. All other CAS inference settings follow Sec.~3.

\paragraph{Finetuning settings.}
For the main one-round comparison, Vanilla-ST and CAS-ST are initialized from the same Base checkpoint and use identical finetuning settings within each policy--benchmark comparison. Table~\ref{tab:st_optimization} summarizes the training configurations.

\begin{table}[H]
\centering
\caption{Finetuning settings for the main one-round Vanilla-ST and CAS-ST comparisons. Settings are matched within each policy--benchmark comparison; reported checkpoints are listed as LIBERO / CALVIN.}
\label{tab:st_optimization}
\small
\setlength{\tabcolsep}{3.0pt}
\renewcommand{\arraystretch}{1.08}
\begin{tabular*}{\textwidth}{@{\extracolsep{\fill}}lll@{}}
\toprule
Setting & $\pi_{0.5}$ & Fast-WAM \\
\midrule
Initialization & Task-finetuned base checkpoint & Task-finetuned base checkpoint \\
Fine-tuning objective & Full-parameter policy fine-tuning & Joint video--action fine-tuning \\
Global batch size & 32 & 32 \\
Training seed & 42 & 42 \\
Optimizer & AdamW, gradient clip 1.0 & AdamW \\
Peak learning rate & $5\times10^{-5}$ & $1\times10^{-4}$ \\
Warmup & 10k steps & 1k steps; cosine decay \\
EMA decay & 0.999 & -- \\
Video/action loss weights & -- & $1/1$ \\
Reported checkpoint & 10k / 20k & 10k / 20k \\
\bottomrule
\end{tabular*}
\end{table}

\paragraph{Iterative CAS self-training.}
We further study whether CAS-guided self-training can be applied iteratively to progressively improve recovery from later and more challenging switching states. Round 1 corresponds to the $\pi_{0.5}$ CAS-ST model on LIBERO-Goal-Switch described above: starting from the task-finetuned $\pi_{0.5}$ checkpoint, we collect CAS-guided trajectories at $t_s=25$ and finetune the policy to obtain the reported 10k checkpoint.

For Round 2, we initialize both data collection and subsequent finetuning from the Round-1 10k checkpoint and move the collection switch to the later setting $t_s=40$. We perform 450 CAS-guided rollout attempts. Among these rollouts, 154 successfully complete the switched target task, and all 154 successful trajectories are retained. As before, only the post-switch portion of each retained successful trajectory is used for self-training and is conditioned on the target instruction $I_{\mathrm{tgt}}$.

The Round-2 finetuning dataset consists of the same 500 official LIBERO-Goal demonstrations together with the 154 newly collected Round-2 CAS trajectories, for a total of 654 training trajectories. Importantly, we do not replay the 148 retained Round-1 self-training trajectories in the Round-2 finetuning dataset. Instead, Round 1 influences the second round through two complementary paths. First, the Round-1 10k checkpoint initializes the Round-2 policy. Second, the post-switch portions of the 148 retained Round-1 CAS trajectories are added to the task-trajectory library used for KNN-based hijacking detection during Round-2 CAS data collection. Thus, Round 2 uses an updated trajectory library containing recovery experience from the preceding round, while its policy finetuning data contain only the official demonstrations and the newly collected Round-2 CAS trajectories.

We finetune the Round-1 checkpoint for an additional 10k steps using the same $\pi_{0.5}$ optimization configuration as in Round 1. The resulting Round-2 checkpoint is evaluated without CAS, so improvements after the second round reflect recovery behavior incorporated into the policy rather than inference-time steering.

\paragraph{Evaluation.}
For the main one-round comparison, we evaluate both the data-collection setting and an unseen later-switch setting: $t_s=25$ and $t_s=30$ for LIBERO-Goal-Switch, and $[9,15]$ and $[15,21]$ for CALVIN-Custom-Switch.

With matched numbers of self-training trajectories, CAS-ST consistently outperforms Vanilla-ST across both benchmarks and both policies, at both the data-collection settings and the unseen later switches. Because CAS is disabled during evaluation, these gains reflect recovery behaviors learned from the CAS-guided trajectories rather than inference-time steering. The consistent improvements at unseen later switches further demonstrate that the learned de-hijacking behavior generalizes beyond the switching regime used for self-training.

For iterative CAS-ST, we evaluate the Round-1 and Round-2 checkpoints at $t_s\in\{40,50,60\}$ on LIBERO-Goal-Switch. These switching steps are all later than the Round-1 collection setting at $t_s=25$. For Round 2, $t_s=40$ matches its data-collection setting, while $t_s=50$ and $t_s=60$ are unseen later switching conditions. Both checkpoints are evaluated without CAS.

\subsection{PCD Implementation Details}
\label{app:pcd_implementation_details}

\paragraph{PCD comparison method.}
Policy Contrastive Decoding (PCD)~\citep{wu2025policy} is a training-free inference-time method that contrasts the action distribution predicted from the original observation with that predicted from an object-masked observation. The object-masked observation removes task-relevant objects from the image, so the resulting contrast is intended to amplify actions that depend on the task object rather than on spurious visual cues. We apply PCD to the flow-matching-based policies $\pi_0$ and $\pi_{0.5}$ using KDE-based probabilistic modeling over sampled actions. We keep the original PCD decoding procedure unchanged and set the contrast strength to $\alpha=0.2$, following the setting used for $\pi_0$ in the original paper.

\paragraph{Adapting PCD to our benchmarks.}
The main adaptation needed in our setting is the construction of object-masked observations. PCD requires identifying the task-relevant objects to be removed, but our benchmarks provide only RGB observations and natural-language instructions at test time. We therefore implement an instruction-grounded open-vocabulary detection pipeline. Given an instruction, we first extract object names using a rule-based parser tailored to the instruction patterns in LIBERO-Goal. Single-object commands such as \emph{open X}, \emph{push X}, or \emph{pick up X} return the manipulated object; placement commands such as \emph{put X on Y} or \emph{stack X on Y} return both the manipulated object and the receptacle; and compound instructions joined by ``and'' are split into sub-commands, from which all referenced objects are collected.

The extracted object phrases are then normalized before being passed to the detector.
We remove articles and location modifiers after the object noun; for example,
\emph{the bowl between the plate and the ramekin} is rewritten as \emph{bowl}.
Since bare LIBERO object names are often difficult for open-vocabulary detectors
to ground reliably, we further apply a small hand-built alias map that rewrites
them into more visually grounded queries, such as \emph{bowl} to ``black bowl'',
\emph{plate} to ``white plate'', \emph{block} to ``wooden cube'', and
drawer-specific names to ``drawer''. The resulting queries are sent to Grounding
DINO for object detection, followed by segmentation and inpainting to construct
the contrast observation. We also preserve the robot arm during masking, so that
only task objects are removed. If no valid object is parsed or detected at a step,
we use the original observation as the contrast observation, in which case PCD
reduces to the base policy for that step.

\clearpage
\section{\rev{Comparisons with Related Methods}}
\label{app:related_method_comparisons}
\begingroup
We complement the PCD comparison with evaluations of the language-grounding intervention CAG-TF, AVA-VLA, OpenVLA-OFT+ from LIBERO-Plus, and ReSteer. All results in this section use LIBERO-Goal-Switch, with 90 ordered non-diagonal source--target task pairs per switching time. Each evaluated policy executes the source phase before receiving the target instruction. We report target-task success rates (\%); higher is better. CAS, when enabled, is applied only after the instruction switch.

\subsection{Language-Grounding Intervention: CAG-TF}
\label{app:cag_tf_comparison}
CAG-TF~\citep{fang2026vision}, termed TF-CAG in the latest version of the work, is a training-free language-grounding intervention. It contrasts actions predicted with and without language conditioning to strengthen the influence of the target instruction. CAS also uses inference-time contrastive steering, but constructs its counterfactual reference from state-relevant competing task instructions, selected and weighted using trajectory distances. This comparison examines whether explicitly targeting interfering task priors provides benefits beyond amplifying language conditioning.

\begin{table}[htbp]
\centering
\small
\caption{\rev{Comparison with CAG-TF on LIBERO-Goal-Switch. Each policy is evaluated without steering (Base), with CAG-TF, and with CAS. CAS achieves higher target-task success rates than CAG-TF across all three policies and all four switching times.}}
\label{tab:cag_tf_comparison}
\begin{tabular*}{\linewidth}{@{\extracolsep{\fill}}llrrrr@{}}
\toprule
Policy & Method & $t_s=10$ & $t_s=15$ & $t_s=20$ & $t_s=25$ \\
\midrule
\multirow{3}{*}{OpenVLA-OFT} & Base & 41.1 & 22.2 & 12.2 & 12.2 \\
& CAG-TF & 35.6 & 24.4 & 16.7 & 11.1 \\
& CAS & \textbf{65.6} & \textbf{31.1} & \textbf{21.1} & \textbf{15.6} \\
\midrule
\multirow{3}{*}{$\pi_0$} & Base & 72.2 & 32.2 & 30.0 & 13.3 \\
& CAG-TF & 66.7 & 36.7 & 25.6 & 16.7 \\
& CAS & \textbf{77.8} & \textbf{51.1} & \textbf{37.8} & \textbf{23.3} \\
\midrule
\multirow{3}{*}{$\pi_{0.5}$} & Base & 90.0 & 73.3 & 52.2 & 34.4 \\
& CAG-TF & 91.1 & 82.2 & 56.7 & 41.1 \\
& CAS & \textbf{96.7} & \textbf{85.6} & \textbf{62.2} & \textbf{44.4} \\
\bottomrule
\end{tabular*}
\end{table}

Table~\ref{tab:cag_tf_comparison} shows that CAG-TF improves $\pi_{0.5}$ at every switching time, while its effects on OpenVLA-OFT and $\pi_0$ are mixed. CAS outperforms CAG-TF in all 12 policy--switching-time settings. These results support the value of contrasting against state-relevant competing tasks in mid-execution switching, beyond the gains available from a language-grounding intervention alone.

\subsection{AVA-VLA and OpenVLA-OFT+}
\label{app:ava_oftplus_comparison}
\paragraph{Evaluation setup.}
We evaluate the officially released AVA-VLA model~\citep{xiao2025ava}, jointly trained on the four LIBERO suites, using its native active visual attention and temporal inference components. At the instruction switch, we discard the remaining source-action chunk and reset its temporal context for the new instruction. We also evaluate the released mix-SFT OpenVLA-OFT+ model from LIBERO-Plus~\citep{fei25libero-plus}, using its corresponding action normalization. These policies are evaluated on our instruction-switching benchmark, rather than on the perturbation benchmark of LIBERO-Plus. Both use eight-step action replanning and a 300-step target-task execution budget. We additionally apply CAS to OpenVLA-OFT+ without further policy training.

\begin{table}[htbp]
\centering
\small
\setlength{\tabcolsep}{4pt}
\caption{\rev{Comparison with released AVA-VLA and OpenVLA-OFT+ models on LIBERO-Goal-Switch. CAS improves both OpenVLA-OFT and OpenVLA-OFT+ at every switching time without additional policy training. The average is taken over the four switching times.}}
\label{tab:ava_oftplus_comparison}
\begin{tabular*}{\linewidth}{@{\extracolsep{\fill}}lrrrrr@{}}
\toprule
Method & $t_s=10$ & $t_s=15$ & $t_s=20$ & $t_s=25$ & Average \\
\midrule
OpenVLA-OFT & 41.1 & 22.2 & 12.2 & 12.2 & 21.9 \\
AVA-VLA & 40.0 & 18.9 & 10.0 & 6.7 & 18.9 \\
OpenVLA-OFT+ (LIBERO-Plus) & 45.6 & 17.8 & 12.2 & 6.7 & 20.6 \\
\midrule
OpenVLA-OFT + CAS & \textbf{65.6} & \textbf{31.1} & \textbf{21.1} & \textbf{15.6} & \textbf{33.3} \\
OpenVLA-OFT+ + CAS & 53.3 & 30.0 & 18.9 & 8.9 & 27.8 \\
\bottomrule
\end{tabular*}
\end{table}

\paragraph{Results.}
Table~\ref{tab:ava_oftplus_comparison} shows that mid-execution instruction switching remains challenging for the released AVA-VLA and OpenVLA-OFT+ policies, which achieve average success rates of 18.9\% and 20.6\%, respectively. CAS improves OpenVLA-OFT+ at all four switching times, increasing its average success rate from 20.6\% to 27.8\%. This demonstrates that counterfactual steering can complement the training improvements in OpenVLA-OFT+. Since these released models differ in training and architecture, the cross-model comparison assesses the complete policies; the paired OpenVLA-OFT+ comparison directly measures the benefit of adding CAS to that policy.

\subsection{ReSteer}
\label{app:resteer_comparison}
\paragraph{Implementation and training.}
We adapt the authors' public ReSteer implementation~\citep{chen2026resteer} to $\pi_{0.5}$ and LIBERO-Goal-Switch, using its step-matched SteerGen bridge generation and success-rate-based self-refining behavioral cloning (SRBC). These are models trained by us using the public code, rather than author-released pretrained checkpoints. Starting from the task-finetuned base policy and 500 original LIBERO-Goal demonstrations, the first stage adds 148 bridge trajectories covering all 90 ordered task pairs. The second stage starts from the first-stage model and adds 154 successful switching rollouts. These additional trajectory counts correspond to those used in the first and second CAS-ST rounds, respectively.

The comparison evaluates the resulting training pipelines with matched numbers of additional trajectories. SRBC uses complete successful rollouts, whereas CAS-ST trains on post-switch target-task segments. Evaluation uses five-step action replanning and a 300-step target-task budget. SteerGen, SRBC, and CAS-ST are evaluated without inference-time CAS.

\begin{table}[htbp]
\centering
\small
\caption{\rev{Inference-time CAS and our adaptation of ReSteer's SteerGen on LIBERO-Goal-Switch. CAS matches or exceeds SteerGen at all four switching times without additional policy training.}}
\label{tab:resteer_early_comparison}
\begin{tabular*}{\linewidth}{@{\extracolsep{\fill}}lrrrr@{}}
\toprule
Method & $t_s=10$ & $t_s=15$ & $t_s=20$ & $t_s=25$ \\
\midrule
Base $\pi_{0.5}$ & 90.0 & 73.3 & 52.2 & 34.4 \\
ReSteer: SteerGen & 93.3 & 82.2 & \textbf{62.2} & 38.9 \\
$\pi_{0.5}$ + CAS & \textbf{96.7} & \textbf{85.6} & \textbf{62.2} & \textbf{44.4} \\
\bottomrule
\end{tabular*}
\end{table}

\begin{table}[htbp]
\centering
\small
\caption{\rev{Comparison of ReSteer and CAS-guided self-training at later switching times on LIBERO-Goal-Switch. ReSteer results use our adaptation of the authors' public code. All trained policies are evaluated without inference-time CAS.}}
\label{tab:resteer_late_comparison}
\begin{tabular*}{\linewidth}{@{\extracolsep{\fill}}lrrr@{}}
\toprule
Method & $t_s=40$ & $t_s=50$ & $t_s=60$ \\
\midrule
Base $\pi_{0.5}$ & 28.9 & 24.4 & 17.8 \\
CAS-ST (one round) & 36.7 & 32.2 & 28.9 \\
ReSteer: SteerGen & 26.7 & 27.8 & 26.7 \\
ReSteer: SteerGen + SRBC & 31.1 & 32.2 & 30.0 \\
CAS-ST (two rounds) & \textbf{46.7} & \textbf{42.2} & \textbf{37.8} \\
\bottomrule
\end{tabular*}
\end{table}

\paragraph{Results.}
SteerGen improves over the base policy at each of the four main switching times (Table~\ref{tab:resteer_early_comparison}). Inference-time CAS achieves higher success rates at three times and matches SteerGen at $t_s=20$, demonstrating its effectiveness without switching-specific policy training. At later switches (Table~\ref{tab:resteer_late_comparison}), SRBC improves over SteerGen at all three times. Two-round CAS-ST further exceeds SRBC by 15.6, 10.0, and 7.8 percentage points at $t_s=40,50,60$, respectively. These results support CAS both as an immediate inference-time intervention and as a source of useful supervision for subsequent self-training.

\endgroup

\stopcontents[appendix]

\end{document}

%% file: TABLE_MAIN_RESULTS.tex
\begin{table*}[t]
\centering
\caption{
\textbf{Simulation results for inference-time instruction switching.}
We report target-task success rates (\%) on LIBERO-Goal-Switch and CALVIN-Custom-Switch.
% We compare CAS with the corresponding base policy and, where applicable, the training-free PCD baseline.
}
\label{tab:main_results}

\resizebox{0.9\linewidth}{!}{%
\begin{tabular}{llccccccc}
\toprule
\multirow{2}{*}{Policy} & \multirow{2}{*}{Method}
& \multicolumn{4}{c}{LIBERO-Goal-Switch}
& \multicolumn{3}{c}{CALVIN-Custom-Switch} \\
\cmidrule(lr){3-6}\cmidrule(lr){7-9}
& & $t_s=10$ & $t_s=15$ & $t_s=20$ & $t_s=25$
& $[9,15]$ & $[15,21]$ & $[21,27]$ \\
\midrule
\multirow{2}{*}{OFT} & Base & 41.1 & 22.2 & 12.2 & 12.2 & 11.3 & 5.4 & 0.0 \\
 & + CAS & \textbf{65.6} & \textbf{31.1} & \textbf{21.1} & \textbf{15.6} & \textbf{25.4} & \textbf{13.8} & \textbf{4.6} \\
\midrule
\multirow{3}{*}{$\pi_0$} & Base & 72.2 & 32.2 & 30.0 & 13.3 & 22.1 & 6.7 & 0.0 \\
 & + PCD & 62.2 & 35.6 & 28.9 & 17.8 & 22.5 & 5.4 & 0.0 \\
 & + CAS & \textbf{77.8} & \textbf{51.1} & \textbf{37.8} & \textbf{23.3} & \textbf{73.8} & \textbf{45.4} & \textbf{20.4} \\
\midrule
\multirow{3}{*}{$\pi_{0.5}$} & Base & 90.0 & 73.3 & 52.2 & 34.4 & 23.3 & 5.8 & 0.8 \\
 & + PCD & 95.6 & 74.4 & 51.1 & 32.2 & 26.3 & 5.4 & 0.0 \\
 & + CAS & \textbf{96.7} & \textbf{85.6} & \textbf{62.2} & \textbf{44.4} & \textbf{77.1} & \textbf{54.6} & \textbf{14.2} \\
\midrule
\multirow{2}{*}{Fast-WAM} & Base & 82.2 & 56.7 & 33.3 & 20.0 & 16.7 & 0.4 & 0.0 \\
 & + CAS & \textbf{93.3} & \textbf{67.8} & \textbf{44.4} & \textbf{34.4} & \textbf{49.6} & \textbf{19.6} & \textbf{0.4} \\
\midrule
\multirow{2}{*}{LingBot-VA} & Base & 94.4 & 82.2 & 67.8 & 58.9 & 25.0 & 9.6 & 0.0 \\
 & + CAS & \textbf{96.7} & \textbf{90.0} & \textbf{75.6} & \textbf{65.6} & \textbf{40.0} & \textbf{25.0} & \textbf{17.1} \\
\bottomrule
\end{tabular}%
}

\end{table*}

%% file: TABLE_SELF_TRAINING.tex
% Self-training main table; source values are archived under draft/ST_main.
\begin{table}[t]
\centering
\caption{
\textbf{Evaluation of CAS-guided self-training.}
We compare CAS-ST with Vanilla-ST using equal numbers of successful trajectories
at the data-collection setting and an unseen later switch.
}
\label{tab:ST_main}
\footnotesize
\renewcommand{\arraystretch}{1.04}
\setlength{\tabcolsep}{0.4pt}
\begin{tabular*}{\textwidth}{@{\extracolsep{\fill}}lcccccccc@{}}
\toprule
& \multicolumn{4}{c}{\textbf{Data-collection settings}} & \multicolumn{4}{c}{\textbf{Generalization to Unseen Later Switches}} \\
\cmidrule(lr){2-5}\cmidrule(lr){6-9}
& \multicolumn{2}{c}{{\fontsize{8.5}{9.5}\selectfont LIBERO-Goal-Switch}} & \multicolumn{2}{c}{{\fontsize{8.5}{9.5}\selectfont CALVIN-Custom-Switch}} & \multicolumn{2}{c}{{\fontsize{8.5}{9.5}\selectfont LIBERO-Goal-Switch}} & \multicolumn{2}{c}{{\fontsize{8.5}{9.5}\selectfont CALVIN-Custom-Switch}} \\
\cmidrule(l{0.15em}r{0.15em}){2-3}\cmidrule(l{0.15em}r{0.15em}){4-5}\cmidrule(l{0.15em}r{0.15em}){6-7}\cmidrule(l{0.15em}r{0.15em}){8-9}
Method & $\pi_{0.5}$ & Fast-WAM & $\pi_{0.5}$ & Fast-WAM & $\pi_{0.5}$ & Fast-WAM & $\pi_{0.5}$ & Fast-WAM \\
\midrule
Base & 34.4 & 20.0 & 23.3 & 16.7 & 27.8 & 16.7 & 5.8 & 0.4 \\
Vanilla-ST & 48.9 & 24.4 & 32.5 & 21.3 & 32.2 & 21.1 & 8.3 & 7.9 \\
CAS-ST & \textbf{52.2} & \textbf{27.8} & \textbf{57.1} & \textbf{50.0} & \textbf{40.0} & \textbf{27.8} & \textbf{23.8} & \textbf{15.4} \\
\bottomrule
\end{tabular*}
\end{table}